%% file: main.tex
\documentclass[pdflatex,sn-apa]{sn-jnl}

\usepackage{graphicx}
\usepackage{multirow}
\usepackage{amsmath,amssymb}
\usepackage{booktabs}
\usepackage{hyperref}
\usepackage[nameinlink,noabbrev]{cleveref}
\usepackage{amsthm}
\usepackage{tabularx}

\usepackage{xcolor}

\newtheorem{definition}{Definition}

\newcommand{\method}{\texttt{FiLSim}}

\begin{document}

\title[Scaling Graph Similarity Learning with Graph Neural Networks]{Scaling Graph Similarity Learning with Graph Neural Networks}

\author*[1]{\fnm{Ahmet} \sur{T{\"u}zen}}\email{ahmet.tuzen@ntnu.no}

\author[1]{\fnm{Helge} \sur{Langseth}}\email{helge.langseth@ntnu.no}

\author[1]{\fnm{Kjetil} \sur{N{\o}rv{\aa}g}}\email{noervaag@ntnu.no}

\affil*[1]{\orgdiv{Department of Computer Science}, \orgname{Norwegian University of Science and Technology}, \orgaddress{\city{Trondheim}, \country{Norway}}}

\abstract{
Graph similarity computation is an important task in graph mining with many real-world applications, such as chemical compound analysis, program dependence comparison, and social network analysis. 
The problem is computationally challenging: Graph Edit Distance (GED) and Maximum Common Subgraph (MCS), which quantify the distance between graphs and identify their largest shared subgraph, respectively, are NP-hard to compute exactly.
To address these computational challenges, Graph Neural Networks (GNNs) have been applied to this problem.
However, existing GNN-based approaches have primarily focused on small-scale graph datasets, while scalability to larger scales remains underexplored.
In this work, we propose \method{}, a scalable GNN-based framework, and investigate whether a simplified GNN pipeline can maintain competitive performance on standard benchmarks while scaling to large graphs. 
We additionally propose a sequential graph-editing strategy to construct large-scale graph datasets with reliable GED labels to evaluate scalability.
The proposed framework is evaluated on four benchmark datasets and four newly created large-scale datasets, demonstrating its effectiveness and scalability.
\method{} remains accurate on graphs with more than 150{,}000 nodes, where the baseline methods run out of memory or fail to learn, and scores all test pairs within seconds on a single GPU.
}


\maketitle

\input{chapter1}
\input{chapter3}
\input{chapter2}
\input{chapter4}

\input{chapter5_1}
\input{chapter5_2}

\section{Conclusion}

In this work, we investigated the scalability of graph similarity learning by showing that a simplified pipeline without expensive cross-graph node interactions can achieve strong performance on standard benchmarks while scaling to large graphs. Our approach adopts a FiLM-based prediction module and extends pairwise GED prediction to graphs with more than 150{,}000 nodes, where none of the baselines remain accurate, and to graphs averaging over 300{,}000 nodes in the extended experiments in the appendix. We further show that asymptotic efficiency alone does not guarantee scalability: existing graph-level methods of equivalent complexity fail to converge, degrade ranking quality, or exhaust memory at this scale, whereas \method{} remains accurate.

Through experiments on four benchmark datasets and four large-scale synthetic datasets, we demonstrated that \method{} achieves strong performance across diverse graph types and sizes. Our large-scale dataset generation methodology provides a systematic approach for creating large-scale datasets with GED upper-bound labels for all pairs, without requiring expensive pairwise distance computations. The generated datasets allow graph similarity methods to be evaluated on graphs with hundreds of thousands of nodes.

Our study has several limitations, each of which points to future work. Our labels on the large-scale datasets are upper bounds on the GED rather than exact values, so the reported errors are measured against these bounds. The large-scale dataset generation only covers GED labels, and extending it to MCS labels and evaluation remains open. Our framework predicts scalar similarity scores and does not recover edit paths or per-operation counts; applications requiring interpretable edit sequences would need structured-output methods.

\bibliography{references}

\newcounter{mainfig}\setcounter{mainfig}{\value{figure}}
\newcounter{maintab}\setcounter{maintab}{\value{table}}
\begin{appendices}
\renewcommand{\thefigure}{\arabic{figure}}
\renewcommand{\thetable}{\arabic{table}}
\setcounter{figure}{\value{mainfig}}
\setcounter{table}{\value{maintab}}

\crefalias{section}{appendix}
\crefalias{subsection}{appendix}

\input{appendix}

\end{appendices}

\end{document}

%% file: chapter1.tex
\section{Introduction}

Graphs represent entities and the relations between them, and many real-world data, such as social and biological networks, are naturally described in this form. Graph structures are therefore widely adopted across many areas, such as social networks and biological networks, and are a common data structure in machine learning and data mining.

Graph matching is the problem of finding an optimal node-to-node correspondence between two graphs. In many real-world applications, requiring exact matching is impractical, and approximate formulations that measure graph similarity are more appropriate. This has motivated research on graph similarity learning, which aims to quantify the degree of similarity between graphs as a continuous score rather than finding correspondences. Both graph matching and graph similarity computation are computationally challenging. Graph Edit Distance (GED)~\citep{ged} quantifies the similarity between two graphs by the cost of edit operations (e.g., node and edge insertions and deletions) required to transform one graph into the other, with lower values indicating higher similarity. Maximum Common Subgraph (MCS)~\citep{mcs} identifies the largest isomorphic subgraph shared by two graphs. Computing these metrics is NP-hard and quickly becomes intractable as the graph size increases~\citep{yang2024deep}. This computational barrier limits their applicability in real-world settings where graphs may contain hundreds or thousands of nodes. A toy example of graph matching and GED is given in \Cref{app:gm_ged_example}.

Graph Neural Networks (GNNs)~\citep{gnn_1, gnn_2} are widely used for learning on graph-structured data. GNNs iteratively aggregate information from neighboring nodes, which allows them to capture both local neighborhood properties and broader graph patterns~\citep{gnn_paper}. GNNs have demonstrated strong performance across a wide range of graph-level downstream tasks, including node classification, graph classification, anomaly detection, and graph generation. Recent studies have adopted GNN architectures for graph matching and graph similarity learning~\citep{ma2021deep}. However, existing GNN-based graph similarity methods are limited in scale~\citep{survey_scale}. Most approaches are designed for and evaluated on datasets containing small graphs. Many rely on operations such as node-level cross-graph attention or pairwise node comparisons, which typically scale quadratically with graph size. In addition, the lack of large-scale benchmark datasets has prevented a systematic study of whether these methods can scale to the graph sizes encountered in domains such as social network analysis.

To address these challenges, we propose \method{} (\textbf{Fi}LM-conditioned \textbf{L}arge-graph \textbf{Sim}ilarity), a scalable GNN-based framework for graph similarity learning. Our framework combines a shared GNN-based encoder with simple feature augmentation, graph-level pooling over the outputs of all encoder layers, and a prediction module based on Feature-wise Linear Modulation (FiLM)~\citep{FiLM} that operates on graph-level embeddings. This design allows each graph to be encoded independently and avoids expensive node-level alignment. We also introduce a systematic data-generation approach that produces large-scale synthetic datasets from real graphs, which allows us to train and evaluate \method{} at previously unexplored scales. We consider static, undirected graphs whose nodes are either unlabeled or carry discrete labels. We use GED with a unit cost for every edit operation. The main contributions of this work are summarized as follows:

\begin{itemize}

\item \textbf{Scalable Pipeline:} We demonstrate that a simplified GNN-based framework replacing node-level cross-graph alignment with efficient graph-level pooling can compute graph similarity at scales of more than 150{,}000 nodes without sacrificing performance on small benchmarks.

\item \textbf{Large-Scale Graph Datasets:} We construct large-scale synthetic graph datasets from real graphs with more than 150{,}000 nodes in the main experiments and more than 300{,}000 nodes in the extended experiments, i.e., graphs more than 1000$\times$ larger than those in existing benchmark datasets. We provide a sequential approach to generating these datasets, in which each label is an upper bound on the GED by construction, and all pairs among the generated graphs are labeled.

\item \textbf{Comprehensive Evaluation:} Through experiments on all datasets, we demonstrate that our method achieves the lowest error on all benchmark datasets and is the only method that remains accurate on all large-scale datasets, whereas each baseline runs out of memory or fails to learn on at least one of them.

\item \textbf{Cost and Design Analysis:} We measure the cost of every stage of training and inference and ablate each component of the model, showing that the encoding cost grows linearly with the graph size and quantifying the contribution of each design choice.

\end{itemize}

%% file: chapter3.tex
\section{Preliminaries}
\label{sec:preliminaries}

In this section, we introduce the notation used throughout the paper, define the graph distance and similarity measures we predict, formalize the learning problem, and provide the necessary background on GNNs.

\begin{definition}[Graphs]
A graph is defined as $G = (V, E, X)$, where $V = \{v_1, v_2, \dots, v_n \} $ is a set of $n$ nodes, and $E \subseteq V \times V$ is the set of $m = |E|$ undirected edges. $X \in \mathbb{R}^{n \times d_0}$ is the node feature matrix, whose $i$-th row is the $d_0$-dimensional feature vector of $v_i$. If a graph carries a node feature matrix, it is an attributed graph. If a graph does not carry node features, i.e., $G = (V, E)$, we refer to it as a plain graph. For a pair of graphs $G_1$ and $G_2$, with $G_i = (V_i, E_i, X_i)$, we write $n_i = |V_i|$ and $m_i = |E_i|$.
\end{definition}

\begin{definition}[Graph edit distance]
Let the edit operations be node insertion, node deletion, edge insertion, edge deletion, and, for graphs with node labels, node relabeling, each with a cost of 1. $\mathrm{GED}(G_1, G_2)$ is the minimum number of edit operations that transform $G_1$ into a graph isomorphic to $G_2$.
\end{definition}

\begin{definition}[Maximum common subgraph]
$\mathrm{MCS}(G_1, G_2)$ is the number of nodes of the largest induced subgraph of $G_1$ that is isomorphic to an induced subgraph of $G_2$.
\end{definition}

Both GED and MCS are symmetric, and computing them exactly is NP-hard~\citep{aids_1, mcsplit}.

\subsection{Graph Similarity Learning}

Given a pair of graphs $(G_1, G_2)$, the goal of graph similarity learning is to learn a function $f: \mathcal{G} \times \mathcal{G} \rightarrow \mathbb{R}_{\ge 0}$ that predicts a similarity or distance $y$ between the two inputs, where $\mathcal{G}$ denotes the space of all graphs. Since GED and MCS are symmetric, $f$ should satisfy $f(G_1, G_2) = f(G_2, G_1)$.
 
The function $f$ is typically parameterized by neural networks and learned from a training set $\mathcal{D} = \{(G_1^{i}, G_2^{i}, y^{i})\}_{i=1}^{N}$, where $y^{i}$ is the ground-truth GED or MCS between $G_1^{i}$ and $G_2^{i}$.

In practice, graph similarity models are used to compare a graph against many others, e.g., to rank a collection of graphs by their similarity to a query graph, which requires scoring a large number of pairs. Learned models that compute cross-graph node interactions typically cost $O(n_1 n_2)$ per pair, which becomes computationally expensive for graphs with thousands of nodes and for collections with many graphs.

\subsection{Graph Neural Networks} \label{sec:GNN}

GNNs are a class of deep models designed to operate on graph structures. At each layer $l$, a node $u$ updates its representation $\mathbf{h}_u^{(l)}$ based on its previous representation and the aggregated message $\mathbf{agg}_u^{(l)}$ from its neighbors $\mathcal{N}(u) = \{v : (u, v) \in E\}$:
\begin{align*}
    \mathbf{h}_u^{(l)} &= \text{UPDATE}^{(l)} \left( \mathbf{h}_u^{(l-1)}, \mathbf{agg}_u^{(l)} \right), \\
    \mathbf{agg}_u^{(l)} &= \text{AGGREGATE}^{(l)} \left( \left\{ \mathbf{h}_v^{(l-1)} : v \in \mathcal{N}(u) \right\} \right).
\end{align*}

Both $\text{UPDATE}^{(l)}$ and $\text{AGGREGATE}^{(l)}$ are differentiable functions, typically implemented as neural networks, and $\text{AGGREGATE}$ is permutation-invariant (e.g., sum, mean, max). The initial representations $\mathbf{h}_u^{(0)}$ are computed from the node inputs (\Cref{sec:features}).

\textbf{Gated Graph Neural Networks.} The Gated Graph Neural Network (GGNN)~\citep{GATED} sums transformed neighbor representations and updates node states with a Gated Recurrent Unit (GRU)~\citep{GRU}:

\begin{align*}
    \mathbf{agg}_u^{(l)} &= \sum_{v \in \mathcal{N}(u)} \mathbf{\Theta}^{(l)} \mathbf{h}_v^{(l-1)}, \\
    \mathbf{h}_u^{(l)} &= \text{GRU}\left(\mathbf{agg}_u^{(l)}, \mathbf{h}_u^{(l-1)}\right),
\end{align*}

where $\mathbf{\Theta}^{(l)}$ is a learnable weight matrix. The GRU gating mechanism enables each node to decide how much of its previous representation to keep at each layer. In the original formulation, the same GRU is shared across all propagation steps.

After $L$ layers of message passing, a graph-level representation $\mathbf{z}_G$ is obtained via a readout function:
\begin{equation*}
    \mathbf{z}_G = \text{READOUT}\left(\left\{\mathbf{h}_u^{(L)} : u \in V\right\}\right).
\end{equation*}

\noindent READOUT is typically implemented using pooling methods such as sum, mean, max, or attention-based pooling. The choice of pooling mechanism affects which graph-level information is preserved, and we treat it as a hyperparameter.

%% file: chapter2.tex
\section{Related Work}
\label{sec:related}
 
\subsection{Computing Graph Similarity Measures}
 
Exact computation of graph similarity measures such as GED and MCS is computationally expensive, as both problems are NP-hard~\citep{aids_1}. A* search~\citep{astar, astar_ged} computes exact GED; however, exact GED computation remains infeasible for graphs with more than 16 nodes~\citep{astar_scale}. To address this, various approximation approaches have been proposed, including beam search~\citep{beam_ged2} and reductions to an assignment problem between the node sets~\citep{VJ}. For MCS, McSplit~\citep{mcsplit} and GLSearch~\citep{GLSearch} rely on branch-and-bound search whose worst-case time complexity remains exponential. All of these approaches operate on each graph pair individually, so their cost grows with both the graph size and the number of pairs within a dataset.
 
\subsection{GNN-Based Graph Similarity Learning}
 
A comprehensive overview of deep graph similarity learning is given by \citet{LiuSurvey}; here, we focus on GNN-based methods for GED and MCS prediction. Initial studies applying GNNs to graph similarity learning established two designs that later work builds on. SimGNN~\citep{SimGNN} employs graph-level embedding functions to summarize each graph and combines these representations with histogram features extracted from pairwise comparison to predict similarity scores. Instead of deriving graph-level representations, GraphSim~\citep{GraphSim} computes a node-node pairwise similarity matrix after each GCN~\citep{GCN} layer, then uses CNNs to capture fine-grained structural differences. These two models reflect a trade-off: node-level comparison captures finer differences between graphs, but it must be computed for every pair, and its cost scales with the product of the graph sizes. We therefore group subsequent work by how it treats cross-graph information.
 
\subsubsection{Node-Level Cross-Graph Matching}
 
A substantial amount of research has been conducted to strengthen graph similarity prediction through explicit node-level alignment between input graphs. A common design for this alignment is cross-graph attention, in which nodes in one graph attend to nodes in the other during encoding. GMN~\citep{GMN} introduces a cross-graph attention mechanism to compute node similarities before aggregating to graph-level similarity scores.
 
Another line of work utilizes node-level matching more directly for edit operations. GraphEdX~\citep{GraphEdX} formulates GED with general edit costs as a quadratic assignment problem and replaces each cost term with a neural set divergence computed under node and node-pair alignments. GEN~\citep{GEN} incorporates edit-operation costs into node-matching costs and models dependencies between matches within and across graphs, enabling globally guided alignment in a single step without iterative refinement.
 
These methods differ in how they align the two graphs, but they share a similar computational cost. Even when node embeddings are computed independently, the interaction or alignment step operates on all cross-graph node pairs; its cost therefore scales with the product of the graph sizes and must be repeated for every graph pair.
 
\subsubsection{Pair-Independent Graph-Level Embeddings}
 
Methods in this group encode each graph independently and compute similarity only from the two resulting graph-level embeddings. Because no information is exchanged between the two graphs, the embedding of a graph can be computed once and reused when it reappears. GraSP~\citep{GraSP} simplifies the overall architecture by removing node-level comparisons and enriching node representations with random-walk positional encodings.
 
Other models use cross-graph information during training but not at inference.
EGSC~\citep{EGSC} trains a teacher model that fuses the graph-level embeddings of both graphs, and then distills the resulting representation into a Siamese Graph Isomorphism Network (GIN)~\citep{GIN} student that has no cross-graph interaction. ERIC~\citep{ERIC} applies node-graph alignment as a regularization term during training and removes it at inference, so the similarity is computed from graph-level embeddings alone. It further combines a Neural Tensor Network (NTN)~\citep{NTN} and an $\ell_2$ distance across GNN layers into a multi-scale GED discriminator.
 
Other methods shape the embedding space itself so that its geometry reflects the target measure. GREED~\citep{GREED} adopts a Siamese GIN architecture and predicts GED and SED (subgraph edit distance) by using graph-level embeddings acquired by a series of neural networks. TaGSim~\citep{TAGSim} decomposes GED into counts of individual edit operation types and predicts graph edit vectors from type-aware graph-level embeddings. G2R~\citep{G2R} represents nodes as closed regions and summarizes each graph as a region whose shape reflects its structure and whose volume reflects its size. The overlap between two graph regions approximates the MCS, and their disjoint parts are used for GED.
 
\subsubsection{Edit-Path Recovery}
 
Methods in this group additionally predict a node matching from which an edit path is recovered. Since every recovered path is a feasible sequence of edit operations, its length is an upper bound on the true GED. Noah~\citep{Noah} and GENN-A*~\citep{GENN} use learned node matching to guide an A*-based search for the edit path. Other methods derive the edit path directly from a predicted matching matrix. GEDGNN~\citep{GEDGNN} predicts a node matching matrix and a cost matrix, and recovers an edit path through $k$-best matching. \citet{GEDIOT} propose three sets of algorithms: a supervised method based on a learnable Sinkhorn algorithm (GEDIOT), an unsupervised approach based on optimal transport and Gromov-Wasserstein discrepancy (GEDGW), and their ensemble method (GEDHOT).
 
Recovering an edit path requires a node correspondence between the two graphs, so all methods in this group compute at least an $n_1 \times n_2$ matching matrix for every pair, and those that add search or sampling repeat this cost over several candidates. However, these methods provide an interpretable edit path and a GED upper bound, which other methods do not.
 
\subsection{Ground-Truth Labels and Synthetic Pair Generation}
\label{sec:rw_labels}
 
Since exact GED computation is limited to small graphs, the commonly used benchmarks introduced with SimGNN~\citep{SimGNN} provide exact labels only for graphs with at most 10 nodes, while labels for larger graphs are obtained from approximate solvers. \citet{XuChang} observe that some of the provided IMDB dataset~\citep{imdb} labels are therefore not exact and can be either smaller or larger than the true GED, and regenerate them with an exact algorithm.
 
For graphs beyond the reach of exact solvers, several works generate pairs synthetically by applying a known number of edit operations $\Delta$ to a graph and using $\Delta$ as the label. GEDGNN~\citep{GEDGNN} and GEN~\citep{GEN} follow this strategy, the latter on graphs with 100 to 400 nodes. Since a shorter edit path between the two graphs may exist, $\Delta$ is in general only an upper bound on the true GED.
 
\subsection{Node Features and Embedding Comparison}
\label{sec:rel_ntn_film}
 
Message-passing GNNs are at most as expressive as the 1-WL test~\citep{GIN, MorrisWL}. Positional encodings (PEs) address this by adding structural information to the node features, such as Laplacian eigenvector encodings (LapPE)~\citep{DwivediBenchmark} and random-walk encodings (RWPE)~\citep{RWPE}. For graphs without node attributes, a cheaper alternative is the Local Degree Profile (LDP)~\citep{LDP}, which describes each node by its degree and summary statistics of neighborhood degrees and can be computed in time linear in the number of edges.
 
Once the two graphs are encoded, a head combines their embeddings into a similarity score. The most common choice is the NTN~\citep{NTN}, adopted by SimGNN, which models interactions between two graph-level embeddings via a bilinear tensor. It has since been used in many models, including TaGSim, ERIC, GEDGNN, GENN-A*, and GraSP. For $d$-dimensional embeddings and $k$ output channels, NTN requires $O(d^2k)$ parameters and computation per pair, and its bilinear form is not symmetric in the two graphs, whereas both GED and MCS are symmetric.
 
Feature-wise linear modulation (FiLM)~\citep{FiLM} conditions a network on external information by scaling and shifting each feature channel. GNN-FiLM~\citep{GNNFiLM} uses it within message passing, where each node modulates the messages it receives from its neighbors. Outside graph similarity, Count-GNN~\citep{CountGNN} predicts the number of subgraph isomorphisms with edge-centric message passing to capture fine-grained structural information. It modulates the input-graph representation conditioned on the query graph to improve matching. Since this modulation is applied to the edge representations of the input graph, the input-graph representation has to be recomputed for each query.
 
\subsection{Scale in Graph Similarity Learning}
 
Most graph similarity learning methods are evaluated on the benchmarks introduced with SimGNN and GraphSim, whose graphs are small (AIDS: avg. $\sim$9 nodes, LINUX: avg. $\sim$8 nodes, IMDB: avg. $\sim$13 nodes)~\citep{SimGNN, GraphSim}. There are a few exceptions, but they either remain in the range of hundreds of nodes or rely on search operations. GEDGNN and GEN are trained on synthetic pairs with up to 400 nodes, and INFMCS~\citep{INFMCS} is evaluated on graphs with a few hundred nodes. G2R is trained on graphs with at most 50 nodes and evaluated zero-shot on graphs with 1,000 to 2,000 nodes for MCS, but its error increases beyond a few hundred nodes. GLSearch handles pairs with up to roughly one million nodes, but it is a search procedure under a fixed budget per pair. FGWAlign~\citep{FGWAlign} computes upper bounds on GED for graphs with more than 10,000 nodes without learning. To the best of our knowledge, no learned GED or MCS predictor has been trained and evaluated on graph pairs with thousands of nodes.
 

%% file: chapter4.tex
\section{Methodology}

\begin{figure}[t]
    \centering
    \includegraphics[width=\textwidth]{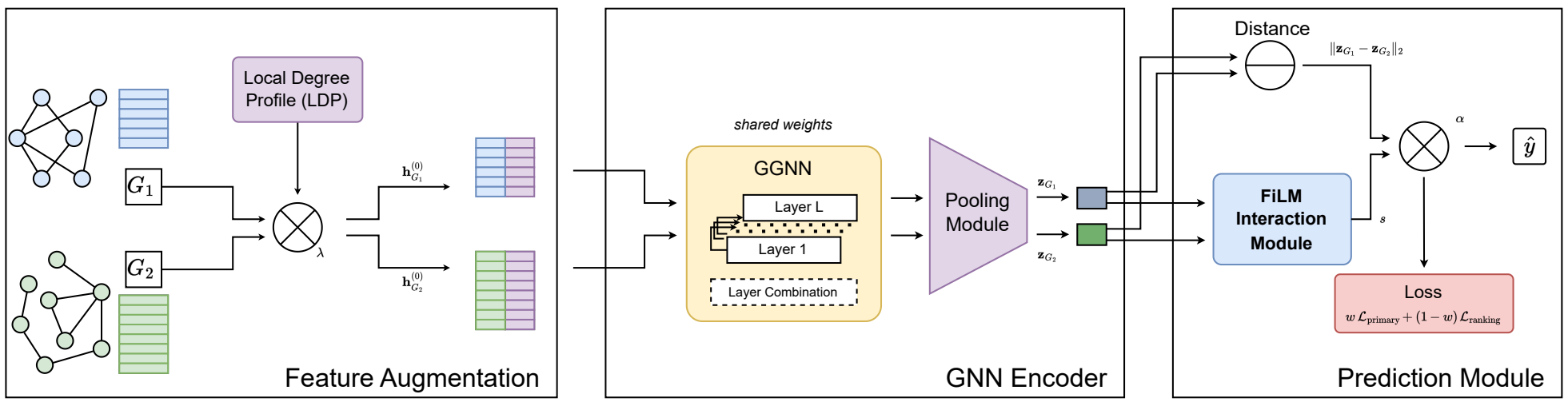}
    \caption{Overview of \method{}. Given two input graphs $G_1$ and $G_2$, the pipeline consists of three stages. First, the feature augmentation module combines the node features with the Local Degree Profile (LDP) into the initial node representations $\mathbf{h}^{(0)}$. Second, a shared GGNN encoder computes node representations, whose layer outputs are combined and pooled into fixed-size graph embeddings $\mathbf{z}_{G_1}$ and $\mathbf{z}_{G_2}$. Since each graph is encoded independently, these embeddings can be cached and reused across pairs. Finally, the prediction module combines a FiLM-based interaction score $s$ with the distance between the two embeddings to predict $\hat{y}$, and is trained on a weighted sum of a regression and a ranking loss.} 
    \label{architecture}
\end{figure}

In this section, we introduce our proposed framework for graph similarity learning. \method{} encodes each graph once, independently of the graph it is compared to, and scores pairs only from the resulting graph embeddings. We describe the node input features (\Cref{sec:features}), the encoder and pooling (\Cref{sec:encoder}), the similarity prediction module (\Cref{sec:prediction}), the training objective (\Cref{sec:loss}), the construction of our large-scale datasets (\Cref{sec:benchmark}), and the complexity of the framework (\Cref{sec:complexity}). \Cref{architecture} illustrates the overall architecture.

\subsection{Feature Augmentation}
\label{sec:features}

Although graph similarity metrics fundamentally rely on structural information, node features can provide important complementary information. Many real-world graphs, however, lack meaningful node features or contain only categorical labels that do not reflect topology. We therefore augment the node input of every graph with a structural descriptor that can be computed in time linear in the graph size.

For plain graphs, we create node features $\mathbf{x}_u$ of node $u \in V$ depending on the graph type. If the graph has node labels, $\mathbf{x}_u$ is the one-hot encoding of the node label. If there are no node labels, $\mathbf{x}_u$ is the one-hot encoding of degree $\deg(u)$. 

For all graphs, we add the LDP~\citep{LDP} $\mathbf{f}_u \in \mathbb{R}^5$ of each node $u$: the degree of $u$ and the minimum, maximum, mean, and standard deviation of its neighbors' degrees. Both inputs are computed once per graph in $O(n + m)$ time.

For high-degree and large graphs, we limit the size of the one-hot encoding and apply $\log(1+\cdot)$ to LDP to compress its range. 

We project the node features and LDP with learnable linear layers $\mathbf{W}_x$ and $\mathbf{W}_f$ and combine them either by a learned weighted sum,
\begin{equation*}
    \mathbf{h}_u^{(0)} = \lambda\, \mathbf{W}_x \mathbf{x}_u + (1 - \lambda)\, \mathbf{W}_f \mathbf{f}_u,
\end{equation*}
where $\lambda = \sigma(\theta_\lambda)$ is the sigmoid of a learnable parameter, or by concatenation, $\mathbf{h}_u^{(0)} = [\mathbf{W}_x \mathbf{x}_u \,\|\, \mathbf{W}_f \mathbf{f}_u]$, where each projection then has dimension $d/2$. The combination method is a hyperparameter that is selected per dataset.

\subsection{GNN Encoder and Graph-Level Pooling}
\label{sec:encoder}

Starting from the initial node representations $\mathbf{h}_u^{(0)}$, both input graphs are processed by a shared GGNN encoder in a Siamese~\citep{siamese} architecture. Both graphs of a pair pass through the same network with shared weights, which encodes each graph independently. This places both graphs in the same embedding space, which is required for comparing their embeddings directly. Since the encoder never sees the other graph of a pair, each graph is encoded once, and its embedding can be reused for every pair it appears in. Sharing the weights also halves the number of encoder parameters.

We investigate two variants of the GGNN encoder that differ in how the $L$ message-passing steps are performed:

\textbf{Single-Block GGNN.} This variant applies one GGNN block with $L$ message-passing steps. Each step has its own weight matrix $\mathbf{\Theta}^{(l)}$, while a single GRU is shared across all steps. A nonlinearity and dropout are applied once, after the last step:
\begin{equation*}
    \mathbf{H}^{(L)} = \mathrm{Dropout}\Big(\mathrm{ReLU}\big(\mathrm{GGNN}_L(\mathbf{H}^{(0)})\big)\Big),
\end{equation*}
where $\mathbf{H}^{(l)} \in \mathbb{R}^{n \times d}$ stacks the node representations $\mathbf{h}_u^{(l)}$. Sharing the GRU reduces the number of parameters compared to the stacked variant, while the per-step computation remains the same.

\textbf{Stacked GGNN.} This variant stacks $L$ single-step GGNN layers, each with its own weight matrix and GRU, and adds a residual connection~\citep{he2016deep} after every layer:
\begin{equation*}
    \mathbf{H}^{(l)} = \mathbf{H}^{(l-1)} + \mathrm{Dropout}\Big(\mathrm{ReLU}\big(\mathrm{GGNN}^{(l)}(\mathbf{H}^{(l-1)})\big)\Big), \quad l = 1, \dots, L.
\end{equation*}
The residual connections preserve the representation of the previous layer.

Before pooling, we combine the outputs of all layers (for the single-block variant, $\mathbf{h}_u^{(0)}$ and $\mathbf{h}_u^{(L)}$). We then obtain $\mathbf{z}_G$ by pooling the combined node representations.

\subsection{Similarity Prediction Module}
\label{sec:prediction}

The graph embeddings $\mathbf{z}_{G_1}$ and $\mathbf{z}_{G_2}$ obtained from pooling are passed to the prediction module, the only component of the framework that is evaluated per pair. It combines a learned interaction score with the distance between the two embeddings.

We condition each embedding on the other with FiLM~\citep{FiLM}:
\begin{equation}
\begin{aligned}
    \tilde{\mathbf{z}}_1 &= \big(1 + \boldsymbol{\gamma}(\mathbf{z}_{G_2})\big) \odot \mathbf{W} \mathbf{z}_{G_1} + \boldsymbol{\beta}(\mathbf{z}_{G_2}), \\
    \tilde{\mathbf{z}}_2 &= \big(1 + \boldsymbol{\gamma}(\mathbf{z}_{G_1})\big) \odot \mathbf{W} \mathbf{z}_{G_2} + \boldsymbol{\beta}(\mathbf{z}_{G_1}),
\end{aligned}
\label{eq:film}
\end{equation}
where $\boldsymbol{\gamma}$ and $\boldsymbol{\beta}$ are produced by a shared two-layer MLP, and $\mathbf{W}$ is a shared linear projection. The last layer of the MLP is initialized to zero, so the modulation is inactive at the start of training ($\boldsymbol{\gamma} = \boldsymbol{\beta} = \mathbf{0}$).

We then combine $\tilde{\mathbf{z}}_1$ and $\tilde{\mathbf{z}}_2$ through their sum and absolute difference and map the concatenation to a scalar score $s$:
\begin{equation}
    s = \mathrm{MLP}_{\text{score}}\Big(\mathrm{ReLU}\big(\mathrm{MLP}_{\text{out}}([\,\tilde{\mathbf{z}}_1 + \tilde{\mathbf{z}}_2 \,\|\, |\tilde{\mathbf{z}}_1 - \tilde{\mathbf{z}}_2|\,])\big)\Big).
    \label{eq:combine}
\end{equation}
Both operations are invariant to the order of the two graphs, and the FiLM parameters are shared between the two directions, so $s$ is symmetric in $G_1$ and $G_2$.

The final prediction combines $s$ with the Euclidean distance between the graph embeddings through a learnable weight $\alpha \in [0, 1]$:
\begin{align*}
    \hat{y}_{\text{GED}} &= -\alpha\, s + (1 - \alpha)\, \|\mathbf{z}_{G_1} - \mathbf{z}_{G_2}\|_2,\\
    \hat{y}_{\text{MCS}} &= \alpha\, s - (1 - \alpha)\, \|\mathbf{z}_{G_1} - \mathbf{z}_{G_2}\|_2.
\end{align*}
For GED, a larger value indicates less similar graphs, whereas for MCS a larger value indicates more similar graphs. Therefore, the signs of both terms are opposite between the two formulas. Intuitively, the distance term ties the prediction directly to the geometry of the embedding space, while the interaction score can model dependencies between the two graphs that a distance alone cannot express.

Compared to the NTN widely adopted in prior work (\Cref{sec:rel_ntn_film}), which learns a weight for every pair of embedding dimensions in each of its slices, the prediction module of \method{} operates on a projection of the embeddings and therefore has far fewer parameters.

\subsection{Training Objective}
\label{sec:loss}

We train the model end-to-end on raw GED and MCS values with a loss that combines a primary regression loss with a ranking loss. The primary loss penalizes the absolute error of each prediction, and the ranking term encourages the predictions to align with the targets across pairs. The total loss is $\mathcal{L} = w\, \mathcal{L}_{\text{primary}} + (1 - w)\, \mathcal{L}_{\text{ranking}}$ with a fixed weight $w \in (0, 1]$.

\subsection{Large-Scale Dataset Generation}
\label{sec:benchmark}

Existing graph similarity benchmarks consist of small graphs, which prevents a systematic evaluation of scalability to real-world applications where graphs may contain thousands of nodes. At the same time, computing exact GED or MCS for such graphs is prohibitively expensive, so labels cannot be obtained by direct computation. We therefore derive collections of graphs from real-world graphs through controlled edit sequences, for which an upper bound on the GED of every pair is known by construction.

A simpler construction would derive each graph independently from the source graph. However, at large scale, two such graphs share almost no edits, so the label of a pair is approximately the sum of the edit counts of the two graphs. In that case, the closest graphs to any query graph are simply the graphs with the fewest edits, regardless of the query graph. The task would be trivial, since even the ranking would not require comparing the two graphs. To avoid this, we generate graphs in a tree-like structure, in which each graph inherits the edits of its parent. Two graphs then share the edits of their common ancestors, so the label depends on the pair rather than on each graph separately.

We use the source graphs as plain graphs and discard their node features and class labels if they exist. Our label counts only node and edge edits, so features would not affect it. In addition, inserting nodes would require artificial features, and their use might introduce bias. The large-scale datasets therefore evaluate only structural similarity.

Starting from a real-world source graph $G_0$, we generate $M$ graphs one after another. The first graph is derived from $G_0$, and each subsequent graph is derived from a parent randomly chosen among the recently generated graphs. Each derivation step applies a random number of edits, proportional to the size of $G_0$, using four operations:

\begin{itemize}
    \item node deletion, together with its edges,
    \item edge deletion,
    \item edge insertion between two nodes that share a neighbor,
    \item node insertion, connecting a new node to an existing node and some of its neighbors.
\end{itemize}

\noindent A deleted node or edge is never added back. The source graph itself is not part of the dataset.

Each node keeps its identity throughout the generation, so it is known which nodes two graphs have in common. To transform $G_a$ into $G_b$, it is then sufficient to delete the nodes and edges that appear only in $G_a$ and to insert those that appear only in $G_b$. We use the number of these operations as the label, $y(G_a, G_b) = |V_a \triangle V_b| + |E_a \triangle E_b|$, where $\triangle$ denotes the symmetric difference. Since GED is the cost of the cheapest edit path, this label is an upper bound on the exact GED.

Unlike prior synthetic constructions (\Cref{sec:rw_labels}), which label only a graph and its edited copy, we label all pairs among the generated graphs. For the large-scale datasets, we evaluate only GED prediction, since the construction provides labels only for GED.

\subsection{Complexity Analysis}
\label{sec:complexity}

The cost of \method{} can be considered per graph and per pair. For a graph with $n$ nodes and $m$ edges, the input features are computed once in $O(n + m)$ time. Each of the $L$ GGNN steps aggregates messages over the edges in $O(md)$ and applies the weight matrix and the GRU to every node in $O(nd^2)$, so the encoder costs $O(L(md + nd^2))$. Pooling is linear in the number of nodes. The cost per graph is therefore linear in both $n$ and $m$.

Only the prediction module depends on the pair, and its cost does not depend on the size of the graphs. Graph embeddings can therefore be computed once and reused: comparing $Q$ query graphs against $R$ reference graphs requires $Q + R$ encoder passes and $Q \cdot R$ evaluations of the prediction module. We use this procedure in all evaluations.

Models that compare the two graphs at the node level typically introduce at least an $O(n_1 n_2 d)$ term per pair. Even when they encode each graph independently, this term has to be computed for every pair, so the comparison cannot be reduced or precomputed. We give the complexity comparison in \Cref{app:complexity_comp}.

%% file: chapter5_1.tex
\section{Experiments}
\label{sec:experiments}

In this section, we evaluate \method{}. We first describe the datasets, baseline methods, evaluation metrics, and implementation, and then present the results.

\subsection{Datasets}
\label{sec:datasets}

We evaluate our method on four widely used benchmark datasets for graph similarity learning and on large-scale datasets constructed from real-world graphs. \Cref{tab:datasets} summarizes the key statistics of the datasets.

\subsubsection{Benchmark Datasets}

Following GraphSim~\citep{GraphSim}, we use four benchmark datasets with their ground-truth GED and MCS values. For AIDS700nef and LINUX, GED values are computed exactly with A*~\citep{astar, astar_ged}. For IMDBMulti and PTC, where exact computation is expensive, the GED label is the minimum of three approximate methods (Beam~\citep{beam_ged2}, Hungarian~\citep{beam_ged}, and VJ~\citep{VJ}). MCS values are computed with McSplit~\citep{mcsplit}. We keep the original 80/20 training/test split for each dataset and randomly select one-fifth of the training graphs for validation. This split is fixed for all runs.

The datasets include \textbf{AIDS700nef}~\citep{aids_1, aids_linux} (molecular graphs from the AIDS antiviral screen), \textbf{LINUX}~\citep{aids_linux} (program dependence graphs), \textbf{IMDBMulti}~\citep{imdb} (actor co-appearance networks), and \textbf{PTC}~\citep{ptc} (chemical compound graphs). LINUX and IMDBMulti consist of plain graphs, for which we use one-hot node degrees as node features, whereas AIDS700nef and PTC have node labels, for which we use their one-hot encodings (\Cref{sec:features}).

\subsubsection{Large-Scale Datasets}

To systematically evaluate scalability, we use the large-scale datasets described in \Cref{sec:benchmark}. In the main text, we report results on four of them, chosen to cover different sizes and densities: \textbf{Cora}~\citep{planetoid_datasets} (small and sparse), \textbf{PubMed}~\citep{planetoid_datasets} (more nodes, still sparse), \textbf{Amazon Computers}~\citep{amazon_datasets} (fewer nodes than PubMed, but much denser), and \textbf{ogbn-arxiv}~\citep{ogb_datasets} (169K nodes and 1.16M edges). The remaining datasets are provided in \Cref{app:datasets}. For each real-world graph, we generate $M \in \{50, 200, 1000\}$ graphs, label each pair with an upper bound on its GED, and split the graphs similarly to benchmark datasets.

\begin{table}[t]
  \centering
  \caption{Statistics of the datasets. For the large-scale datasets, values are given for $M=200$. Other large-scale datasets and $M$ values are listed in \Cref{tab:app_datasets}.}
  \label{tab:datasets}
  \begin{tabular}{lrrrr}
    \toprule
    Dataset & \# Graphs & Avg. Nodes & Avg. Edges & Avg. Degree\\
    \midrule
    PTC & 344 & 25.56 & 25.96 & 2.03 \\
    AIDS700nef & 700 & 8.90 & 8.80 & 1.98 \\
    LINUX & 1000 & 7.58 & 6.93 & 1.83 \\
    IMDBMulti & 1500 & 13.00 & 65.94 & 10.14 \\
    \midrule
    Cora & 200 & 2,703 & 5,272 & 3.90 \\
    PubMed & 200 & 19,633 & 44,053 & 4.49 \\
    Amazon Computers & 200 & 13,717 & 244,029 & 35.58 \\
    ogbn-arxiv & 200 & 169,344 & 1,156,152 & 13.65 \\
    \bottomrule
  \end{tabular}
\end{table}

\subsection{Baseline Models}

We compare against the following graph similarity learning methods, grouped as in \Cref{sec:related}: methods with node-level cross-graph matching (\textbf{SimGNN}~\citep{SimGNN}, \textbf{GMN}~\citep{GMN}, \textbf{GraphEdX}~\citep{GraphEdX}, \textbf{GEN}~\citep{GEN}), methods that compare pair-independent graph-level embeddings (\textbf{EGSC}~\citep{EGSC}, \textbf{ERIC}~\citep{ERIC}, \textbf{GREED}~\citep{GREED}, \textbf{GraSP}~\citep{GraSP}, \textbf{G2R}~\citep{G2R}), and methods that recover edit paths (\textbf{GEDGNN}~\citep{GEDGNN}, \textbf{GEDIOT}~\citep{GEDIOT}).

All baselines use the same datasets, splits, and evaluation code as our method. Each baseline is run with its official implementation and its own training target. The implementation details are given in \Cref{app:baselines}.

\subsection{Evaluation Metrics}

All metrics are computed on raw GED and MCS values. Each test graph is compared with all reference graphs, i.e., the training and validation graphs. We report the Mean Absolute Error (MAE) over all test pairs and the Relative Absolute Error (RAE)~\citep{witten}, the ratio of the model's total absolute error to that of always predicting the mean training label, so that values below 1 indicate an improvement over this predictor. Additionally, we use Spearman's rank correlation coefficient ($\rho$)~\citep{spearman} and Kendall's rank correlation coefficient ($\tau$)~\citep{kendall} to assess how well the predicted ranking aligns with the ground-truth ranking. To complement them, we include Precision at 10 (P@10), the fraction of the ground-truth 10 most similar reference graphs that also appear among the predicted 10 most similar ones, where the most similar graphs are those with the smallest GED or the largest MCS. While $\rho$ and $\tau$ measure the quality of the whole ranking, P@10 focuses on the accuracy of top-ranked predictions. For efficiency, we report the inference time per 10,000 graph pairs, measured on the test pairs with each method's own implementation on the same experimental setup. Further details of the evaluation metrics are in \Cref{app:evaluation}.

\subsection{Implementation}

We implement \method{} in Python with PyTorch~\citep{pytorch} and PyTorch Geometric~\citep{pyg}, building on the implementation of GraSP~\citep{GraSP}. All experiments run on a single NVIDIA A100 GPU with 40 GB of memory and 64 GB of host memory. The code is available at \url{https://github.com/ahmetTuzen/FiLSim}.

Early stopping and hyperparameter selection use only the validation set. The test set is used once for the final evaluation. Further training details are given in \Cref{app:implementation}.

\subsection{Experimental Setup}

All methods, including \method{}, are trained under the same time limits: 2 hours of training (and validation, if the method uses it) on the benchmark datasets and 6 hours on the large-scale datasets. The same limits apply to inference. Runs that cannot complete a single pass over the training pairs within the time limit or that exhaust memory are reported as TO or OOM, respectively. To account for randomness, we repeat each experiment with 10 seeds on the benchmark datasets and 3 seeds on the large-scale datasets. The data split is fixed across seeds. Due to space limitations, the main tables report the mean over seeds, and the standard deviations are given in \Cref{app:experiments}. In the main text, we report MAE for the benchmark datasets and RAE for the large-scale datasets; the other measure, along with P@20, is listed in \Cref{app:experiments}. Baselines whose implementations report only GED are additionally trained with MCS labels, except for GEDGNN and GEDIOT, which require GED node matchings and are therefore not evaluated on MCS. Here, we report only the main findings; detailed explanations and discussions are left to the extended experiments in \Cref{app:experiments}.

%% file: chapter5_2.tex
\subsection{Results on Benchmark Datasets}
\label{sec:res_bench}

\Cref{tab:bench_ged,tab:bench_mcs} report the results on the four benchmark datasets for GED and MCS, respectively. Across both tasks, \method{} has the lowest error on all four datasets and the best ranking metrics on all datasets except LINUX.
 
\begin{table}[htb]
\footnotesize
\setlength{\tabcolsep}{3pt}
\centering
\caption{GED prediction results on the benchmark datasets, mean over 10 seeds. $\downarrow$: lower is better, $\uparrow$: higher is better. Time: inference time in seconds per 10,000 test pairs. Best result in \textbf{bold}, second best \underline{underlined}, and third best in \textit{italic}. \textbf{TO}: timeout.}
\label{tab:bench_ged}
\begin{tabular}{l*{10}{c}}
\toprule
\multirow{2}{*}{\textbf{Model}} & \multicolumn{5}{c}{\textbf{AIDS700nef}} & \multicolumn{5}{c}{\textbf{IMDBMulti}} \\
\cmidrule(lr){2-6}\cmidrule(lr){7-11}
& MAE$\downarrow$ & $\rho$$\uparrow$ & $\tau$$\uparrow$ & P@10$\uparrow$ & Time$\downarrow$ & MAE$\downarrow$ & $\rho$$\uparrow$ & $\tau$$\uparrow$ & P@10$\uparrow$ & Time$\downarrow$ \\
\midrule
SimGNN & 1.376 & 0.738 & 0.594 & 0.325 & 2.158 & 34.989 & 0.767 & 0.686 & 0.677 & 2.118 \\
GMN & 0.730 & 0.880 & 0.751 & 0.561 & 13.505 & 4.341 & 0.938 & \textit{0.870} & 0.829 & 14.933 \\
GraphEdX & \underline{0.542} & \underline{0.931} & \underline{0.823} & \underline{0.807} & 1.866 & \multicolumn{5}{c}{\textbf{TO}} \\
GEN & \textit{0.590} & \textit{0.925} & \textit{0.814} & \textit{0.773} & 0.098 & \textit{3.830} & \textit{0.939} & \textit{0.870} & 0.843 & 0.280 \\
\midrule
EGSC & 0.736 & 0.890 & 0.764 & 0.628 & 0.001 & 25.251 & 0.888 & 0.825 & 0.851 & $<$0.001 \\
ERIC & 0.665 & 0.905 & 0.785 & 0.712 & 0.001 & 35.344 & 0.797 & 0.745 & 0.849 & $<$0.001 \\
GREED & 0.695 & 0.901 & 0.779 & 0.675 & $<$0.001 & \underline{3.672} & 0.929 & 0.856 & 0.833 & $<$0.001 \\
GraSP & 0.618 & 0.923 & \textit{0.814} & 0.767 & 0.002 & 4.088 & \underline{0.945} & \underline{0.890} & \textit{0.859} & 0.001 \\
G2R & 0.656 & 0.912 & 0.795 & 0.746 & 0.001 & 31.392 & 0.805 & 0.740 & \underline{0.890} & 0.001 \\
\midrule
GEDGNN & 0.877 & 0.853 & 0.718 & 0.572 & 23.383 & 32.968 & 0.737 & 0.644 & 0.674 & 23.209 \\
GEDIOT & 0.729 & 0.901 & 0.779 & 0.721 & 34.152 & 18.502 & 0.824 & 0.742 & 0.757 & 35.416 \\
\midrule
\method{} & \textbf{0.532} & \textbf{0.937} & \textbf{0.834} & \textbf{0.830} & 0.005 & \textbf{2.235} & \textbf{0.959} & \textbf{0.916} & \textbf{0.905} & 0.003 \\
\midrule[\heavyrulewidth]
\multirow{2}{*}{\textbf{Model}} & \multicolumn{5}{c}{\textbf{LINUX}}& \multicolumn{5}{c}{\textbf{PTC}} \\
\cmidrule(lr){2-6}\cmidrule(lr){7-11}
& MAE$\downarrow$ & $\rho$$\uparrow$ & $\tau$$\uparrow$ & P@10$\uparrow$ & Time$\downarrow$ & MAE$\downarrow$ & $\rho$$\uparrow$ & $\tau$$\uparrow$ & P@10$\uparrow$ & Time$\downarrow$ \\
\midrule
SimGNN & 1.060 & 0.878 & 0.778 & 0.872 & 2.134 & 9.898 & 0.922 & 0.816 & 0.415 & 2.348 \\
GMN & 0.134 & 0.972 & 0.905 & 0.982 & 12.877 & \underline{3.039} & 0.933 & 0.830 & 0.475 & 14.494 \\
GraphEdX & 0.139 & 0.972 & 0.906 & \textit{0.991} & 0.837 & 9.512 & 0.848 & 0.724 & 0.481 & 751.344 \\
GEN & 0.089 & \textit{0.973} & 0.904 & \underline{0.992} & 0.078 & 3.793 & 0.937 & 0.841 & \textit{0.589} & 0.531 \\
\midrule
EGSC & 0.141 & \underline{0.974} & \underline{0.920} & 0.981 & $<$0.001 & 4.850 & 0.934 & 0.835 & 0.573 & 0.001 \\
ERIC & 0.078 & \underline{0.974} & \textit{0.916} & \textit{0.991} & 0.001 & 3.990 & 0.939 & 0.844 & 0.574 & 0.002 \\
GREED & 0.323 & 0.967 & 0.904 & 0.972 & $<$0.001 & \textit{3.635} & 0.919 & 0.810 & 0.437 & 0.002 \\
GraSP & \textit{0.059} & \textbf{0.975} & \textbf{0.922} & 0.990 & 0.001 & 3.672 & \textit{0.946} & \textit{0.853} & 0.581 & 0.004 \\
G2R & \underline{0.031} & \textit{0.973} & 0.898 & \textbf{0.995} & 0.001 & 3.857 & \underline{0.949} & \underline{0.856} & \underline{0.616} & 0.004 \\
\midrule
GEDGNN & 0.271 & 0.968 & 0.906 & 0.952 & 23.267 & 38.990 & 0.639 & 0.500 & 0.316 & 23.805 \\
GEDIOT & 0.228 & 0.972 & 0.912 & 0.967 & 34.533 & 38.133 & 0.558 & 0.508 & 0.350 & 38.201 \\
\midrule
\method{} & \textbf{0.020} & \underline{0.974} & 0.909 & \underline{0.992} & 0.003 & \textbf{2.436} & \textbf{0.958} & \textbf{0.872} & \textbf{0.642} & 0.007 \\
\bottomrule
\end{tabular}
\end{table}

For GED, \method{} is the best method in all accuracy metrics on AIDS700nef, IMDBMulti, and PTC, with the largest gains in MAE. On LINUX, it has the lowest error, while the ranking metrics of many methods are close to the ceiling. Within the training budget, methods with node-level interaction do not outperform \method{} on these datasets. The results for MCS follow a similar pattern: \method{} is the best method in all accuracy metrics on AIDS700nef, IMDBMulti, and PTC, and has the lowest error on LINUX.

\begin{table}[htb]
\footnotesize
\setlength{\tabcolsep}{3pt}
\centering
\caption{MCS prediction results on the benchmark datasets, mean over 10 seeds. $\downarrow$: lower is better, $\uparrow$: higher is better. Time: inference time in seconds per 10,000 test pairs. Best result in \textbf{bold}, second best \underline{underlined}, and third best in \textit{italic}. \textbf{TO}: timeout.}
\label{tab:bench_mcs}
\begin{tabular}{l*{10}{c}}
\toprule
\multirow{2}{*}{\textbf{Model}} & \multicolumn{5}{c}{\textbf{AIDS700nef}} & \multicolumn{5}{c}{\textbf{IMDBMulti}} \\
\cmidrule(lr){2-6}\cmidrule(lr){7-11}
& MAE$\downarrow$ & $\rho$$\uparrow$ & $\tau$$\uparrow$ & P@10$\uparrow$ & Time$\downarrow$ & MAE$\downarrow$ & $\rho$$\uparrow$ & $\tau$$\uparrow$ & P@10$\uparrow$ & Time$\downarrow$ \\
\midrule
SimGNN & 0.711 & 0.756 & 0.623 & 0.387 & 2.203 & 0.451 & 0.870 & 0.784 & 0.909 & 2.121 \\
GMN & 0.422 & 0.888 & 0.778 & 0.769 & 16.762 & 0.197 & \textit{0.881} & 0.798 & 0.937 & 15.248 \\
GraphEdX & 0.482 & 0.880 & 0.764 & 0.771 & 1.576 & \multicolumn{5}{c}{\textbf{TO}} \\
GEN & \underline{0.303} & \underline{0.933} & \underline{0.836} & \underline{0.874} & 0.098 & 0.295 & 0.868 & 0.772 & 0.879 & 0.277 \\
\midrule
EGSC & 0.436 & 0.892 & 0.785 & 0.740 & 0.001 & 0.233 & 0.880 & \textit{0.801} & 0.912 & $<$0.001 \\
ERIC & 0.339 & 0.907 & 0.806 & 0.801 & 0.001 & 0.361 & 0.862 & 0.774 & 0.868 & $<$0.001 \\
GREED & 0.472 & 0.881 & 0.769 & 0.761 & $<$0.001 & 0.289 & 0.876 & 0.787 & 0.899 & $<$0.001 \\
GraSP & \textit{0.331} & \textit{0.926} & \textit{0.829} & \textit{0.871} & 0.002 & \textit{0.141} & \underline{0.883} & \underline{0.804} & \textit{0.947} & 0.001 \\
G2R & \underline{0.303} & 0.920 & 0.820 & 0.823 & 0.001 & \underline{0.087} & 0.880 & 0.781 & \underline{0.959} & 0.001 \\
\midrule
\method{} & \textbf{0.264} & \textbf{0.935} & \textbf{0.839} & \textbf{0.889} & 0.003 & \textbf{0.081} & \textbf{0.885} & \textbf{0.808} & \textbf{0.964} & 0.002 \\
\midrule[\heavyrulewidth]
\multirow{2}{*}{\textbf{Model}} & \multicolumn{5}{c}{\textbf{LINUX}}& \multicolumn{5}{c}{\textbf{PTC}} \\
\cmidrule(lr){2-6}\cmidrule(lr){7-11}
& MAE$\downarrow$ & $\rho$$\uparrow$ & $\tau$$\uparrow$ & P@10$\uparrow$ & Time$\downarrow$ & MAE$\downarrow$ & $\rho$$\uparrow$ & $\tau$$\uparrow$ & P@10$\uparrow$ & Time$\downarrow$ \\
\midrule
SimGNN & 0.368 & 0.751 & 0.654 & 0.666 & 2.157 & 1.707 & 0.725 & 0.586 & 0.397 & 2.325 \\
GMN & 0.084 & \underline{0.813} & \textit{0.715} & 0.992 & 13.539 & 1.581 & 0.790 & 0.653 & 0.511 & 16.718 \\
GraphEdX & 0.105 & \underline{0.813} & \textit{0.715} & \textbf{0.998} & 0.912 & 5.339 & 0.516 & 0.398 & 0.415 & 751.328 \\
GEN & 0.048 & \underline{0.813} & 0.714 & 0.994 & 0.077 & 1.392 & 0.865 & 0.748 & \textit{0.634} & 0.673 \\
\midrule
EGSC & 0.081 & \textit{0.812} & \underline{0.719} & 0.980 & 0.001 & 1.482 & 0.813 & 0.687 & 0.499 & 0.001 \\
ERIC & 0.046 & \textit{0.812} & \textit{0.715} & 0.990 & 0.001 & 1.201 & 0.840 & 0.717 & 0.553 & 0.002 \\
GREED & 0.158 & \textit{0.812} & \underline{0.719} & 0.992 & 0.001 & 1.988 & 0.797 & 0.665 & 0.540 & 0.002 \\
GraSP & \textit{0.035} & \textbf{0.814} & \textbf{0.721} & \textit{0.995} & 0.001 & \underline{1.143} & \textit{0.868} & \textit{0.753} & \underline{0.636} & 0.004 \\
G2R & \underline{0.021} & \textit{0.812} & 0.701 & 0.994 & 0.001 & \textit{1.149} & \underline{0.884} & \underline{0.771} & 0.633 & 0.004 \\
\midrule
\method{} & \textbf{0.015} & \textbf{0.814} & \textit{0.715} & \underline{0.997} & 0.003 & \textbf{0.804} & \textbf{0.924} & \textbf{0.827} & \textbf{0.721} & 0.009 \\
\bottomrule
\end{tabular}
\end{table}

Although \method{} compares the two graphs only on their embeddings, it is the most accurate method on the benchmark datasets. While its scoring time is slightly longer than that of other graph-level methods, \method{} offers the best overall performance.

\subsection{Results on Large-Scale Datasets}
\label{sec:res_large}

\Cref{tab:large_main} reports the results on the large-scale datasets with $M = 200$. \method{} has the best ranking metrics on all four large-scale datasets and the lowest RAE on two of them; on Cora and PubMed, G2R has a slightly lower RAE. The gap to the best baseline is largest on Amazon Computers and ogbn-arxiv. On ogbn-arxiv, the other methods either run out of memory or fail to learn a useful similarity, and only \method{} remains accurate.
 
\begin{table}[htb]
\footnotesize
\setlength{\tabcolsep}{3pt}
\centering
\caption{GED prediction on the large-scale datasets with $M = 200$ graphs, mean over 3 seeds. RAE: relative absolute error. Time: inference time in seconds per 10,000 test pairs. Best result in \textbf{bold}, second best \underline{underlined}, and third best in \textit{italic}. \textbf{OOM}: out of memory; -: constant prediction (ranking metrics) or RAE above 1,000.}
\label{tab:large_main}
\begin{tabular}{l*{10}{c}}
\toprule
\multirow{2}{*}{\textbf{Model}} & \multicolumn{5}{c}{\textbf{Cora}} & \multicolumn{5}{c}{\textbf{PubMed}} \\
\cmidrule(lr){2-6}\cmidrule(lr){7-11}
& RAE$\downarrow$ & $\rho$$\uparrow$ & $\tau$$\uparrow$ & P@10$\uparrow$ & Time$\downarrow$ & RAE$\downarrow$ & $\rho$$\uparrow$ & $\tau$$\uparrow$ & P@10$\uparrow$ & Time$\downarrow$ \\
\midrule
SimGNN & 2.734 & - & - & 0.045 & 19.795 & 88.598 & -0.271 & -0.188 & 0.061 & 958.379 \\
GMN & 0.309 & 0.950 & 0.822 & 0.723 & 268.426 & \textit{0.564} & \textit{0.755} & \textit{0.582} & \textit{0.562} & 2511.166 \\
GraphEdX & \multicolumn{5}{c}{\textbf{OOM}} & \multicolumn{5}{c}{\textbf{OOM}}\\
GEN & 1.058 & 0.195 & 0.144 & 0.077 & 2879.634 & \multicolumn{5}{c}{\textbf{OOM}} \\
\midrule
EGSC & 2.734 & - & - & 0.045 & 0.191 & 2.591 & - & - & 0.085 & 0.653 \\
ERIC & 0.992 & 0.033 & 0.023 & 0.078 & 0.030 & 1.010 & -0.270 & -0.187 & 0.013 & 0.203 \\
GREED & \textit{0.224} & \textit{0.987} & \textit{0.914} & \textit{0.844} & 0.102 & 1.089 & 0.181 & 0.105 & 0.298 & 0.686 \\
GraSP & 0.301 & 0.955 & 0.835 & 0.711 & 0.267 & \multicolumn{5}{c}{\textbf{OOM}} \\
G2R & \textbf{0.188} & \underline{0.989} & \underline{0.925} & \underline{0.866} & 0.428 & \textbf{0.177} & \underline{0.984} & \underline{0.908} & \underline{0.833} & 3.134 \\
\midrule
GEDGNN & 2.734 & 0.162 & 0.112 & 0.176 & 204.246 & \multicolumn{5}{c}{\textbf{OOM}} \\
GEDIOT & 2.692 & 0.187 & 0.128 & 0.187 & 63.290 & \multicolumn{5}{c}{\textbf{OOM}} \\
\midrule
\method{} & \underline{0.190} & \textbf{0.991} & \textbf{0.939} & \textbf{0.913} & 0.285 & \underline{0.185} & \textbf{0.986} & \textbf{0.914} & \textbf{0.880} & 1.573 \\
\midrule[\heavyrulewidth]
\multirow{2}{*}{\textbf{Model}} & \multicolumn{5}{c}{\textbf{Amazon Computers}} & \multicolumn{5}{c}{\textbf{ogbn-arxiv}} \\
\cmidrule(lr){2-6}\cmidrule(lr){7-11}
& RAE$\downarrow$ & $\rho$$\uparrow$ & $\tau$$\uparrow$ & P@10$\uparrow$ & Time$\downarrow$ & RAE$\downarrow$ & $\rho$$\uparrow$ & $\tau$$\uparrow$ & P@10$\uparrow$ & Time$\downarrow$ \\
\midrule
SimGNN & 34.948 & -0.165 & -0.114 & 0.077 & 517.722 & \multicolumn{5}{c}{\textbf{OOM}} \\
GMN & \textit{0.715} & \textit{0.697} & \textit{0.538} & \textit{0.502} & 2060.480 & \multicolumn{5}{c}{\textbf{OOM}} \\
GraphEdX & \multicolumn{5}{c}{\textbf{OOM}} & \multicolumn{5}{c}{\textbf{OOM}} \\
GEN & \multicolumn{5}{c}{\textbf{OOM}} & \multicolumn{5}{c}{\textbf{OOM}} \\
\midrule
EGSC & 1.609 & -0.242 & -0.167 & 0.063 & 16.889 & \multicolumn{5}{c}{\textbf{OOM}} \\
ERIC & 1.070 & -0.150 & -0.104 & 0.056 & 15.390 & \multicolumn{5}{c}{\textbf{OOM}} \\
GREED & - & 0.189 & 0.151 & 0.181 & 1.637 & - & \textit{0.252} & \textit{0.187} & \textit{0.173} & 13.015 \\
GraSP & 2.768 & - & - & 0.085 & 6.484 & \multicolumn{5}{c}{\textbf{OOM}} \\
G2R & \underline{0.380} & \underline{0.908} & \underline{0.762} & \underline{0.652} & 3.926 & \underline{0.958} & \underline{0.289} & \underline{0.212} & \underline{0.183} & 34.922 \\
\midrule
GEDGNN & \multicolumn{5}{c}{\textbf{OOM}} & \multicolumn{5}{c}{\textbf{OOM}} \\
GEDIOT & 2.763 & 0.179 & 0.129 & 0.150 & 879.037 & \multicolumn{5}{c}{\textbf{OOM}} \\
\midrule
\method{} & \textbf{0.241} & \textbf{0.979} & \textbf{0.895} & \textbf{0.852} & 2.048 & \textbf{0.342} & \textbf{0.956} & \textbf{0.831} & \textbf{0.713} & 12.452 \\
\bottomrule
\end{tabular}
\end{table}

\method{} compares graphs only through their embeddings, so its encoding cost grows linearly with graph size (\Cref{sec:res_efficiency}). However, a graph-level design alone is not sufficient, as most graph-level baselines fail on the large-scale datasets. Our experiments therefore lead to two findings: (1) strong performance on benchmark datasets does not necessarily extend to the large-scale datasets, and (2) a linear cost does not guarantee accurate predictions on large graphs.

\subsection{Efficiency}
\label{sec:res_efficiency}

In addition to the time measurements in \Cref{sec:res_bench,sec:res_large}, we measure the training and inference cost of \method{}. At inference, each test graph is scored with all reference graphs. \method{} encodes each graph independently, so every graph is encoded once, and only its embedding is passed to the prediction module, which reduces the computation time substantially. To test the effect of this design, we also run the same model in pairwise mode, where both graphs are re-encoded for each pair, similar to models with cross-graph comparisons. 
 
\begin{table}[htb]
\footnotesize
\setlength{\tabcolsep}{3pt}
\centering
\caption{Training and scoring cost of \method{}. Time is the median, including validation time.}
\label{tab:res_efficiency}
\begin{tabular}{@{}lrrrrr@{}}
\toprule
 & \multicolumn{2}{c}{Training} & \multicolumn{3}{c}{Inference} \\
\cmidrule(lr){2-3}\cmidrule(lr){4-6}
Dataset & Time (min) & Peak GPU (GB) & Encode once (s) & Pairwise (s) & Peak GPU (GB) \\
\midrule
AIDS700nef & 9 & 0.18 & 0.04 & 0.6 & 0.89 \\
IMDBMulti & 22 & 0.29 & 0.10 & 4.3 & 0.79 \\
LINUX & 16 & 0.15 & 0.05 & 1.0 & 0.65 \\
PTC & 7 & 0.41 & 0.01 & 0.3 & 0.20 \\
\midrule
Cora & 120 & 8.72 & 0.18 & 11.6 & 1.75 \\
PubMed & 154 & 8.27 & 1.01 & 66.0 & 1.72 \\
Amazon C. & 165 & 8.87 & 1.31 & 85.0 & 4.02 \\
ogbn-arxiv & 306 & 10.50 & 7.97 & 509.7 & 1.66 \\
\bottomrule
\end{tabular}
\end{table}

\Cref{tab:res_efficiency} compares the two encoding approaches, along with the total time and memory required for both training and inference. The measured times match the expected ratio: for the large-scale datasets with $M = 200$, there are 40 test graphs and 160 reference graphs, so pairwise encoding requires $2 \times 40 \times 160$ encoder passes, whereas our approach requires $40 + 160$, which is $64\times$ fewer. Training is more expensive, but it fits on a single GPU for every dataset.

\begin{figure}[htb]
    \centering
    \includegraphics[width=\textwidth]{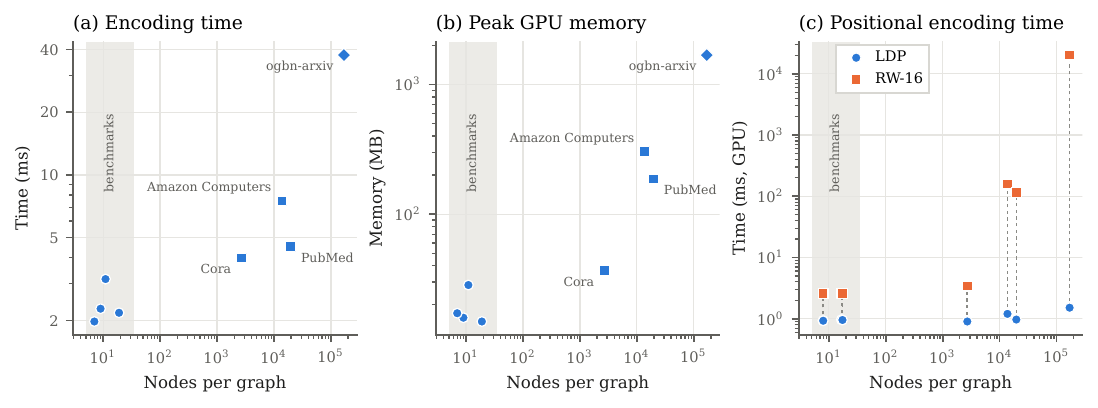}
    \caption{Cost of \method{} by graph size. (a) Time and (b) peak GPU memory to encode one graph (median size). (c) Time to compute structural encodings per graph: LDP and random-walk encodings with 16 steps. Shaded area: benchmark datasets.}
    \label{fig:scaling}
\end{figure}
 
\Cref{fig:scaling} illustrates the cost of encoding a single graph, measured on the median-sized graph of each dataset. On the benchmark graphs, encoding takes about 2 to 3 milliseconds regardless of size. For the large-scale datasets, both encoding time and memory usage increase with the total number of nodes and edges. LDP remains cheap on every dataset, since computing it requires only a single pass over the edges. On the other hand, random-walk encodings take longer as the graph size increases. Although random-walk encodings can be precomputed, their cost grows faster than the graph size, whereas the cost of LDP does not.

\begin{figure}[htb]
    \centering
    \includegraphics[width=\textwidth]{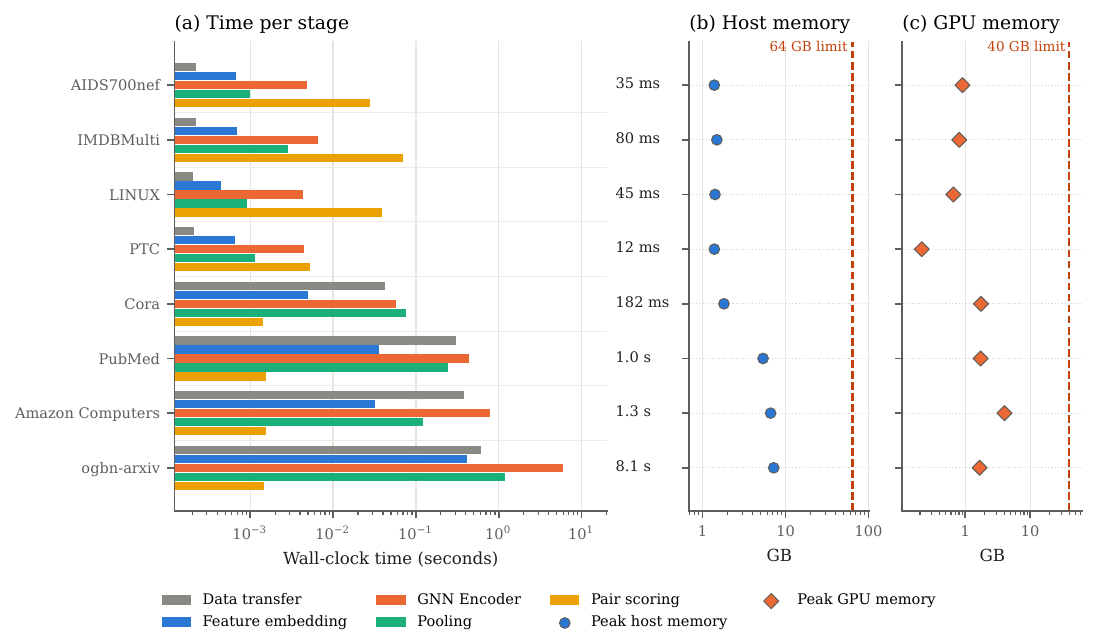}
    \caption{Stage breakdown of inference of \method. (a) Wall-clock time per stage and total time. (b) Peak host memory of the process and (c) peak GPU memory. Dashed lines: memory limits.}
    \label{fig:stages}
\end{figure}
 
The inference cost of \method{} is illustrated in \Cref{fig:stages}. For large-scale datasets, the cost is dominated by the GNN encoder, data transfer, and pooling, the order of which depends on the dataset. On the benchmark datasets, the order is different: pair scoring is the most expensive part, because the graphs are small and the number of pairs is high. Even on the largest datasets, memory usage of \method{} is far below the limits. Stages are timed separately, so their sum can differ slightly from the end-to-end scoring time in \Cref{tab:res_efficiency}.

\subsection{Ablation Study}
\label{sec:res_ablation}

To assess the contribution of each design choice, we remove or replace one component of \method{} at a time while keeping all other components unchanged. We first compare FiLM with NTN, the most widely adopted prediction module, across all datasets (\Cref{tab:ablation_prediction}). The remaining variants (\Cref{tab:ablation}) are evaluated only on AIDS700nef and Amazon Computers due to their computational cost. Ablations are grouped by the component of \method{} they target, with horizontal lines separating them.

\begin{table}[htb]
    \scriptsize
    \setlength{\tabcolsep}{5pt}
    \centering
    \caption{Comparison of the FiLM and NTN prediction modules. Better scores are in \textbf{bold}.}
    \label{tab:ablation_prediction}
  \begin{tabular}{llcc|cc|cc}
    \toprule
    \multirow{2}{*}{\textbf{Task}} & \multirow{2}{*}{\textbf{Dataset}} & \multicolumn{2}{c}{\textbf{RAE$\downarrow$}} & \multicolumn{2}{c}{\textbf{$\rho\uparrow$}} & \multicolumn{2}{c}{\textbf{P@10$\uparrow$}} \\
    \cmidrule(lr){3-4}\cmidrule(lr){5-6}\cmidrule(lr){7-8} 
 & & NTN & FiLM & NTN & FiLM & NTN & FiLM \\
    \midrule
    \multirow{8}{*}{GED} 
 & AIDS700nef& 0.293& \textbf{0.265} & 0.922& \textbf{0.937} & 0.738& \textbf{0.830} \\
  & IMDBMulti & 0.043& \textbf{0.027} & 0.948& \textbf{0.959} & 0.863& \textbf{0.905} \\
  & LINUX& 0.046& \textbf{0.009} & 0.973& \textbf{0.974} & 0.986& \textbf{0.992} \\
  & PTC  & 0.156& \textbf{0.102} & 0.943& \textbf{0.958} & 0.583& \textbf{0.642} \\
  & Cora & 0.199& \textbf{0.190} & 0.990& \textbf{0.991} & 0.900& \textbf{0.913} \\
  & PubMed & 0.201& \textbf{0.185} & 0.978& \textbf{0.986} & 0.812& \textbf{0.880} \\
  & Amazon C. & 0.347& \textbf{0.241} & 0.935& \textbf{0.979} & 0.360& \textbf{0.852} \\
  & ogbn-arxiv & \textbf{0.333} & 0.342 & \textbf{0.958}& 0.956 & 0.680 & \textbf{0.713} \\
    \midrule
    \multirow{4}{*}{MCS} 
 & AIDS700nef& 0.212& \textbf{0.198} & \textbf{0.935} & \textbf{0.935} & \textbf{0.897} & 0.889\\
 & IMDBMulti & 0.093& \textbf{0.047} & 0.881& \textbf{0.885} & 0.935& \textbf{0.964} \\
 & LINUX& 0.070& \textbf{0.019} & 0.812& \textbf{0.814} & 0.994& \textbf{0.997} \\
 & PTC  & 0.226& \textbf{0.223} & 0.912& \textbf{0.924} & 0.712& \textbf{0.721} \\
    \bottomrule
  \end{tabular}
\end{table}

The choice of the prediction module has a clear effect on the results. Among the 36 comparisons, NTN is better in only 3 cases, and the two modules are tied in 1 case, making FiLM the better choice for this task. Additionally, FiLM has fewer learnable parameters, making it an even stronger option between the two.

\begin{table}[htb]
\footnotesize
\setlength{\tabcolsep}{5pt}
\centering
\caption{Ablation of \method{} on AIDS700nef for GED and MCS (mean over 5 seeds; \method{} row: mean over 10 seeds), and on Amazon Computers ($M = 200$) for GED (mean over 3 seeds).}
\label{tab:ablation}
  \begin{tabular}{l|rrr|rrr|rrr}
  \toprule
    & \multicolumn{3}{c}{\textbf{AIDS700nef, GED}} & \multicolumn{3}{c}{\textbf{AIDS700nef, MCS}}& \multicolumn{3}{c}{\textbf{Amazon C., GED}} \\
    \cmidrule(lr){2-4}\cmidrule(lr){5-7}\cmidrule(lr){8-10} 
    & RAE$\downarrow$ & $\rho \uparrow$ & P@10$\uparrow $ & RAE$\downarrow$ & $\rho \uparrow$ & P@10$\uparrow $ & RAE$\downarrow$ & $\rho \uparrow$ & P@10$\uparrow$ \\
    \midrule
    \method & 0.265 & 0.937 & 0.830 & 0.198 & 0.935 & 0.889 & 0.241 & 0.979 & 0.852 \\
    \midrule
    Score only ($\alpha{=}1$) & 0.267 & 0.936 & 0.826 & 0.181 & 0.932 & 0.882& 0.230 & 0.979 & 0.857\\
    Distance only ($\alpha{=}0$) & 0.326 & 0.913 & 0.732 & 2.851 & 0.864 & 0.715& 2.757 & 0.613 & 0.530 \\
    \midrule
    w/o LDP & 0.271 & 0.935 & 0.818 & 0.195 & 0.935 & 0.884 & 0.270 & 0.961 & 0.793 \\
    RW-16 instead of LDP & 0.255 & 0.940 & 0.831 & 0.194 & 0.934 & 0.893& \multicolumn{3}{c}{OOM} \\
    w/o node features & 0.524 & 0.773 & 0.463 & 0.803 & 0.451 & 0.212& 0.341 & 0.945 & 0.773 \\
    \midrule
    Primary loss only & 0.268 & 0.937 & 0.829 & 0.201 & 0.934 & 0.884& 0.221 & 0.976 & 0.865 \\
    Ranking loss only & 0.328 & 0.921 & 0.744 & 1.991 & 0.910 & 0.820& 2.831 & 0.980 & 0.645 \\
    \midrule
    w/o message passing & 0.292 & 0.921 & 0.735 & 0.360 & 0.857 & 0.654& 0.290 & 0.978 & 0.845 \\
    w/o MP and LDP & 0.471 & 0.817 & 0.399 & 0.575 & 0.718 & 0.401& 0.260 & 0.965 & 0.802 \\
    \bottomrule
  \end{tabular}
\end{table}

The ablation results show that the contribution of each component depends on the dataset and the task. The FiLM-based score is the main contributor to the prediction: using only the distance degrades all metrics, whereas using only the score changes the results slightly and even lowers the RAE on AIDS700nef (MCS) and Amazon Computers. LDP is a cheap alternative to random-walk encodings, which perform similarly on AIDS700nef but run out of memory when computed for all graphs of Amazon Computers. The primary loss alone gives results comparable to \method{}, while the ranking loss alone leads to large RAE values on MCS and on Amazon Computers. In the without-message-passing variant, we remove all edges, so each node is updated only from its own input features. This degrades all metrics on AIDS700nef, whereas on Amazon Computers the ranking metrics remain close to those of \method{}. The FiLM-based score is the main contributor to the prediction: using only the distance degrades all metrics, whereas using only the score changes the results slightly and even makes it better in some cases. However, the FiLM-based score alone is not robust; on ogbn-arxiv, using only this score yields constant predictions, whereas combining it with the distance does not.

%% file: appendix.tex
\section{Graph Matching and Graph Edit Distance Examples}
\label{app:gm_ged_example}

\begin{figure}[htb]
    \centering
    \includegraphics[width=0.7\textwidth]{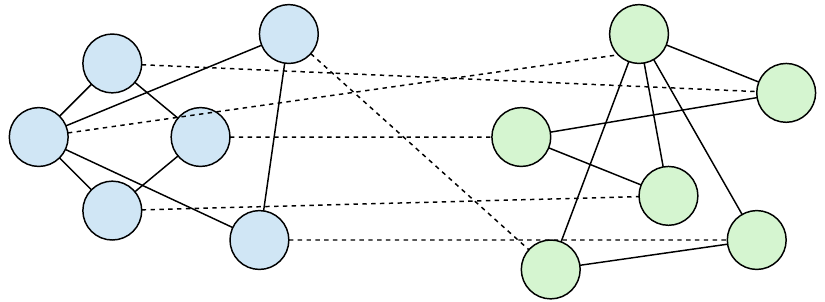}
    \caption{Graph matching example: two isomorphic graphs, with dashed lines showing corresponding nodes.}
    \label{fig:graph_matching}
\end{figure}

\begin{figure}[htb]
    \centering
    \includegraphics[width=0.85\textwidth]{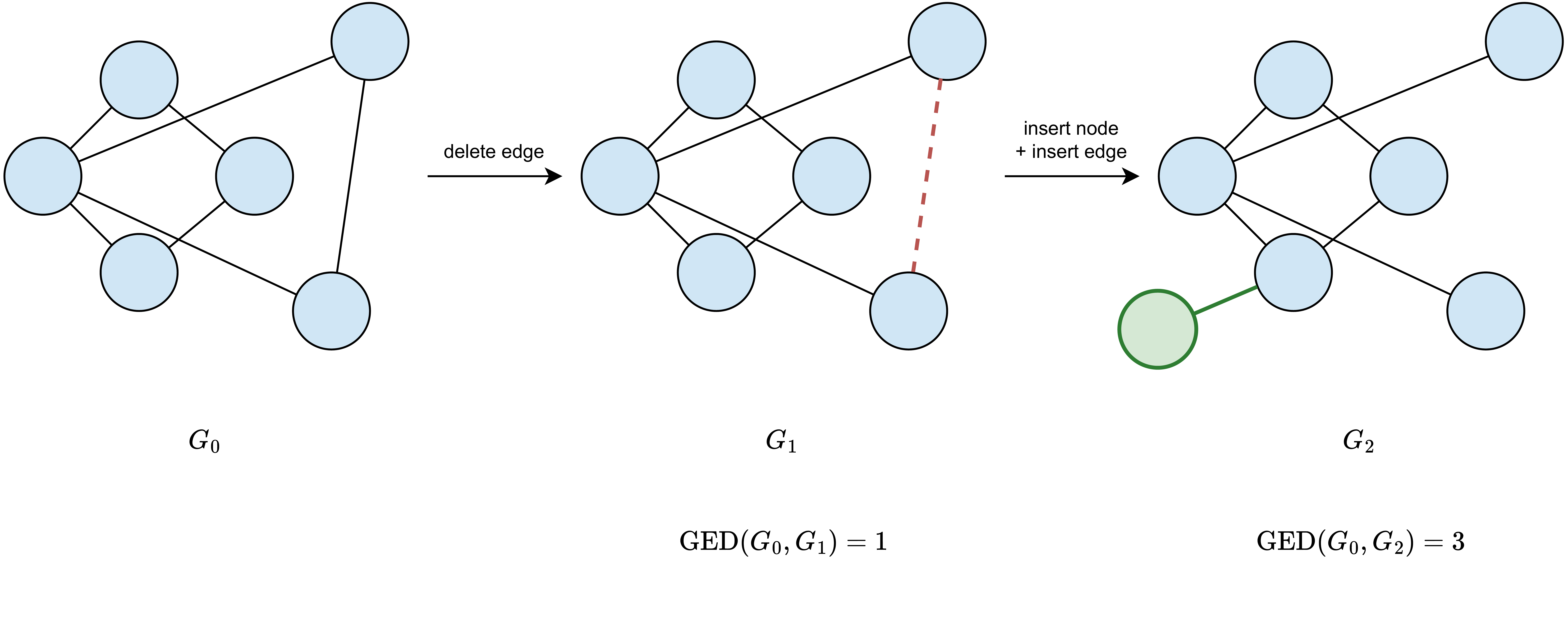}
    \caption{Example of graph edit distance. Each step transforms the previous graph by node or edge insertions or deletions; $\mathrm{GED}(G_0, G_k)$ is the minimum number of such operations between $G_0$ and $G_k$.}
    \label{fig:ged}
\end{figure}

Classical exact graph-matching approaches aim to find precise node-to-node correspondences between graph pairs, as shown in \Cref{fig:graph_matching}. The graph matching problem is a challenge across many fields, such as computer vision (visual tracking, action recognition), bioinformatics, and source/binary code analysis. Since the two graphs in \Cref{fig:graph_matching} are isomorphic, their GED is 0. A toy example of GED is illustrated in \Cref{fig:ged}.

\section{Complexity Comparison}
\label{app:complexity_comp}

\Cref{tab:complexity} compares the cost of \method{} with the baselines in our main experiments. We group the baselines as in \Cref{sec:related} and report the cost per graph and per pair. The original papers report their complexity in different forms; hence, we rewrite them in the form used in \Cref{sec:complexity}, where an $L$-layer message-passing encoder with MLP or GRU updates incurs a cost of $O(L(md + nd^2))$ per graph. The table compares cost classes rather than exact operation counts.

Except for the pair-independent group, the per-pair cost contains at least one term that grows with $n_1 n_2$, since these methods compute cross terms between all nodes of the two graphs. In GMN, these interactions are applied inside the encoder, so node representations depend on the pair and no part of the computation can be shared across pairs. The remaining methods in these groups encode each graph independently, but still compare the two graphs at the node level. SimGNN is discussed separately in \Cref{sec:related} as one of the first GNN-based models, but we include it in the node-level group.

Edit-path recovery methods additionally extract an edit path from the predicted node matching. We list the cost of this extra step in a separate column, since it is not required to predict the GED value. GEDGNN and GEDIOT extract $K$ candidate node matchings with $k$-best matching in $O(Kn^3)$, where $n = \max(n_1, n_2)$.

In the pair-independent group, the per-pair cost depends only on the embedding dimension. It ranges from $O(d_z)$ for GREED and G2R to $O(d_z^2 k)$ for the NTN used in GraSP. EGSC combines the two embeddings using an MLP at a cost of $O(d_z^2)$. The cost of the prediction module of \method{} lies between these two extremes: it is cheaper than an NTN but more expensive than a norm. None of these per-pair costs depends on the graph size.

\begin{table}[t]
\caption{Time complexity of \method{} and the baselines in our main experiments, grouped as in \Cref{sec:related}. $n, m$: nodes and edges of a graph; $n_1, n_2$: sizes of the two graphs in a pair; $d$: hidden dimension (also used in our prediction module); $d_z$: graph embedding dimension; $k$: output dimension of the NTN and of the interaction module of \method; $L$: number of layers; $K$: number of candidate matchings.}\label{tab:complexity}
\setlength{\tabcolsep}{2pt}
\footnotesize
\begin{tabularx}{\textwidth}{@{}>{\raggedright\arraybackslash}p{1.5cm}>{\raggedright\arraybackslash}X>{\raggedright\arraybackslash}Xll>{\raggedright\arraybackslash}p{1.7cm}@{}}
\toprule
\textbf{Group} & \textbf{Methods} & \textbf{Cross-graph operation} & \textbf{Per graph} & \textbf{Per pair} & \textbf{Edit path} \\
\midrule
Node-level\newline matching & SimGNN\footnotemark[1], \newline GMN\footnotemark[2],\newline GraphEdX\footnotemark[3], \newline GEN & histogram, attention, or alignment over node pairs & $O(L(md + nd^2))$\footnotemark[2] & $\geq O(n_1 n_2 d)$ & - \\
\midrule
Edit-path\newline recovery & GEDGNN\footnotemark[1], GEDIOT\footnotemark[1] & node matching matrix & $O(L(md + nd^2))$ & $\geq O(n_1 n_2 d)$ & $O(Kn^3)$ \\
\midrule
Pair-\newline independent & GREED,\newline G2R,\newline EGSC,\newline ERIC\footnotemark[4], \newline GraSP\footnotemark[1]\textsuperscript{,}\footnotemark[5] & none & $O(L(md + nd^2))$ & $O(d_z)$ to $O(d_z^2 k)$ & - \\
\midrule
\method{} & - & none & $O(L(md + nd^2))$ & $O(d_z d + d^2 + d k)$ & - \\
\bottomrule
\end{tabularx}
\footnotetext[1]{Uses an NTN, which adds $O(d_z^2 k)$ per pair.}
\footnotetext[2]{GMN applies cross-graph interactions inside the encoder; the encoder cost is paid per pair.}
\footnotetext[3]{The official implementation pads each graph to the largest graph size $n_{\max}$ in the dataset and aligns all node pairs of the two graphs, which costs $O(n_{\max}^4 d)$ time and $O(n_{\max}^4)$ memory per pair.}
\footnotetext[4]{Low-rank NTN, $O(d_z d' k)$ with $d' \ll d_z$.}
\footnotetext[5]{Additionally precomputes random-walk positional encodings, reported as $O(rn^2)$ for walk length $r$.}
\end{table}

\section{Datasets}
\label{app:datasets}

\subsection{Benchmark Datasets}

In addition to the details in \Cref{sec:datasets}, we provide further information here. \Cref{tab:app_bench} gives the number of graphs in each split and the average and maximum graph sizes in both the number of nodes and edges.

\begin{table}[t]
  \centering
  \caption{Benchmark dataset statistics.}
  \label{tab:app_bench}
  \begin{tabular}{lcccccc}
    \toprule
    Dataset & \# Graphs & Train / Val / Test & Avg. $n$ & Max. $n$ & Avg. $m$ & Max. $m$ \\
    \midrule
    PTC & 344 & 220 / 55 / 69 & 25.56 & 109 & 25.96 & 108 \\
    AIDS700nef & 700 & 448 / 112 / 140 & 8.90 & 10 & 8.80 & 14 \\
    LINUX & 1000 & 640 / 160 / 200 & 7.58 & 10 & 6.93 & 13 \\
    IMDBMulti & 1500 & 960 / 240 / 300 & 13.00 & 89 & 65.94 & 1467 \\
    \bottomrule
  \end{tabular}
\end{table}

\subsection{Large-Scale Datasets}
\label{app:large_datasets}

\Cref{tab:app_main_datasets} lists the main properties of the ten real-world graphs used to build the large-scale datasets. For each graph and each $M$, the generated graphs are split into 64\%, 16\%, and 20\% for training, validation, and testing. The number of graphs and pairs in each split is given in \Cref{tab:app_pairs}. Training pairs are sampled from all distinct pairs of training graphs, and each validation or test graph is compared with all training and validation graphs, excluding the graph itself. \Cref{tab:app_datasets} gives the statistics of the generated graph datasets.

\begin{table}[htb]
  \centering
  \caption{Main statistics of real-world graphs used to build the large-scale datasets.}
  \label{tab:app_main_datasets}
  \begin{tabular}{lccc}
    \toprule
    Graph & $n$ & $m$ & Avg. Degree \\
    \midrule
    Wiki~\citep{attributed_datasets} & 2,405 & 11,596 & 9.64 \\
    Cora~\citep{planetoid_datasets} & 2,708 & 5,278 & 3.90 \\
    CiteSeer~\citep{planetoid_datasets} & 3,327 & 4,552 & 2.74 \\
    BlogCatalog~\citep{attributed_datasets} & 5,196 & 171,743 & 66.11 \\
    Flickr~\citep{attributed_datasets} & 7,575 & 239,738 & 63.30 \\
    Amazon Computers~\citep{amazon_datasets} & 13,752 & 245,861 & 35.76 \\
    PubMed~\citep{planetoid_datasets} & 19,717 & 44,324 & 4.50 \\
    \midrule
    ogbn-arxiv~\citep{ogb_datasets} & 169,343 & 1,157,799 & 13.67 \\
    com-DBLP~\citep{snap_datasets} & 317,080 & 1,049,866 & 6.62 \\
    com-Amazon~\citep{snap_datasets} & 334,863 & 925,872 & 5.53 \\
    \bottomrule
  \end{tabular}
\end{table}

\begin{table}[htb]
  \centering
    \caption{Number of graphs and pairs per split for the large-scale datasets. The counts are the same for all ten graphs.}    
  \label{tab:app_pairs}
  \begin{tabular}{crrrrr}
    \toprule
    $M$ & Train / Val / Test & Train pairs & Val pairs & Test pairs \\
    \midrule
    50 & 32 / 8 / 10 & 496 & 312 & 400 \\
    200 & 128 / 32 / 40 & 8,128 & 5,088 & 6,400 \\
    1000 & 640 / 160 / 200 & 204,480 & 127,840 & 160,000 \\
    \bottomrule
  \end{tabular}
\end{table}

\begin{table}[htb]
  \centering
  \caption{Statistics of the generated graphs. GED label: min / median / max of the GED upper bound used as the label, over all pairs.}
  \label{tab:app_datasets}
  \begin{tabular}{lcrrr}
    \toprule
    Graph & $M$ & Avg. $n$ & Avg. $m$ & GED label \\
    \midrule
    \multirow{3}{*}{Wiki} & 50 & 2,403 & 11,582 & 18 / 301 / 736 \\
     & 200 & 2,403 & 11,593 & 17 / 524 / 1,310 \\
     & 1000 & 2,398 & 11,554 & 14 / 781 / 1,825 \\
    \midrule
    \multirow{3}{*}{Cora} & 50 & 2,703 & 5,267 & 14 / 199 / 467 \\
     & 200 & 2,703 & 5,272 & 15 / 369 / 866 \\
     & 1000 & 2,703 & 5,259 & 11 / 472 / 1,013 \\
    \midrule
    \multirow{3}{*}{CiteSeer} & 50 & 3,327 & 4,557 & 16 / 203 / 492 \\
     & 200 & 3,327 & 4,560 & 13 / 349 / 840 \\
     & 1000 & 3,324 & 4,531 & 11 / 498 / 1,269 \\
    \midrule
    \multirow{3}{*}{BlogCatalog} & 50 & 5,186 & 171,302 & 299 / 4,648 / 11,949 \\
     & 200 & 5,186 & 171,227 & 221 / 7,570 / 18,114 \\
     & 1000 & 5,190 & 171,495 & 203 / 10,350 / 20,810 \\
    \midrule
    \multirow{3}{*}{Flickr} & 50 & 7,575 & 239,786 & 192 / 4,703 / 11,428 \\
     & 200 & 7,568 & 239,904 & 225 / 8,079 / 19,403 \\
     & 1000 & 7,570 & 239,032 & 174 / 12,638 / 28,778 \\
    \midrule
    \multirow{3}{*}{Amazon C.} & 50 & 13,737 & 244,742 & 381 / 4,816 / 10,920 \\
     & 200 & 13,717 & 244,029 & 368 / 11,118 / 26,696 \\
     & 1000 & 13,721 & 244,284 & 312 / 14,461 / 30,219 \\
    \midrule
    \multirow{3}{*}{PubMed} & 50 & 19,685 & 44,275 & 119 / 1,601 / 4,255 \\
     & 200 & 19,633 & 44,053 & 98 / 2,967 / 6,738 \\
     & 1000 & 19,639 & 43,939 & 102 / 4,118 / 8,698 \\
    \midrule
    \multirow{3}{*}{ogbn-arxiv} & 50 & 169,212 & 1,156,984 & 2,355 / 28,868 / 69,312 \\
     & 200 & 169,344 & 1,156,152 & 1,943 / 51,288 / 126,624 \\
     & 1000 & 169,186 & 1,155,069 & 1,895 / 67,648 / 148,871 \\
    \midrule
    \multirow{3}{*}{com-DBLP} & 50 & 316,587 & 1,047,236 & 2,809 / 33,969 / 79,419 \\
     & 200 & 316,810 & 1,051,380 & 2,514 / 54,375 / 133,354 \\
     & 1000 & 316,439 & 1,044,017 & 2,313 / 70,367 / 166,320 \\
    \midrule
    \multirow{3}{*}{com-Amazon} & 50 & 334,856 & 928,747 & 2,887 / 30,334 / 74,171 \\
     & 200 & 334,415 & 921,061 & 2,653 / 55,931 / 125,546 \\
     & 1000 & 334,323 & 925,118 & 2,605 / 75,092 / 162,620 \\
    \bottomrule
  \end{tabular}
\end{table}

\section{Implementation Details}
\label{app:implementation}

\noindent\textbf{Software.} 
The package versions and installation instructions are given in the repository. To ensure reproducibility, each run is seeded. All experiments are run on a single NVIDIA A100 GPU with 40 GB of memory and 64 GB of host memory.

\noindent\textbf{Training.} 
We train in steps, where each step uses one batch of training pairs. In each pass over the training set, each graph appears once as the first and once as the second graph of a pair. We use the Adam optimizer with weight decay and gradient clipping along with a validation-based learning-rate scheduler that reduces the learning rate if there is no improvement on the validation set. 

We define the validation objective for \method{} as 
\[
  \rho + \tau + 0.5\,\mathrm{P@10} + 0.5\,\mathrm{P@20} - 0.5\,\mathrm{RAE}.
\]
Every 250 steps, we evaluate on the validation set and keep the best checkpoint based on the validation objective. Training stops when the validation objective does not improve or when the time limit is reached. The weights $\alpha$ and $\lambda$ are initialized to 0.5. The one-hot degree is capped at 128 (only meaningful for large graphs) and $\log(1+\cdot)$ is applied to LDP. For hyperparameter optimization, we adopt Optuna~\citep{optuna}.

\noindent\textbf{Benchmark datasets.} We run up to 250 trials per dataset. The three best configurations are retrained with 10 seeds, and we report the one with the best mean validation objective. The batch size is fixed to 128.

\noindent\textbf{Large-scale datasets.} Since training on large datasets is more expensive than on the benchmarks, we limit the hyperparameter search. The GNN encoder is fixed to the Stacked GGNN ($d = 128$, $L = 4$. We search only for the other hyperparameters at $M = 200$ within a 24-hour wall-clock time limit, resulting in 64 to 82 completed trials for the seven smaller graphs and 38 to 49 trials for the three largest graphs (ogbn-arxiv, com-DBLP, and com-Amazon). As with the benchmark datasets, we retrain the configuration with the best validation objective, but with 3 seeds instead of 10.

\section{Hyperparameters}
\label{app:hyperparameters}

We select the hyperparameters of \method{} for each dataset and task separately, with Optuna and the validation objective defined in \Cref{app:implementation}. On all datasets, each optimization trial trains for at most 20,000 steps, and the validation objective is measured every 250 steps. In all searches, a trial stops early when the validation objective does not improve for 10 to 40 validations, depending on the dataset size. We then retrain the best configurations with 10 seeds on the benchmark datasets and 3 seeds on the large-scale datasets. The detailed hyperparameter space can be found in our repository.

Some settings are fixed rather than searched: training is measured in optimization steps, not epochs. Training pairs are sampled at random from the training graphs in every pass. The model with the best validation objective is evaluated on the test set. We use the Adam optimizer with a plateau-based learning-rate scheduler that reduces the learning rate when the validation objective no longer improves.

\section{Evaluation Metrics}
\label{app:evaluation}

All metrics are computed on raw GED or MCS values. Let $\mathcal{Q}$ be the set of query graphs (test or validation graphs) and $\mathcal{R}$ the reference set, which contains all training and validation graphs, with $Q = |\mathcal{Q}|$ and $R = |\mathcal{R}|$ as in \Cref{sec:complexity}. For a query graph $q$, we compare it with every reference graph $r \in \mathcal{R}$ (excluding $r = q$ during validation). We write $y_{qr}$ for the label and $\hat{y}_{qr}$ for the prediction, and $\mathcal{P}$ for the set of all evaluated pairs.

\noindent\textbf{Error.} MAE and RAE are computed over all evaluated pairs:
\begin{align*}
  \mathrm{MAE} &= \frac{1}{|\mathcal{P}|} \sum_{(q,r) \in \mathcal{P}} |\hat{y}_{qr} - y_{qr}|, \\
  \mathrm{RAE} &= \frac{\sum_{(q,r) \in \mathcal{P}} |\hat{y}_{qr} - y_{qr}|}{\sum_{(q,r) \in \mathcal{P}} |y_{qr} - \bar{y}|},
\end{align*}

where $\bar{y}$ is the mean label over all pairs of training graphs. RAE does not depend on the scale of the labels; therefore, it is more suitable than MAE for comparing results across graph sizes.

\noindent\textbf{Ranking.} For each $q$, we rank the reference graphs by their labels and by the predictions, and compute Spearman's $\rho$ and Kendall's $\tau$ using SciPy~\citep{scipy}. We report the mean over query graphs.

\noindent\textbf{Precision at $k$.} For each $q$, the true top-$k$ set contains the $k$ most similar reference graphs, i.e., those with the smallest GED or the largest MCS. Reference graphs may share the same label, and we include all graphs tied with the $k$-th graph in the true set. Hence, the true set can contain more than $k$ graphs. P@$k$ is the number of graphs among the $k$ most similar ones according to the predictions that are in the true set, divided by $k$. We report precision at $k = 10$ and $k = 20$ as the mean over all query graphs.

\noindent\textbf{Comparison across $M$.} On the large-scale datasets, the labels grow with $M$ as shown in \Cref{tab:app_datasets}, and the reference set contains more graphs for larger $M$ as shown in \Cref{tab:app_pairs}. MAE and P@$k$ therefore cannot be compared directly across different values of $M$, while RAE, $\rho$, and $\tau$ are better suited for this comparison. 

\section{Baselines}
\label{app:baselines}

For all baselines, we use their public repositories. Each method is trained on the same split as \method{}, with its own input features and target. Their predictions are converted back to raw GED or MCS values, and all metrics are computed with our evaluation code. The training budget is 2 hours on the benchmark datasets and 6 hours on the large-scale datasets, and runs that cannot complete a single pass over the training pairs within the budget are reported as TO. When a run exceeds the memory limit, we repeatedly halve the batch size, down to 4 on the benchmark datasets and 1 on the large-scale datasets. We use 10 seeds on the benchmark datasets and 3 on the large-scale datasets. For graph-level methods, the reported time results are derived from cached graph embeddings. Below, we list all deviations from the published setups and the design choices we make for each baseline.

\begin{itemize}
    \item \textbf{SimGNN:} We enable the histogram features and increase the number of epochs for the large graphs.
    \item \textbf{GMN:} The repository trains with a margin loss, which does not produce GED values. Instead, we regress the distance between the two graph vectors onto the target using MSE, as is also done in other papers~\citep{GraphEdX}.
    \item \textbf{GraphEdX:} We use $\lambda = 1$, which matches the unit-cost GED.
    \item \textbf{GEN:} We run it as published.
    \item \textbf{GraSP:} The repository evaluates on the validation set only in the last 5\% of the epochs; we change it to match the budget.
    \item \textbf{GREED:} We run as published.
    \item \textbf{ERIC:} The validation window follows the budget, as in the GraSP modification.
    \item \textbf{EGSC:} The teacher and the student share one budget, divided in the ratio of their epoch numbers in the repository. Only the student is used at inference.
    \item \textbf{G2R:} We run as published.
    \item \textbf{GEDGNN} and \textbf{GEDIOT:} We report the predicted GED value and do not run the $k$-best edit-path search, which is applied after training and is extremely slow for all test pairs. As ground-truth matchings, we use the TaGED matchings of the repository. Both are trained on the GPU, like the other methods. Since training needs GED node matchings, they are not evaluated on MCS.
\end{itemize}

\section{Extended Experiments}
\label{app:experiments}

\input{appendix_experiments}

\subsection{Case Studies}
\label{app:case}

\begin{figure}[htb]
    \centering
    \includegraphics[width=\textwidth]{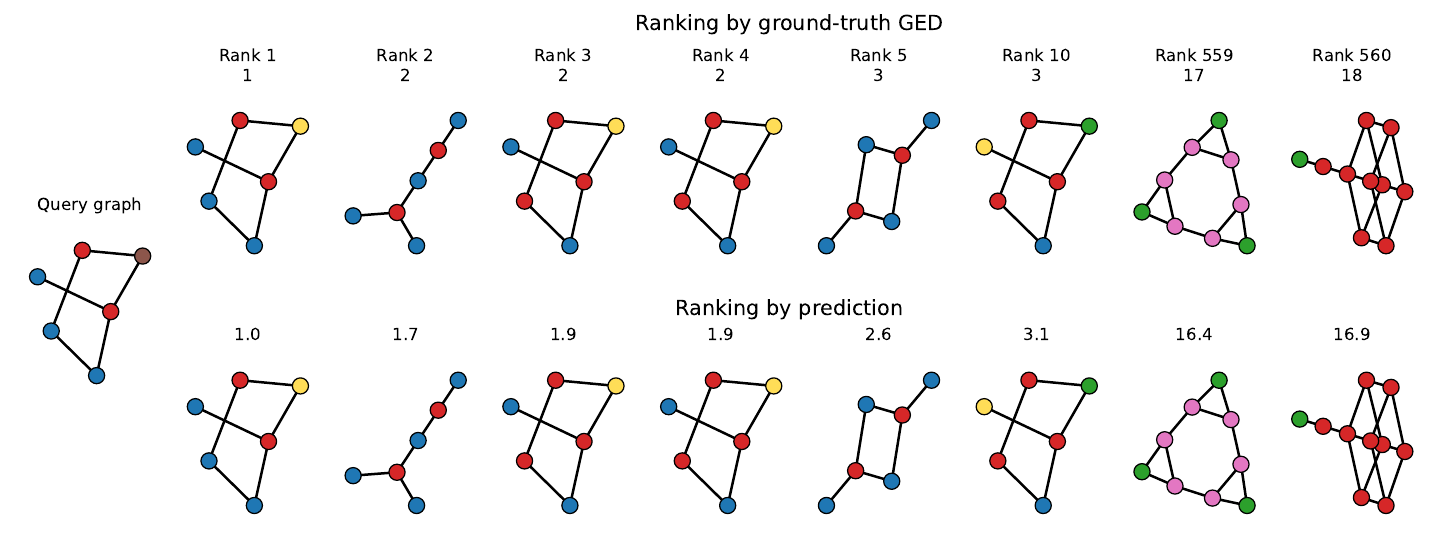}\\
    \includegraphics[width=\textwidth]{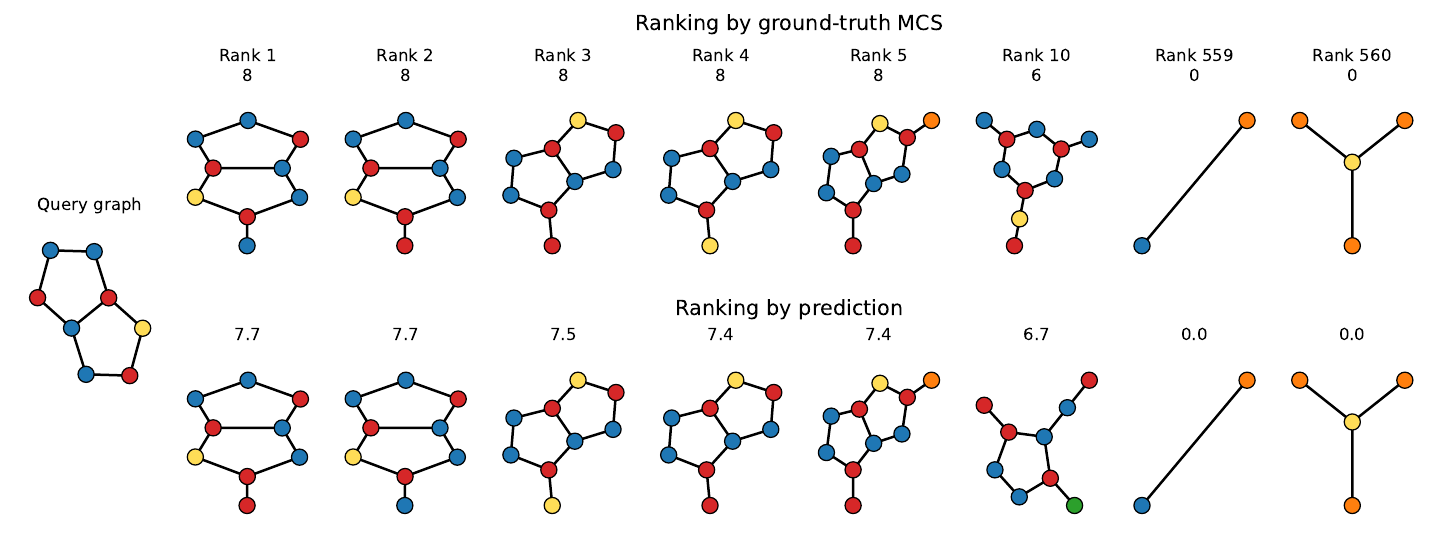}
    \caption{Example queries from the AIDS700nef test set, GED (top) and MCS (bottom).}
    \label{fig:case_aids}
\end{figure}

\begin{figure}[htb]
    \centering
    \includegraphics[width=\textwidth]{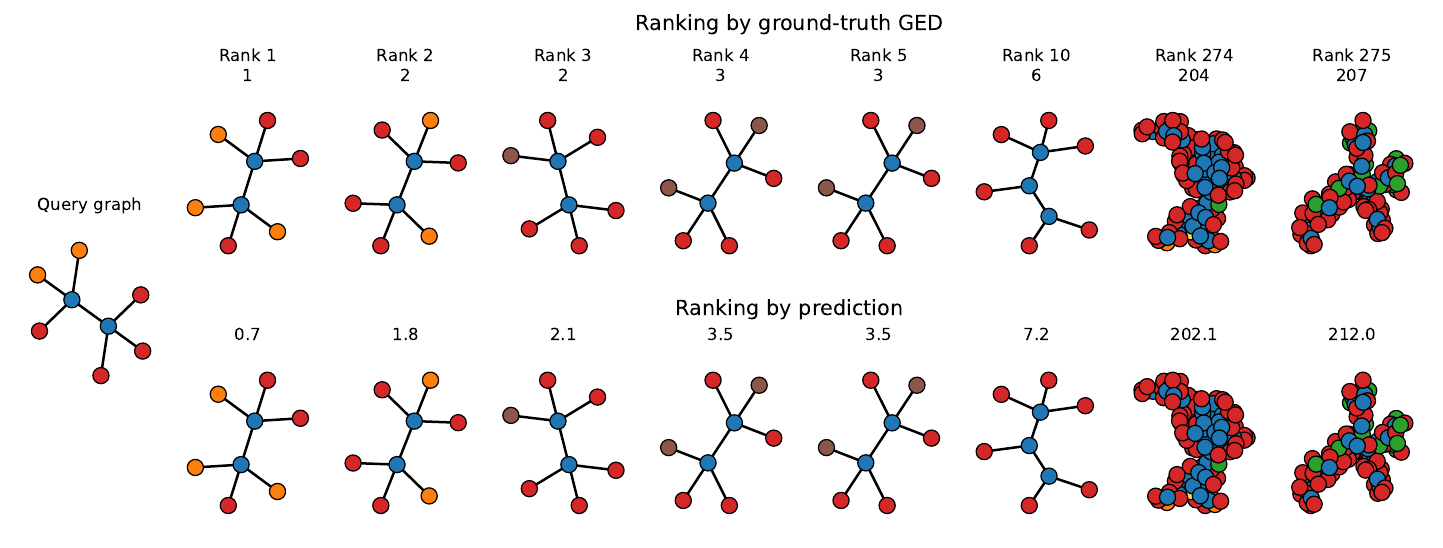}\\
    \includegraphics[width=\textwidth]{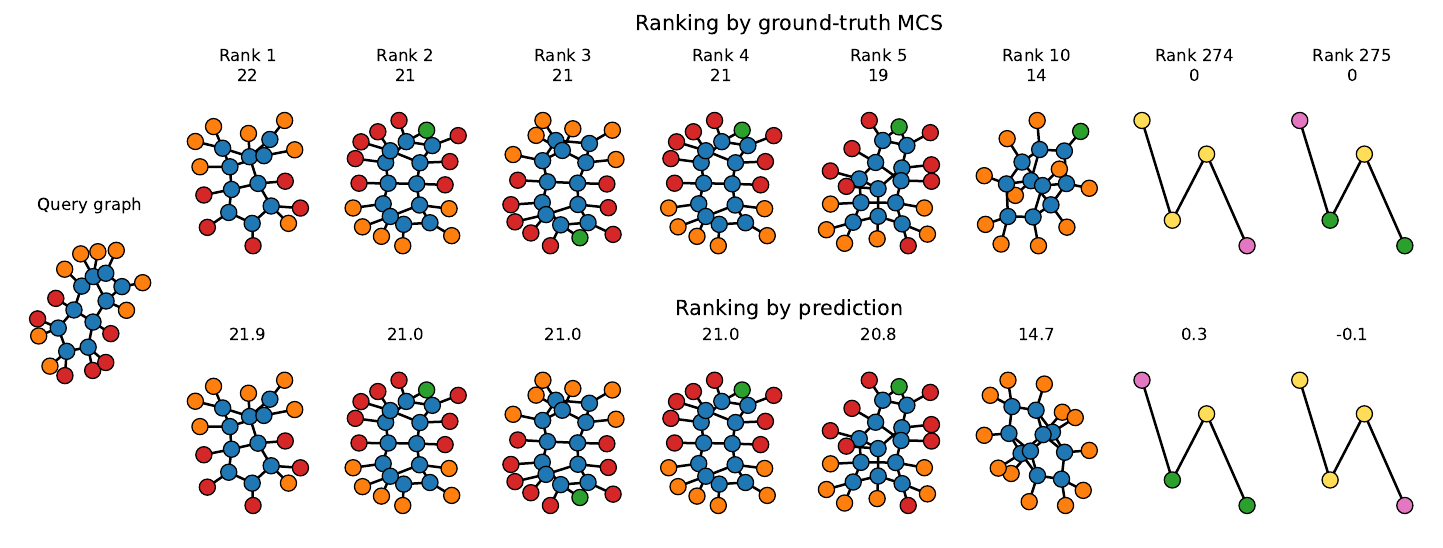}
    \caption{Example queries from the PTC test set, GED (top) and MCS (bottom).}
    \label{fig:case_ptc}
\end{figure}

\Cref{fig:case_aids,fig:case_ptc} show example queries from the AIDS700nef and PTC test sets for GED and MCS. In each example, the top row shows the reference graphs ranked by the ground-truth value, and the bottom row shows the reference graphs ranked by the prediction of \method{}. The main differences from the ground truth lie within the groups of tied graphs, since several reference graphs share the same GED or MCS value, and \method{} orders them by small differences in its continuous predictions.

%% file: appendix_experiments.tex
This section reports the complete results behind the tables and figures of \Cref{sec:res_bench,sec:res_large,sec:res_efficiency}: all metrics with standard deviations on the benchmark datasets, all large-scale datasets and collection sizes,  the training cost (\Cref{app:cost}), and the case study (\Cref{app:case}).

\subsection{Benchmark Datasets}
\label{app:bench_all}

\Cref{tab:app_bench_ged,tab:app_bench_mcs} extend \Cref{tab:bench_ged,tab:bench_mcs} with the relative absolute error (RAE) and P@20, and give the standard deviation over the same 10 seeds.

\begin{sidewaystable}[htbp]
\footnotesize
\renewcommand{\arraystretch}{0.8}
\centering
\caption{GED prediction on the benchmark datasets, all metrics. Mean $\pm$ standard deviation over 10 seeds. MAE: mean absolute error, RAE: relative absolute error, TO: timeout. Best result in \textbf{bold}, second best \underline{underlined}, and third best in \textit{italic}.}
\label{tab:app_bench_ged}
\begin{tabular}{@{}llcccccc@{}}
\toprule
Dataset & Model & MAE & RAE & $\rho$ & $\tau$ & P@10 & P@20 \\
\midrule
AIDS700nef & SimGNN & 1.376{\scriptsize$\pm$0.114} & 0.686{\scriptsize$\pm$0.057} & 0.738{\scriptsize$\pm$0.027} & 0.594{\scriptsize$\pm$0.025} & 0.325{\scriptsize$\pm$0.031} & 0.446{\scriptsize$\pm$0.038} \\
 & GMN & 0.730{\scriptsize$\pm$0.015} & 0.364{\scriptsize$\pm$0.007} & 0.880{\scriptsize$\pm$0.006} & 0.751{\scriptsize$\pm$0.007} & 0.561{\scriptsize$\pm$0.026} & 0.661{\scriptsize$\pm$0.022} \\
 & GraphEdX & \underline{0.542{\scriptsize$\pm$0.005}} & \underline{0.270{\scriptsize$\pm$0.002}} & \underline{0.931{\scriptsize$\pm$0.001}} & \underline{0.823{\scriptsize$\pm$0.002}} & \underline{0.807{\scriptsize$\pm$0.007}} & \underline{0.840{\scriptsize$\pm$0.005}} \\
 & GEN & \textit{0.590{\scriptsize$\pm$0.006}} & \textit{0.294{\scriptsize$\pm$0.003}} & \textit{0.925{\scriptsize$\pm$0.001}} & \textit{0.814{\scriptsize$\pm$0.002}} & \textit{0.773{\scriptsize$\pm$0.010}} & \textit{0.827{\scriptsize$\pm$0.007}} \\
 & EGSC & 0.736{\scriptsize$\pm$0.017} & 0.367{\scriptsize$\pm$0.009} & 0.890{\scriptsize$\pm$0.002} & 0.764{\scriptsize$\pm$0.002} & 0.628{\scriptsize$\pm$0.013} & 0.716{\scriptsize$\pm$0.010} \\
 & ERIC & 0.665{\scriptsize$\pm$0.011} & 0.331{\scriptsize$\pm$0.006} & 0.905{\scriptsize$\pm$0.002} & 0.785{\scriptsize$\pm$0.003} & 0.712{\scriptsize$\pm$0.010} & 0.778{\scriptsize$\pm$0.009} \\
 & GREED & 0.695{\scriptsize$\pm$0.011} & 0.346{\scriptsize$\pm$0.005} & 0.901{\scriptsize$\pm$0.002} & 0.779{\scriptsize$\pm$0.003} & 0.675{\scriptsize$\pm$0.011} & 0.754{\scriptsize$\pm$0.009} \\
 & GraSP & 0.618{\scriptsize$\pm$0.008} & 0.308{\scriptsize$\pm$0.004} & 0.923{\scriptsize$\pm$0.001} & \textit{0.814{\scriptsize$\pm$0.002}} & 0.767{\scriptsize$\pm$0.012} & \textit{0.827{\scriptsize$\pm$0.005}} \\
 & G2R & 0.656{\scriptsize$\pm$0.022} & 0.327{\scriptsize$\pm$0.011} & 0.912{\scriptsize$\pm$0.002} & 0.795{\scriptsize$\pm$0.003} & 0.746{\scriptsize$\pm$0.010} & 0.805{\scriptsize$\pm$0.010} \\
 & GEDGNN & 0.877{\scriptsize$\pm$0.018} & 0.437{\scriptsize$\pm$0.009} & 0.853{\scriptsize$\pm$0.006} & 0.718{\scriptsize$\pm$0.007} & 0.572{\scriptsize$\pm$0.015} & 0.653{\scriptsize$\pm$0.019} \\
 & GEDIOT & 0.729{\scriptsize$\pm$0.060} & 0.363{\scriptsize$\pm$0.030} & 0.901{\scriptsize$\pm$0.013} & 0.779{\scriptsize$\pm$0.015} & 0.721{\scriptsize$\pm$0.027} & 0.783{\scriptsize$\pm$0.021} \\
 & \method{} & \textbf{0.532{\scriptsize$\pm$0.008}} & \textbf{0.265{\scriptsize$\pm$0.004}} & \textbf{0.937{\scriptsize$\pm$0.001}} & \textbf{0.834{\scriptsize$\pm$0.002}} & \textbf{0.830{\scriptsize$\pm$0.009}} & \textbf{0.867{\scriptsize$\pm$0.007}} \\
\midrule
IMDBMulti & SimGNN & 34.989{\scriptsize$\pm$2.112} & 0.429{\scriptsize$\pm$0.026} & 0.767{\scriptsize$\pm$0.012} & 0.686{\scriptsize$\pm$0.013} & 0.677{\scriptsize$\pm$0.038} & 0.684{\scriptsize$\pm$0.036} \\
 & GMN & 4.341{\scriptsize$\pm$1.023} & 0.053{\scriptsize$\pm$0.013} & 0.938{\scriptsize$\pm$0.005} & \textit{0.870{\scriptsize$\pm$0.009}} & 0.829{\scriptsize$\pm$0.012} & 0.840{\scriptsize$\pm$0.016} \\
 & GraphEdX & TO & TO & TO & TO & TO & TO \\
 & GEN & \textit{3.830{\scriptsize$\pm$0.155}} & \textit{0.047{\scriptsize$\pm$0.002}} & \textit{0.939{\scriptsize$\pm$0.001}} & \textit{0.870{\scriptsize$\pm$0.002}} & 0.843{\scriptsize$\pm$0.015} & 0.860{\scriptsize$\pm$0.011} \\
 & EGSC & 25.251{\scriptsize$\pm$1.033} & 0.310{\scriptsize$\pm$0.013} & 0.888{\scriptsize$\pm$0.013} & 0.825{\scriptsize$\pm$0.012} & 0.851{\scriptsize$\pm$0.007} & 0.859{\scriptsize$\pm$0.006} \\
 & ERIC & 35.344{\scriptsize$\pm$2.406} & 0.433{\scriptsize$\pm$0.030} & 0.797{\scriptsize$\pm$0.024} & 0.745{\scriptsize$\pm$0.022} & 0.849{\scriptsize$\pm$0.009} & 0.855{\scriptsize$\pm$0.005} \\
 & GREED & \underline{3.672{\scriptsize$\pm$0.154}} & \underline{0.045{\scriptsize$\pm$0.002}} & 0.929{\scriptsize$\pm$0.006} & 0.856{\scriptsize$\pm$0.009} & 0.833{\scriptsize$\pm$0.019} & 0.846{\scriptsize$\pm$0.016} \\
 & GraSP & 4.088{\scriptsize$\pm$0.249} & 0.050{\scriptsize$\pm$0.003} & \underline{0.945{\scriptsize$\pm$0.003}} & \underline{0.890{\scriptsize$\pm$0.005}} & \textit{0.859{\scriptsize$\pm$0.006}} & \textit{0.875{\scriptsize$\pm$0.005}} \\
 & G2R & 31.392{\scriptsize$\pm$1.637} & 0.385{\scriptsize$\pm$0.020} & 0.805{\scriptsize$\pm$0.017} & 0.740{\scriptsize$\pm$0.014} & \underline{0.890{\scriptsize$\pm$0.006}} & \underline{0.895{\scriptsize$\pm$0.004}} \\
 & GEDGNN & 32.968{\scriptsize$\pm$10.265} & 0.404{\scriptsize$\pm$0.126} & 0.737{\scriptsize$\pm$0.041} & 0.644{\scriptsize$\pm$0.040} & 0.674{\scriptsize$\pm$0.065} & 0.700{\scriptsize$\pm$0.045} \\
 & GEDIOT & 18.502{\scriptsize$\pm$7.793} & 0.227{\scriptsize$\pm$0.096} & 0.824{\scriptsize$\pm$0.037} & 0.742{\scriptsize$\pm$0.039} & 0.757{\scriptsize$\pm$0.024} & 0.770{\scriptsize$\pm$0.028} \\
 & \method{} & \textbf{2.235{\scriptsize$\pm$0.068}} & \textbf{0.027{\scriptsize$\pm$0.001}} & \textbf{0.959{\scriptsize$\pm$0.001}} & \textbf{0.916{\scriptsize$\pm$0.001}} & \textbf{0.905{\scriptsize$\pm$0.005}} & \textbf{0.924{\scriptsize$\pm$0.003}} \\
\midrule
LINUX & SimGNN & 1.060{\scriptsize$\pm$0.204} & 0.492{\scriptsize$\pm$0.095} & 0.878{\scriptsize$\pm$0.041} & 0.778{\scriptsize$\pm$0.051} & 0.872{\scriptsize$\pm$0.078} & 0.898{\scriptsize$\pm$0.058} \\
 & GMN & 0.134{\scriptsize$\pm$0.007} & 0.062{\scriptsize$\pm$0.003} & 0.972{\scriptsize$\pm$0.000} & 0.905{\scriptsize$\pm$0.001} & 0.982{\scriptsize$\pm$0.002} & 0.987{\scriptsize$\pm$0.002} \\
 & GraphEdX & 0.139{\scriptsize$\pm$0.006} & 0.065{\scriptsize$\pm$0.003} & 0.972{\scriptsize$\pm$0.000} & 0.906{\scriptsize$\pm$0.001} & \textit{0.991{\scriptsize$\pm$0.002}} & 0.987{\scriptsize$\pm$0.002} \\
 & GEN & 0.089{\scriptsize$\pm$0.009} & 0.042{\scriptsize$\pm$0.004} & \textit{0.973{\scriptsize$\pm$0.000}} & 0.904{\scriptsize$\pm$0.001} & \underline{0.992{\scriptsize$\pm$0.005}} & \underline{0.992{\scriptsize$\pm$0.002}} \\
 & EGSC & 0.141{\scriptsize$\pm$0.021} & 0.066{\scriptsize$\pm$0.010} & \underline{0.974{\scriptsize$\pm$0.001}} & \underline{0.920{\scriptsize$\pm$0.001}} & 0.981{\scriptsize$\pm$0.008} & 0.980{\scriptsize$\pm$0.007} \\
 & ERIC & 0.078{\scriptsize$\pm$0.006} & 0.036{\scriptsize$\pm$0.003} & \underline{0.974{\scriptsize$\pm$0.000}} & \textit{0.916{\scriptsize$\pm$0.002}} & \textit{0.991{\scriptsize$\pm$0.003}} & \textbf{0.993{\scriptsize$\pm$0.002}} \\
 & GREED & 0.323{\scriptsize$\pm$0.003} & 0.150{\scriptsize$\pm$0.001} & 0.967{\scriptsize$\pm$0.001} & 0.904{\scriptsize$\pm$0.001} & 0.972{\scriptsize$\pm$0.005} & 0.970{\scriptsize$\pm$0.003} \\
 & GraSP & \textit{0.059{\scriptsize$\pm$0.035}} & \textit{0.027{\scriptsize$\pm$0.016}} & \textbf{0.975{\scriptsize$\pm$0.000}} & \textbf{0.922{\scriptsize$\pm$0.001}} & 0.990{\scriptsize$\pm$0.005} & \textit{0.988{\scriptsize$\pm$0.003}} \\
 & G2R & \underline{0.031{\scriptsize$\pm$0.005}} & \underline{0.014{\scriptsize$\pm$0.002}} & \textit{0.973{\scriptsize$\pm$0.000}} & 0.898{\scriptsize$\pm$0.001} & \textbf{0.995{\scriptsize$\pm$0.001}} & \underline{0.992{\scriptsize$\pm$0.002}} \\
 & GEDGNN & 0.271{\scriptsize$\pm$0.032} & 0.126{\scriptsize$\pm$0.015} & 0.968{\scriptsize$\pm$0.002} & 0.906{\scriptsize$\pm$0.005} & 0.952{\scriptsize$\pm$0.011} & 0.951{\scriptsize$\pm$0.009} \\
 & GEDIOT & 0.228{\scriptsize$\pm$0.040} & 0.106{\scriptsize$\pm$0.019} & 0.972{\scriptsize$\pm$0.001} & 0.912{\scriptsize$\pm$0.003} & 0.967{\scriptsize$\pm$0.019} & 0.974{\scriptsize$\pm$0.013} \\
 & \method{} & \textbf{0.020{\scriptsize$\pm$0.004}} & \textbf{0.009{\scriptsize$\pm$0.002}} & \underline{0.974{\scriptsize$\pm$0.000}} & 0.909{\scriptsize$\pm$0.001} & \underline{0.992{\scriptsize$\pm$0.003}} & \textbf{0.993{\scriptsize$\pm$0.002}} \\
\midrule
PTC & SimGNN & 9.898{\scriptsize$\pm$1.258} & 0.415{\scriptsize$\pm$0.053} & 0.922{\scriptsize$\pm$0.032} & 0.816{\scriptsize$\pm$0.038} & 0.415{\scriptsize$\pm$0.084} & 0.542{\scriptsize$\pm$0.087} \\
 & GMN & \underline{3.039{\scriptsize$\pm$0.143}} & \underline{0.127{\scriptsize$\pm$0.006}} & 0.933{\scriptsize$\pm$0.006} & 0.830{\scriptsize$\pm$0.008} & 0.475{\scriptsize$\pm$0.031} & 0.579{\scriptsize$\pm$0.023} \\
 & GraphEdX & 9.512{\scriptsize$\pm$1.192} & 0.398{\scriptsize$\pm$0.050} & 0.848{\scriptsize$\pm$0.013} & 0.724{\scriptsize$\pm$0.015} & 0.481{\scriptsize$\pm$0.018} & 0.547{\scriptsize$\pm$0.018} \\
 & GEN & 3.793{\scriptsize$\pm$0.089} & 0.159{\scriptsize$\pm$0.004} & 0.937{\scriptsize$\pm$0.002} & 0.841{\scriptsize$\pm$0.002} & \textit{0.589{\scriptsize$\pm$0.016}} & 0.683{\scriptsize$\pm$0.008} \\
 & EGSC & 4.850{\scriptsize$\pm$0.652} & 0.203{\scriptsize$\pm$0.027} & 0.934{\scriptsize$\pm$0.005} & 0.835{\scriptsize$\pm$0.005} & 0.573{\scriptsize$\pm$0.016} & 0.675{\scriptsize$\pm$0.012} \\
 & ERIC & 3.990{\scriptsize$\pm$0.333} & 0.167{\scriptsize$\pm$0.014} & 0.939{\scriptsize$\pm$0.004} & 0.844{\scriptsize$\pm$0.005} & 0.574{\scriptsize$\pm$0.015} & 0.678{\scriptsize$\pm$0.010} \\
 & GREED & \textit{3.635{\scriptsize$\pm$0.088}} & \textit{0.152{\scriptsize$\pm$0.004}} & 0.919{\scriptsize$\pm$0.005} & 0.810{\scriptsize$\pm$0.006} & 0.437{\scriptsize$\pm$0.022} & 0.562{\scriptsize$\pm$0.010} \\
 & GraSP & 3.672{\scriptsize$\pm$0.083} & 0.154{\scriptsize$\pm$0.003} & \textit{0.946{\scriptsize$\pm$0.001}} & \textit{0.853{\scriptsize$\pm$0.002}} & 0.581{\scriptsize$\pm$0.011} & \textit{0.689{\scriptsize$\pm$0.007}} \\
 & G2R & 3.857{\scriptsize$\pm$0.396} & 0.162{\scriptsize$\pm$0.017} & \underline{0.949{\scriptsize$\pm$0.003}} & \underline{0.856{\scriptsize$\pm$0.003}} & \underline{0.616{\scriptsize$\pm$0.011}} & \underline{0.719{\scriptsize$\pm$0.004}} \\
 & GEDGNN & 38.990{\scriptsize$\pm$2.172} & 1.633{\scriptsize$\pm$0.091} & 0.639{\scriptsize$\pm$0.173} & 0.500{\scriptsize$\pm$0.147} & 0.316{\scriptsize$\pm$0.074} & 0.398{\scriptsize$\pm$0.080} \\
 & GEDIOT & 38.133{\scriptsize$\pm$9.133} & 1.597{\scriptsize$\pm$0.382} & 0.558{\scriptsize$\pm$0.241} & 0.508{\scriptsize$\pm$0.157} & 0.350{\scriptsize$\pm$0.070} & 0.436{\scriptsize$\pm$0.070} \\
 & \method{} & \textbf{2.436{\scriptsize$\pm$0.036}} & \textbf{0.102{\scriptsize$\pm$0.002}} & \textbf{0.958{\scriptsize$\pm$0.001}} & \textbf{0.872{\scriptsize$\pm$0.001}} & \textbf{0.642{\scriptsize$\pm$0.007}} & \textbf{0.737{\scriptsize$\pm$0.004}} \\
\bottomrule
\end{tabular}
\end{sidewaystable}

\textbf{GED.} The detailed results align with the main experiments. RAE of \method{} ranges from 0.009 on LINUX to 0.265 on AIDS700nef, i.e., the total error of \method{} is between 0.9\% and 26.5\% of that of the mean predictor. P@20 follows P@10, and \method{} has the highest P@20 on all four datasets (tied with ERIC on LINUX). The standard deviations indicate that these results are not sensitive to the seed: the standard deviations of the ranking metrics of \method{} are at most 0.009 across the 10 seeds, whereas several baselines vary much more. On PTC, GEDGNN and GEDIOT have an RAE above 1, meaning that these models produce larger errors than the mean predictor. On IMDBMulti, G2R has the second-highest P@10 and P@20 but also one of the largest errors, so it identifies the most similar graphs but does not recover the scale of GED. On LINUX, all methods except SimGNN lie within a narrow range in $\rho$, $\tau$, and P@$k$, which is a consequence of the tied labels. GraphEdX cannot complete a single pass over the IMDBMulti training pairs within the time limit, so it is reported as a timeout.

\begin{sidewaystable}[htbp]
\footnotesize
\centering
\caption{MCS prediction on the benchmark datasets, all metrics. Mean $\pm$ standard deviation over ten seeds. MAE: mean absolute error, RAE: relative absolute error, TO: timeout. Best result in \textbf{bold}, second best \underline{underlined}, and third best in \textit{italic}.}
\label{tab:app_bench_mcs}
\begin{tabular}{@{}llcccccc@{}}
\toprule
Dataset & Model & MAE & RAE & $\rho$ & $\tau$ & P@10 & P@20 \\
\midrule
AIDS700nef & SimGNN & 0.711{\scriptsize$\pm$0.015} & 0.534{\scriptsize$\pm$0.011} & 0.756{\scriptsize$\pm$0.008} & 0.623{\scriptsize$\pm$0.008} & 0.387{\scriptsize$\pm$0.018} & 0.471{\scriptsize$\pm$0.011} \\
 & GMN & 0.422{\scriptsize$\pm$0.017} & 0.317{\scriptsize$\pm$0.013} & 0.888{\scriptsize$\pm$0.007} & 0.778{\scriptsize$\pm$0.008} & 0.769{\scriptsize$\pm$0.021} & 0.796{\scriptsize$\pm$0.017} \\
 & GraphEdX & 0.482{\scriptsize$\pm$0.005} & 0.362{\scriptsize$\pm$0.004} & 0.880{\scriptsize$\pm$0.002} & 0.764{\scriptsize$\pm$0.002} & 0.771{\scriptsize$\pm$0.005} & 0.792{\scriptsize$\pm$0.006} \\
 & GEN & \underline{0.303{\scriptsize$\pm$0.011}} & \underline{0.227{\scriptsize$\pm$0.008}} & \underline{0.933{\scriptsize$\pm$0.003}} & \underline{0.836{\scriptsize$\pm$0.003}} & \underline{0.874{\scriptsize$\pm$0.013}} & \underline{0.896{\scriptsize$\pm$0.007}} \\
 & EGSC & 0.436{\scriptsize$\pm$0.012} & 0.327{\scriptsize$\pm$0.009} & 0.892{\scriptsize$\pm$0.005} & 0.785{\scriptsize$\pm$0.005} & 0.740{\scriptsize$\pm$0.017} & 0.792{\scriptsize$\pm$0.011} \\
 & ERIC & 0.339{\scriptsize$\pm$0.012} & 0.255{\scriptsize$\pm$0.009} & 0.907{\scriptsize$\pm$0.004} & 0.806{\scriptsize$\pm$0.004} & 0.801{\scriptsize$\pm$0.007} & 0.835{\scriptsize$\pm$0.003} \\
 & GREED & 0.472{\scriptsize$\pm$0.010} & 0.355{\scriptsize$\pm$0.007} & 0.881{\scriptsize$\pm$0.006} & 0.769{\scriptsize$\pm$0.007} & 0.761{\scriptsize$\pm$0.016} & 0.794{\scriptsize$\pm$0.016} \\
 & GraSP & \textit{0.331{\scriptsize$\pm$0.007}} & \textit{0.248{\scriptsize$\pm$0.005}} & \textit{0.926{\scriptsize$\pm$0.002}} & \textit{0.829{\scriptsize$\pm$0.002}} & \textit{0.871{\scriptsize$\pm$0.009}} & \textit{0.895{\scriptsize$\pm$0.005}} \\
 & G2R & \underline{0.303{\scriptsize$\pm$0.010}} & \underline{0.227{\scriptsize$\pm$0.008}} & 0.920{\scriptsize$\pm$0.003} & 0.820{\scriptsize$\pm$0.002} & 0.823{\scriptsize$\pm$0.009} & 0.854{\scriptsize$\pm$0.008} \\
 & \method{} & \textbf{0.264{\scriptsize$\pm$0.007}} & \textbf{0.198{\scriptsize$\pm$0.005}} & \textbf{0.935{\scriptsize$\pm$0.001}} & \textbf{0.839{\scriptsize$\pm$0.001}} & \textbf{0.889{\scriptsize$\pm$0.006}} & \textbf{0.904{\scriptsize$\pm$0.004}} \\
\midrule
IMDBMulti & SimGNN & 0.451{\scriptsize$\pm$0.040} & 0.260{\scriptsize$\pm$0.023} & 0.870{\scriptsize$\pm$0.003} & 0.784{\scriptsize$\pm$0.004} & 0.909{\scriptsize$\pm$0.009} & 0.922{\scriptsize$\pm$0.006} \\
 & GMN & 0.197{\scriptsize$\pm$0.027} & 0.114{\scriptsize$\pm$0.016} & \textit{0.881{\scriptsize$\pm$0.001}} & 0.798{\scriptsize$\pm$0.001} & 0.937{\scriptsize$\pm$0.011} & 0.950{\scriptsize$\pm$0.008} \\
 & GraphEdX & TO & TO & TO & TO & TO & TO \\
 & GEN & 0.295{\scriptsize$\pm$0.033} & 0.170{\scriptsize$\pm$0.019} & 0.868{\scriptsize$\pm$0.003} & 0.772{\scriptsize$\pm$0.005} & 0.879{\scriptsize$\pm$0.024} & 0.889{\scriptsize$\pm$0.019} \\
 & EGSC & 0.233{\scriptsize$\pm$0.030} & 0.134{\scriptsize$\pm$0.017} & 0.880{\scriptsize$\pm$0.002} & \textit{0.801{\scriptsize$\pm$0.002}} & 0.912{\scriptsize$\pm$0.019} & 0.928{\scriptsize$\pm$0.013} \\
 & ERIC & 0.361{\scriptsize$\pm$0.084} & 0.208{\scriptsize$\pm$0.049} & 0.862{\scriptsize$\pm$0.008} & 0.774{\scriptsize$\pm$0.009} & 0.868{\scriptsize$\pm$0.028} & 0.884{\scriptsize$\pm$0.016} \\
 & GREED & 0.289{\scriptsize$\pm$0.011} & 0.167{\scriptsize$\pm$0.006} & 0.876{\scriptsize$\pm$0.002} & 0.787{\scriptsize$\pm$0.003} & 0.899{\scriptsize$\pm$0.017} & 0.914{\scriptsize$\pm$0.014} \\
 & GraSP & \textit{0.141{\scriptsize$\pm$0.017}} & \textit{0.081{\scriptsize$\pm$0.010}} & \underline{0.883{\scriptsize$\pm$0.001}} & \underline{0.804{\scriptsize$\pm$0.001}} & \textit{0.947{\scriptsize$\pm$0.003}} & \textit{0.957{\scriptsize$\pm$0.004}} \\
 & G2R & \underline{0.087{\scriptsize$\pm$0.006}} & \underline{0.050{\scriptsize$\pm$0.004}} & 0.880{\scriptsize$\pm$0.000} & 0.781{\scriptsize$\pm$0.000} & \underline{0.959{\scriptsize$\pm$0.005}} & \underline{0.968{\scriptsize$\pm$0.003}} \\
 & \method{} & \textbf{0.081{\scriptsize$\pm$0.003}} & \textbf{0.047{\scriptsize$\pm$0.002}} & \textbf{0.885{\scriptsize$\pm$0.000}} & \textbf{0.808{\scriptsize$\pm$0.000}} & \textbf{0.964{\scriptsize$\pm$0.002}} & \textbf{0.969{\scriptsize$\pm$0.001}} \\
\midrule
LINUX & SimGNN & 0.368{\scriptsize$\pm$0.054} & 0.464{\scriptsize$\pm$0.068} & 0.751{\scriptsize$\pm$0.021} & 0.654{\scriptsize$\pm$0.021} & 0.666{\scriptsize$\pm$0.046} & 0.740{\scriptsize$\pm$0.048} \\
 & GMN & 0.084{\scriptsize$\pm$0.009} & 0.106{\scriptsize$\pm$0.011} & \underline{0.813{\scriptsize$\pm$0.000}} & \textit{0.715{\scriptsize$\pm$0.001}} & 0.992{\scriptsize$\pm$0.002} & 0.994{\scriptsize$\pm$0.002} \\
 & GraphEdX & 0.105{\scriptsize$\pm$0.003} & 0.133{\scriptsize$\pm$0.004} & \underline{0.813{\scriptsize$\pm$0.000}} & \textit{0.715{\scriptsize$\pm$0.001}} & \textbf{0.998{\scriptsize$\pm$0.002}} & \underline{0.998{\scriptsize$\pm$0.001}} \\
 & GEN & 0.048{\scriptsize$\pm$0.004} & 0.060{\scriptsize$\pm$0.005} & \underline{0.813{\scriptsize$\pm$0.000}} & 0.714{\scriptsize$\pm$0.001} & 0.994{\scriptsize$\pm$0.004} & \textit{0.997{\scriptsize$\pm$0.003}} \\
 & EGSC & 0.081{\scriptsize$\pm$0.008} & 0.102{\scriptsize$\pm$0.011} & \textit{0.812{\scriptsize$\pm$0.002}} & \underline{0.719{\scriptsize$\pm$0.002}} & 0.980{\scriptsize$\pm$0.014} & 0.985{\scriptsize$\pm$0.007} \\
 & ERIC & 0.046{\scriptsize$\pm$0.006} & 0.058{\scriptsize$\pm$0.007} & \textit{0.812{\scriptsize$\pm$0.001}} & \textit{0.715{\scriptsize$\pm$0.001}} & 0.990{\scriptsize$\pm$0.006} & 0.996{\scriptsize$\pm$0.002} \\
 & GREED & 0.158{\scriptsize$\pm$0.001} & 0.200{\scriptsize$\pm$0.002} & \textit{0.812{\scriptsize$\pm$0.000}} & \underline{0.719{\scriptsize$\pm$0.000}} & 0.992{\scriptsize$\pm$0.002} & 0.991{\scriptsize$\pm$0.002} \\
 & GraSP & \textit{0.035{\scriptsize$\pm$0.005}} & \textit{0.044{\scriptsize$\pm$0.007}} & \textbf{0.814{\scriptsize$\pm$0.001}} & \textbf{0.721{\scriptsize$\pm$0.001}} & \textit{0.995{\scriptsize$\pm$0.004}} & \underline{0.998{\scriptsize$\pm$0.002}} \\
 & G2R & \underline{0.021{\scriptsize$\pm$0.004}} & \underline{0.027{\scriptsize$\pm$0.005}} & \textit{0.812{\scriptsize$\pm$0.001}} & 0.701{\scriptsize$\pm$0.001} & 0.994{\scriptsize$\pm$0.003} & 0.996{\scriptsize$\pm$0.002} \\
 & \method{} & \textbf{0.015{\scriptsize$\pm$0.001}} & \textbf{0.019{\scriptsize$\pm$0.002}} & \textbf{0.814{\scriptsize$\pm$0.000}} & \textit{0.715{\scriptsize$\pm$0.001}} & \underline{0.997{\scriptsize$\pm$0.001}} & \textbf{0.999{\scriptsize$\pm$0.001}} \\
\midrule
PTC & SimGNN & 1.707{\scriptsize$\pm$0.100} & 0.473{\scriptsize$\pm$0.028} & 0.725{\scriptsize$\pm$0.027} & 0.586{\scriptsize$\pm$0.025} & 0.397{\scriptsize$\pm$0.034} & 0.492{\scriptsize$\pm$0.025} \\
 & GMN & 1.581{\scriptsize$\pm$0.070} & 0.438{\scriptsize$\pm$0.019} & 0.790{\scriptsize$\pm$0.010} & 0.653{\scriptsize$\pm$0.011} & 0.511{\scriptsize$\pm$0.014} & 0.606{\scriptsize$\pm$0.017} \\
 & GraphEdX & 5.339{\scriptsize$\pm$1.307} & 1.479{\scriptsize$\pm$0.362} & 0.516{\scriptsize$\pm$0.099} & 0.398{\scriptsize$\pm$0.086} & 0.415{\scriptsize$\pm$0.055} & 0.487{\scriptsize$\pm$0.063} \\
 & GEN & 1.392{\scriptsize$\pm$0.073} & 0.386{\scriptsize$\pm$0.020} & 0.865{\scriptsize$\pm$0.011} & 0.748{\scriptsize$\pm$0.016} & \textit{0.634{\scriptsize$\pm$0.053}} & 0.707{\scriptsize$\pm$0.039} \\
 & EGSC & 1.482{\scriptsize$\pm$0.105} & 0.410{\scriptsize$\pm$0.029} & 0.813{\scriptsize$\pm$0.014} & 0.687{\scriptsize$\pm$0.014} & 0.499{\scriptsize$\pm$0.020} & 0.628{\scriptsize$\pm$0.021} \\
 & ERIC & 1.201{\scriptsize$\pm$0.052} & 0.333{\scriptsize$\pm$0.014} & 0.840{\scriptsize$\pm$0.014} & 0.717{\scriptsize$\pm$0.014} & 0.553{\scriptsize$\pm$0.027} & 0.655{\scriptsize$\pm$0.024} \\
 & GREED & 1.988{\scriptsize$\pm$0.112} & 0.551{\scriptsize$\pm$0.031} & 0.797{\scriptsize$\pm$0.007} & 0.665{\scriptsize$\pm$0.007} & 0.540{\scriptsize$\pm$0.024} & 0.632{\scriptsize$\pm$0.018} \\
 & GraSP & \underline{1.143{\scriptsize$\pm$0.048}} & \underline{0.317{\scriptsize$\pm$0.013}} & \textit{0.868{\scriptsize$\pm$0.009}} & \textit{0.753{\scriptsize$\pm$0.010}} & \underline{0.636{\scriptsize$\pm$0.023}} & \underline{0.724{\scriptsize$\pm$0.014}} \\
 & G2R & \textit{1.149{\scriptsize$\pm$0.025}} & \textit{0.318{\scriptsize$\pm$0.007}} & \underline{0.884{\scriptsize$\pm$0.006}} & \underline{0.771{\scriptsize$\pm$0.006}} & 0.633{\scriptsize$\pm$0.016} & \textit{0.719{\scriptsize$\pm$0.012}} \\
 & \method{} & \textbf{0.804{\scriptsize$\pm$0.014}} & \textbf{0.223{\scriptsize$\pm$0.004}} & \textbf{0.924{\scriptsize$\pm$0.003}} & \textbf{0.827{\scriptsize$\pm$0.003}} & \textbf{0.721{\scriptsize$\pm$0.010}} & \textbf{0.800{\scriptsize$\pm$0.008}} \\
\bottomrule
\end{tabular}
\end{sidewaystable}

\textbf{MCS.} Measurements show the same pattern for MCS. \method{} is the best method across all six metrics on AIDS700nef, IMDBMulti, and PTC, including RAE and P@20. On LINUX, it has the lowest MAE and RAE, the best $\rho$ (tied with GraSP), and the best P@20. The exceptions are $\tau$ on LINUX, where GraSP, EGSC, and GREED are slightly higher, and P@10 on LINUX, where GraphEdX is slightly higher. On this dataset, $\rho$ is between 0.812 and 0.814 for all methods except SimGNN, indicating that the ranking metrics are close to their ceiling and that the remaining differences come from how each method orders graphs with equal labels. The variation across seeds is small for most methods. The largest differences between methods are on PTC, which contains the largest graphs among the benchmark datasets.

The additional metrics and standard deviations do not change the conclusions of the main experiments. \method{} is the most accurate method in the error metrics on all datasets and both tasks, and in the ranking metrics on all datasets except LINUX. \method{} is among the most stable methods across seeds. Since it achieves the lowest errors and the highest ranking metrics across most datasets, \method{} learns not only the graph order but also the GED and MCS values.

\subsection{Large-Scale Datasets}
\label{app:large_all}

In addition to the four large-scale datasets in \Cref{sec:datasets}, we provide six further datasets derived from Wiki, CiteSeer, BlogCatalog, Flickr, com-DBLP, and com-Amazon, and we report all ten datasets for $M \in \{50, 200, 1000\}$. Since computation on these datasets is expensive, we select the best configuration on the validation set at $M = 200$ and reuse it for $M = 50$ and $M = 1000$ without further search. The details of these datasets are given in \Cref{app:large_datasets}. Due to computational constraints, we did not extend the baselines to these datasets and $M$ values. \Cref{tab:app_large} extends \Cref{tab:large_main}, without the runs that exceed the memory limit. Values above 1,000 are shown as $>$1,000. \Cref{tab:app_large_all} gives the results of \method{} on all ten datasets and $M$ values, where values above 10,000 are shown as $>$10,000.

\begin{sidewaystable}[htbp]
\footnotesize
\centering
\setlength{\tabcolsep}{4pt}
\caption{GED prediction on the four large datasets, all metrics. Mean $\pm$ standard deviation over three seeds. MAE: mean absolute error, RAE: relative absolute error. Best result in \textbf{bold}, second best \underline{underlined}, and third best in \textit{italic}.}
\label{tab:app_large}
\begin{tabular}{@{}llcccccc@{}}
\toprule
Dataset & Model & MAE & RAE & $\rho$ & $\tau$ & P@10 & P@20 \\
\midrule
Cora & SimGNN & 377.855{\scriptsize$\pm$0.000} & 2.734{\scriptsize$\pm$0.000} & - & - & 0.045{\scriptsize$\pm$0.000} & 0.123{\scriptsize$\pm$0.000} \\
 & GMN & 42.653{\scriptsize$\pm$0.628} & 0.309{\scriptsize$\pm$0.005} & 0.950{\scriptsize$\pm$0.001} & 0.822{\scriptsize$\pm$0.001} & 0.723{\scriptsize$\pm$0.026} & 0.800{\scriptsize$\pm$0.009} \\
 & GEN & 146.256{\scriptsize$\pm$35.369} & 1.058{\scriptsize$\pm$0.256} & 0.195{\scriptsize$\pm$0.360} & 0.144{\scriptsize$\pm$0.251} & 0.077{\scriptsize$\pm$0.030} & 0.147{\scriptsize$\pm$0.063} \\
 & EGSC & 377.855{\scriptsize$\pm$0.000} & 2.734{\scriptsize$\pm$0.000} & - & - & 0.045{\scriptsize$\pm$0.000} & 0.123{\scriptsize$\pm$0.000} \\
 & ERIC & 137.089{\scriptsize$\pm$0.403} & 0.992{\scriptsize$\pm$0.003} & 0.033{\scriptsize$\pm$0.115} & 0.023{\scriptsize$\pm$0.076} & 0.078{\scriptsize$\pm$0.051} & 0.125{\scriptsize$\pm$0.045} \\
 & GREED & \textit{31.019{\scriptsize$\pm$1.861}} & \textit{0.224{\scriptsize$\pm$0.013}} & \textit{0.987{\scriptsize$\pm$0.003}} & \textit{0.914{\scriptsize$\pm$0.010}} & \textit{0.844{\scriptsize$\pm$0.039}} & \textit{0.892{\scriptsize$\pm$0.017}} \\
 & GraSP & 41.610{\scriptsize$\pm$6.636} & 0.301{\scriptsize$\pm$0.048} & 0.955{\scriptsize$\pm$0.020} & 0.835{\scriptsize$\pm$0.042} & 0.711{\scriptsize$\pm$0.046} & 0.823{\scriptsize$\pm$0.032} \\
 & G2R & \textbf{25.978{\scriptsize$\pm$1.829}} & \textbf{0.188{\scriptsize$\pm$0.013}} & \underline{0.989{\scriptsize$\pm$0.003}} & \underline{0.925{\scriptsize$\pm$0.012}} & \underline{0.866{\scriptsize$\pm$0.017}} & \underline{0.921{\scriptsize$\pm$0.012}} \\
 & GEDGNN & 377.855{\scriptsize$\pm$0.000} & 2.734{\scriptsize$\pm$0.000} & 0.162{\scriptsize$\pm$0.001} & 0.112{\scriptsize$\pm$0.001} & 0.176{\scriptsize$\pm$0.004} & 0.257{\scriptsize$\pm$0.001} \\
 & GEDIOT & 372.094{\scriptsize$\pm$0.500} & 2.692{\scriptsize$\pm$0.004} & 0.187{\scriptsize$\pm$0.000} & 0.128{\scriptsize$\pm$0.000} & 0.187{\scriptsize$\pm$0.001} & 0.268{\scriptsize$\pm$0.001} \\
 & \method{} & \underline{26.257{\scriptsize$\pm$1.093}} & \underline{0.190{\scriptsize$\pm$0.008}} & \textbf{0.991{\scriptsize$\pm$0.000}} & \textbf{0.939{\scriptsize$\pm$0.001}} & \textbf{0.913{\scriptsize$\pm$0.012}} & \textbf{0.940{\scriptsize$\pm$0.003}} \\
\midrule
PubMed & SimGNN & $>$1,000 & 88.598{\scriptsize$\pm$148.968} & -0.271{\scriptsize$\pm$--} & -0.188{\scriptsize$\pm$--} & 0.061{\scriptsize$\pm$0.042} & 0.107{\scriptsize$\pm$0.038} \\
 & GMN & \textit{677.724{\scriptsize$\pm$38.684}} & \textit{0.564{\scriptsize$\pm$0.032}} & \textit{0.755{\scriptsize$\pm$0.048}} & \textit{0.582{\scriptsize$\pm$0.043}} & \textit{0.562{\scriptsize$\pm$0.054}} & \textit{0.613{\scriptsize$\pm$0.030}} \\
 & EGSC & $>$1,000 & 2.591{\scriptsize$\pm$0.000} & - & - & 0.085{\scriptsize$\pm$0.000} & 0.129{\scriptsize$\pm$0.000} \\
 & ERIC & $>$1,000 & 1.010{\scriptsize$\pm$0.001} & -0.270{\scriptsize$\pm$0.001} & -0.187{\scriptsize$\pm$0.001} & 0.013{\scriptsize$\pm$0.000} & 0.062{\scriptsize$\pm$0.000} \\
 & GREED & $>$1,000 & 1.089{\scriptsize$\pm$0.192} & 0.181{\scriptsize$\pm$0.103} & 0.105{\scriptsize$\pm$0.081} & 0.298{\scriptsize$\pm$0.149} & 0.328{\scriptsize$\pm$0.118} \\
 & G2R & \textbf{212.715{\scriptsize$\pm$26.104}} & \textbf{0.177{\scriptsize$\pm$0.022}} & \underline{0.984{\scriptsize$\pm$0.003}} & \underline{0.908{\scriptsize$\pm$0.008}} & \underline{0.833{\scriptsize$\pm$0.020}} & \underline{0.892{\scriptsize$\pm$0.017}} \\
 & \method{} & \underline{222.412{\scriptsize$\pm$13.464}} & \underline{0.185{\scriptsize$\pm$0.011}} & \textbf{0.986{\scriptsize$\pm$0.003}} & \textbf{0.914{\scriptsize$\pm$0.008}} & \textbf{0.880{\scriptsize$\pm$0.011}} & \textbf{0.908{\scriptsize$\pm$0.014}} \\
\midrule
Amazon C. & SimGNN & $>$1,000 & 34.948{\scriptsize$\pm$29.231} & -0.165{\scriptsize$\pm$0.134} & -0.114{\scriptsize$\pm$0.091} & 0.077{\scriptsize$\pm$0.043} & 0.107{\scriptsize$\pm$0.062} \\
 & GMN & \textit{$>$1,000} & \textit{0.715{\scriptsize$\pm$0.109}} & \textit{0.697{\scriptsize$\pm$0.147}} & \textit{0.538{\scriptsize$\pm$0.110}} & \textit{0.502{\scriptsize$\pm$0.075}} & \textit{0.568{\scriptsize$\pm$0.051}} \\
 & EGSC & $>$1,000 & 1.609{\scriptsize$\pm$1.003} & -0.242{\scriptsize$\pm$0.000} & -0.167{\scriptsize$\pm$0.000} & 0.063{\scriptsize$\pm$0.019} & 0.090{\scriptsize$\pm$0.032} \\
 & ERIC & $>$1,000 & 1.070{\scriptsize$\pm$0.058} & -0.150{\scriptsize$\pm$0.163} & -0.104{\scriptsize$\pm$0.111} & 0.056{\scriptsize$\pm$0.004} & 0.091{\scriptsize$\pm$0.034} \\
 & GREED & $>$1,000 & $>$1,000 & 0.189{\scriptsize$\pm$0.000} & 0.151{\scriptsize$\pm$0.000} & 0.181{\scriptsize$\pm$0.001} & 0.264{\scriptsize$\pm$0.000} \\
 & GraSP & $>$1,000 & 2.768{\scriptsize$\pm$0.000} & - & - & 0.085{\scriptsize$\pm$0.000} & 0.128{\scriptsize$\pm$0.000} \\
 & G2R & \underline{$>$1,000} & \underline{0.380{\scriptsize$\pm$0.038}} & \underline{0.908{\scriptsize$\pm$0.027}} & \underline{0.762{\scriptsize$\pm$0.034}} & \underline{0.652{\scriptsize$\pm$0.065}} & \underline{0.752{\scriptsize$\pm$0.028}} \\
 & GEDIOT & $>$1,000 & 2.763{\scriptsize$\pm$0.004} & 0.179{\scriptsize$\pm$0.159} & 0.129{\scriptsize$\pm$0.125} & 0.150{\scriptsize$\pm$0.043} & 0.175{\scriptsize$\pm$0.005} \\
 & \method{} & \textbf{975.351{\scriptsize$\pm$120.121}} & \textbf{0.241{\scriptsize$\pm$0.030}} & \textbf{0.979{\scriptsize$\pm$0.002}} & \textbf{0.895{\scriptsize$\pm$0.004}} & \textbf{0.852{\scriptsize$\pm$0.010}} & \textbf{0.893{\scriptsize$\pm$0.006}} \\
\midrule
ogbn-arxiv & GREED & \textit{$>$1,000} & \textit{$>$1,000} & \textit{0.252{\scriptsize$\pm$0.000}} & \textit{0.187{\scriptsize$\pm$0.000}} & \textit{0.173{\scriptsize$\pm$0.002}} & \textit{0.238{\scriptsize$\pm$0.001}} \\
 & G2R & \underline{$>$1,000} & \underline{0.958{\scriptsize$\pm$0.004}} & \underline{0.289{\scriptsize$\pm$0.030}} & \underline{0.212{\scriptsize$\pm$0.019}} & \underline{0.183{\scriptsize$\pm$0.024}} & \underline{0.240{\scriptsize$\pm$0.006}} \\
 & \method{} & \textbf{$>$1,000} & \textbf{0.342{\scriptsize$\pm$0.017}} & \textbf{0.956{\scriptsize$\pm$0.006}} & \textbf{0.831{\scriptsize$\pm$0.010}} & \textbf{0.713{\scriptsize$\pm$0.015}} & \textbf{0.778{\scriptsize$\pm$0.008}} \\
\bottomrule
\end{tabular}
\end{sidewaystable}

\begin{sidewaystable}[htbp]
\footnotesize
\centering
\caption{\method{} on all large-scale datasets and collection sizes $M$. Mean $\pm$ standard deviation over three seeds. RAE: relative absolute error. P@$k$ is not comparable across $M$, since the number of reference graphs grows with $M$.}
\label{tab:app_large_all}
\begin{tabular}{@{}llcccccc@{}}
\toprule
Dataset & $M$ & MAE & RAE & $\rho$ & $\tau$ & P@10 & P@20 \\
\midrule
Cora & 50 & 19.450{\scriptsize$\pm$1.685} & 0.238{\scriptsize$\pm$0.021} & 0.976{\scriptsize$\pm$0.008} & 0.911{\scriptsize$\pm$0.017} & 0.930{\scriptsize$\pm$0.026} & 0.948{\scriptsize$\pm$0.008} \\
 & 200 & 26.257{\scriptsize$\pm$1.093} & 0.190{\scriptsize$\pm$0.008} & 0.991{\scriptsize$\pm$0.000} & 0.939{\scriptsize$\pm$0.001} & 0.913{\scriptsize$\pm$0.012} & 0.940{\scriptsize$\pm$0.003} \\
 & 1000 & 22.124{\scriptsize$\pm$0.270} & 0.168{\scriptsize$\pm$0.002} & 0.989{\scriptsize$\pm$0.000} & 0.917{\scriptsize$\pm$0.001} & 0.867{\scriptsize$\pm$0.002} & 0.891{\scriptsize$\pm$0.002} \\
\midrule
CiteSeer & 50 & 15.603{\scriptsize$\pm$0.524} & 0.172{\scriptsize$\pm$0.006} & 0.972{\scriptsize$\pm$0.004} & 0.904{\scriptsize$\pm$0.004} & 0.947{\scriptsize$\pm$0.006} & 0.938{\scriptsize$\pm$0.008} \\
 & 200 & 26.887{\scriptsize$\pm$2.199} & 0.202{\scriptsize$\pm$0.017} & 0.986{\scriptsize$\pm$0.004} & 0.922{\scriptsize$\pm$0.016} & 0.911{\scriptsize$\pm$0.012} & 0.937{\scriptsize$\pm$0.012} \\
 & 1000 & 27.025{\scriptsize$\pm$5.811} & 0.187{\scriptsize$\pm$0.040} & 0.982{\scriptsize$\pm$0.009} & 0.896{\scriptsize$\pm$0.033} & 0.863{\scriptsize$\pm$0.013} & 0.879{\scriptsize$\pm$0.018} \\
\midrule
PubMed & 50 & 258.053{\scriptsize$\pm$28.756} & 0.367{\scriptsize$\pm$0.041} & 0.981{\scriptsize$\pm$0.000} & 0.912{\scriptsize$\pm$0.001} & 0.913{\scriptsize$\pm$0.006} & 0.970{\scriptsize$\pm$0.000} \\
 & 200 & 222.412{\scriptsize$\pm$13.464} & 0.185{\scriptsize$\pm$0.011} & 0.986{\scriptsize$\pm$0.003} & 0.914{\scriptsize$\pm$0.008} & 0.880{\scriptsize$\pm$0.011} & 0.908{\scriptsize$\pm$0.014} \\
 & 1000 & 258.404{\scriptsize$\pm$35.399} & 0.225{\scriptsize$\pm$0.031} & 0.973{\scriptsize$\pm$0.008} & 0.862{\scriptsize$\pm$0.022} & 0.524{\scriptsize$\pm$0.053} & 0.624{\scriptsize$\pm$0.048} \\
\midrule
Wiki & 50 & 40.362{\scriptsize$\pm$0.601} & 0.290{\scriptsize$\pm$0.004} & 0.952{\scriptsize$\pm$0.002} & 0.862{\scriptsize$\pm$0.002} & 0.863{\scriptsize$\pm$0.021} & 0.948{\scriptsize$\pm$0.003} \\
 & 200 & 45.966{\scriptsize$\pm$2.247} & 0.232{\scriptsize$\pm$0.011} & 0.986{\scriptsize$\pm$0.002} & 0.914{\scriptsize$\pm$0.007} & 0.890{\scriptsize$\pm$0.010} & 0.897{\scriptsize$\pm$0.009} \\
 & 1000 & 52.555{\scriptsize$\pm$2.817} & 0.229{\scriptsize$\pm$0.012} & 0.986{\scriptsize$\pm$0.001} & 0.903{\scriptsize$\pm$0.003} & 0.865{\scriptsize$\pm$0.004} & 0.880{\scriptsize$\pm$0.002} \\
\midrule
BlogCatalog & 50 & 905.177{\scriptsize$\pm$35.666} & 0.414{\scriptsize$\pm$0.016} & 0.918{\scriptsize$\pm$0.015} & 0.802{\scriptsize$\pm$0.018} & 0.830{\scriptsize$\pm$0.010} & 0.913{\scriptsize$\pm$0.010} \\
 & 200 & 597.536{\scriptsize$\pm$66.618} & 0.205{\scriptsize$\pm$0.023} & 0.983{\scriptsize$\pm$0.001} & 0.911{\scriptsize$\pm$0.005} & 0.871{\scriptsize$\pm$0.016} & 0.891{\scriptsize$\pm$0.015} \\
 & 1000 & 735.942{\scriptsize$\pm$17.629} & 0.256{\scriptsize$\pm$0.006} & 0.971{\scriptsize$\pm$0.002} & 0.863{\scriptsize$\pm$0.004} & 0.726{\scriptsize$\pm$0.012} & 0.778{\scriptsize$\pm$0.010} \\
\midrule
Flickr & 50 & 631.294{\scriptsize$\pm$58.721} & 0.288{\scriptsize$\pm$0.027} & 0.961{\scriptsize$\pm$0.010} & 0.874{\scriptsize$\pm$0.019} & 0.877{\scriptsize$\pm$0.023} & 0.940{\scriptsize$\pm$0.005} \\
 & 200 & 732.234{\scriptsize$\pm$44.384} & 0.248{\scriptsize$\pm$0.015} & 0.977{\scriptsize$\pm$0.008} & 0.891{\scriptsize$\pm$0.016} & 0.843{\scriptsize$\pm$0.016} & 0.887{\scriptsize$\pm$0.018} \\
 & 1000 & 875.688{\scriptsize$\pm$32.976} & 0.236{\scriptsize$\pm$0.009} & 0.974{\scriptsize$\pm$0.001} & 0.869{\scriptsize$\pm$0.001} & 0.729{\scriptsize$\pm$0.013} & 0.782{\scriptsize$\pm$0.007} \\
\midrule
Amazon C. & 50 & 684.999{\scriptsize$\pm$13.863} & 0.372{\scriptsize$\pm$0.008} & 0.932{\scriptsize$\pm$0.016} & 0.824{\scriptsize$\pm$0.014} & 0.843{\scriptsize$\pm$0.015} & 0.920{\scriptsize$\pm$0.013} \\
 & 200 & 975.351{\scriptsize$\pm$120.121} & 0.241{\scriptsize$\pm$0.030} & 0.979{\scriptsize$\pm$0.002} & 0.895{\scriptsize$\pm$0.004} & 0.852{\scriptsize$\pm$0.010} & 0.893{\scriptsize$\pm$0.006} \\
 & 1000 & 1213.036{\scriptsize$\pm$30.118} & 0.291{\scriptsize$\pm$0.007} & 0.957{\scriptsize$\pm$0.000} & 0.827{\scriptsize$\pm$0.001} & 0.723{\scriptsize$\pm$0.008} & 0.757{\scriptsize$\pm$0.008} \\
\midrule
ogbn-arxiv & 50 & 4544.548{\scriptsize$\pm$180.218} & 0.352{\scriptsize$\pm$0.014} & 0.896{\scriptsize$\pm$0.013} & 0.795{\scriptsize$\pm$0.009} & 0.877{\scriptsize$\pm$0.015} & 0.897{\scriptsize$\pm$0.003} \\
 & 200 & 6556.686{\scriptsize$\pm$316.644} & 0.342{\scriptsize$\pm$0.017} & 0.956{\scriptsize$\pm$0.006} & 0.831{\scriptsize$\pm$0.010} & 0.713{\scriptsize$\pm$0.015} & 0.778{\scriptsize$\pm$0.008} \\
 & 1000 & $>$10,000 & 0.664{\scriptsize$\pm$0.017} & 0.737{\scriptsize$\pm$0.023} & 0.555{\scriptsize$\pm$0.020} & 0.551{\scriptsize$\pm$0.013} & 0.510{\scriptsize$\pm$0.008} \\
\midrule
com-DBLP & 50 & 4221.987{\scriptsize$\pm$90.587} & 0.325{\scriptsize$\pm$0.007} & 0.922{\scriptsize$\pm$0.002} & 0.799{\scriptsize$\pm$0.002} & 0.893{\scriptsize$\pm$0.021} & 0.892{\scriptsize$\pm$0.003} \\
 & 200 & 6579.477{\scriptsize$\pm$185.073} & 0.355{\scriptsize$\pm$0.010} & 0.925{\scriptsize$\pm$0.011} & 0.793{\scriptsize$\pm$0.013} & 0.730{\scriptsize$\pm$0.015} & 0.757{\scriptsize$\pm$0.005} \\
 & 1000 & $>$10,000 & 0.638{\scriptsize$\pm$0.012} & 0.711{\scriptsize$\pm$0.012} & 0.534{\scriptsize$\pm$0.011} & 0.469{\scriptsize$\pm$0.015} & 0.434{\scriptsize$\pm$0.007} \\
\midrule
com-Amazon & 50 & 4317.737{\scriptsize$\pm$93.764} & 0.326{\scriptsize$\pm$0.007} & 0.963{\scriptsize$\pm$0.004} & 0.865{\scriptsize$\pm$0.007} & 0.920{\scriptsize$\pm$0.020} & 0.922{\scriptsize$\pm$0.010} \\
 & 200 & 6480.177{\scriptsize$\pm$213.594} & 0.313{\scriptsize$\pm$0.010} & 0.957{\scriptsize$\pm$0.003} & 0.834{\scriptsize$\pm$0.008} & 0.744{\scriptsize$\pm$0.020} & 0.805{\scriptsize$\pm$0.007} \\
 & 1000 & $>$10,000 & 0.576{\scriptsize$\pm$0.022} & 0.793{\scriptsize$\pm$0.017} & 0.610{\scriptsize$\pm$0.017} & 0.625{\scriptsize$\pm$0.033} & 0.588{\scriptsize$\pm$0.037} \\
\bottomrule
\end{tabular}
\end{sidewaystable}

\Cref{tab:app_large} confirms the results of \Cref{tab:large_main} with all metrics and their standard deviations. \method{} achieves the best ranking metrics on all four datasets and the lowest RAE on all datasets except Cora and PubMed, where G2R has a slightly lower RAE but lower ranking metrics. The results of \method{} are also stable across seeds, while several baselines vary strongly between seeds or predict a constant value in all seeds (shown as - for $\rho$ and $\tau$). On Cora, GMN, GREED, and GraSP also achieve rank correlations above 0.9, but among the baselines, only G2R does so on more than one dataset, and it also fails on ogbn-arxiv.

\Cref{tab:app_large_all} shows how the results of \method{} change with the collection size $M$. From $M = 50$ to $M = 200$, the rank correlation improves on most datasets, likely because the number of training pairs grows with the square of $M$. At $M = 1000$, $\rho$ remains above 0.95 on all seven datasets derived from the smaller graphs, while P@$k$ decreases, which is expected since the reference set grows with $M$ (\Cref{app:evaluation}). On the three largest datasets, the results at $M = 1000$ are clearly lower. Here, the fixed training budget of 6 hours is not sufficient. At $M = 1000$, validation alone takes about 45\% of the training time on these datasets (\Cref{tab:app_cost}). These results therefore likely reflect the training budget and the configuration selected at $M = 200$, which is not tuned for $M = 1000$, rather than a limit of the model.

\subsection{Training Cost}
\label{app:cost}

\Cref{tab:app_cost} reports the training cost of \method{} for every dataset. It includes the wall-clock time until the best validation step is reached, the total training time, the time per training step, the share of training time spent on validation, and the peak GPU memory. Total times can exceed the 6-hour limit by about one minute because the last validation run is completed after the limit is reached.

\begin{table}[htbp]
\footnotesize
\centering
\setlength{\tabcolsep}{6pt}
\caption{Training cost of \method. Time columns are the median over seeds of the wall-clock time per run. \textbf{to best}: time until the best validation objective, in minutes; \textbf{total}: wall-clock time of the whole training, in minutes; \textbf{ms/step}: time per training step in milliseconds; \textbf{Val. (\%)}: share of validation in the total training time; \textbf{Peak GPU}: maximum over seeds of the peak GPU memory during training, in GB.}
\label{tab:app_cost}
\begin{tabular}{llccccc}
\toprule
Dataset & Task/$M$ & to best & total & ms/step & Val. (\%) & Peak GPU\\
\midrule
\multirow{2}{*}{AIDS700nef} & GED & 6.3 & 9.5 & 18.0 & 5.0 & 0.18 \\
 & MCS & 5.8 & 9.2 & 19.3 & 4.7 & 0.21 \\
\midrule
\multirow{2}{*}{IMDBMulti} & GED & 16.0 & 22.3 & 35.4 & 6.4 & 0.29 \\
 & MCS & 12.6 & 18.0 & 30.4 & 7.1 & 0.27 \\
\midrule
\multirow{2}{*}{LINUX} & GED & 10.0 & 15.6 & 31.0 & 4.4 & 0.15 \\
 & MCS & 15.6 & 21.8 & 35.3 & 3.8 & 0.19 \\
\midrule
\multirow{2}{*}{PTC} & GED & 4.2 & 7.4 & 18.5 & 2.3 & 0.41 \\
 & MCS & 5.2 & 9.5 & 24.6 & 1.8 & 0.64 \\
\midrule
\multirow{3}{*}{Cora} & 50 & 13.7 & 55.3 & 167.8 & 0.2 & 6.08 \\
 & 200 & 84.1 & 120.3 & 218.7 & 0.5 & 8.72 \\
 & 1000 & 70.2 & 109.5 & 231.1 & 2.4 & 8.72 \\
\midrule
\multirow{3}{*}{CiteSeer} & 50 & 27.9 & 58.9 & 175.5 & 0.2 & 8.25 \\
 & 200 & 40.2 & 69.2 & 174.0 & 0.6 & 9.54 \\
 & 1000 & 36.2 & 69.0 & 189.8 & 2.7 & 9.54 \\
\midrule
\multirow{3}{*}{PubMed} & 50 & 79.3 & 106.4 & 169.9 & 0.7 & 8.31 \\
 & 200 & 121.6 & 153.5 & 185.7 & 2.7 & 8.27 \\
 & 1000 & 201.9 & 238.5 & 191.1 & 11.9 & 8.30 \\
\midrule
\multirow{3}{*}{Wiki} & 50 & 10.9 & 50.9 & 154.2 & 0.1 & 4.81 \\
 & 200 & 49.8 & 82.0 & 196.0 & 0.4 & 7.74 \\
 & 1000 & 109.7 & 148.9 & 230.6 & 1.9 & 7.72 \\
\midrule
\multirow{3}{*}{BlogCatalog} & 50 & 14.3 & 122.4 & 371.0 & 0.2 & 10.85\\
 & 200 & 87.9 & 169.4 & 502.3 & 0.7 & 10.86\\
 & 1000 & 301.6 & 360.1 & 560.0 & 3.3 & 10.88\\
\midrule
\multirow{3}{*}{Flickr} & 50 & 33.6 & 187.8 & 569.2 & 0.3 & 10.84\\
 & 200 & 226.5 & 318.4 & 564.4 & 1.0 & 10.85\\
 & 1000 & 351.4 & 360.1 & 550.6 & 5.0 & 10.84\\
\midrule
\multirow{3}{*}{Amazon C.} & 50 & 101.7 & 156.7 & 328.2 & 0.5 & 8.88 \\
 & 200 & 105.9 & 165.4 & 348.2 & 1.8 & 8.87 \\
 & 1000 & 246.1 & 310.6 & 353.5 & 8.5 & 8.87 \\
\midrule
\multirow{3}{*}{ogbn-arxiv} & 50 & 75.9 & 110.0 & 195.9 & 3.9 & 10.46\\
 & 200 & 267.6 & 305.5 & 249.3 & 11.3 & 10.50\\
 & 1000 & 359.9 & 360.7 & 194.9 & 45.2 & 10.51\\
\midrule
\multirow{3}{*}{com-DBLP} & 50 & 163.3 & 215.0 & 298.3 & 3.9 & 19.70\\
 & 200 & 350.4 & 360.2 & 298.4 & 14.0 & 19.79\\
 & 1000 & 356.0 & 361.0 & 300.9 & 45.1 & 19.78\\
\midrule
\multirow{3}{*}{com-Amazon} & 50 & 141.3 & 188.3 & 272.6 & 4.1 & 19.90\\
 & 200 & 330.3 & 360.2 & 272.0 & 14.4 & 19.90\\
 & 1000 & 344.8 & 361.0 & 272.9 & 46.0 & 19.93\\
\bottomrule
\end{tabular}
\end{table}

%% file: main.bbl
\begin{thebibliography}{}
\renewcommand{\doi}[1]{\url{https://doi.org/#1}}
\bibcommenthead

\bibitem [\protect \citeauthoryear {%
Akiba%
, Sano%
, Yanase%
, Ohta%
\BCBL {}\ \BBA {} Koyama%
}{%
Akiba%
\ \protect \BOthers {.}}{%
{\protect \APACyear {2019}}%
}]{%
optuna}
\APACinsertmetastar {%
optuna}%
\begin{APACrefauthors}%
Akiba, T.%
, Sano, S.%
, Yanase, T.%
, Ohta, T.%
\BCBL {} Koyama, M.%
\end{APACrefauthors}%
\unskip\
\newblock
\APACrefYearMonthDay{2019}{}{}.
\newblock
{\BBOQ}\APACrefatitle {{Optuna}: A next-generation hyperparameter optimization framework} {{Optuna}: A next-generation hyperparameter optimization framework}.{\BBCQ}
\newblock
 \APACrefbtitle {{ACM SIGKDD International Conference on Knowledge Discovery and Data Mining}} {{ACM SIGKDD International Conference on Knowledge Discovery and Data Mining}}\ (\BPGS\ 2623--2631).
\PrintBackRefs{\CurrentBib}

\bibitem [\protect \citeauthoryear {%
J.~Bai%
\ \BBA {} Zhao%
}{%
J.~Bai%
\ \BBA {} Zhao%
}{%
{\protect \APACyear {2021}}%
}]{%
TAGSim}
\APACinsertmetastar {%
TAGSim}%
\begin{APACrefauthors}%
Bai, J.%
\BCBT {}\ \BBA {} Zhao, P.%
\end{APACrefauthors}%
\unskip\
\newblock
\APACrefYearMonthDay{2021}{}{}.
\newblock
{\BBOQ}\APACrefatitle {{TaGSim}: Type-aware graph similarity learning and computation} {{TaGSim}: Type-aware graph similarity learning and computation}.{\BBCQ}
\newblock
\APACjournalVolNumPages{Proceedings of the VLDB Endowment}{15}{2}{335--347,}
\newblock
\begin{APACrefDOI} \doi{10.14778/3489496.3489513} \end{APACrefDOI}
\newblock

\newblock

\PrintBackRefs{\CurrentBib}

\bibitem [\protect \citeauthoryear {%
Y.~Bai%
\ \protect \BOthers {.}}{%
Y.~Bai%
\ \protect \BOthers {.}}{%
{\protect \APACyear {2019}}%
}]{%
SimGNN}
\APACinsertmetastar {%
SimGNN}%
\begin{APACrefauthors}%
Bai, Y.%
, Ding, H.%
, Bian, S.%
, Chen, T.%
, Sun, Y.%
\BCBL {} Wang, W.%
\end{APACrefauthors}%
\unskip\
\newblock
\APACrefYearMonthDay{2019}{}{}.
\newblock
{\BBOQ}\APACrefatitle {{SimGNN}: A neural network approach to fast graph similarity computation} {{SimGNN}: A neural network approach to fast graph similarity computation}.{\BBCQ}
\newblock
 \APACrefbtitle {{ACM International Conference on Web Search and Data Mining}} {{ACM International Conference on Web Search and Data Mining}}\ (\BPGS\ 384--392).
\PrintBackRefs{\CurrentBib}

\bibitem [\protect \citeauthoryear {%
Y.~Bai%
, Ding%
, Gu%
, Sun%
\BCBL {}\ \BBA {} Wang%
}{%
Y.~Bai%
\ \protect \BOthers {.}}{%
{\protect \APACyear {2020}}%
}]{%
GraphSim}
\APACinsertmetastar {%
GraphSim}%
\begin{APACrefauthors}%
Bai, Y.%
, Ding, H.%
, Gu, K.%
, Sun, Y.%
\BCBL {} Wang, W.%
\end{APACrefauthors}%
\unskip\
\newblock
\APACrefYearMonthDay{2020}{}{}.
\newblock
{\BBOQ}\APACrefatitle {Learning-based efficient graph similarity computation via multi-scale convolutional set matching} {Learning-based efficient graph similarity computation via multi-scale convolutional set matching}.{\BBCQ}
\newblock
 \APACrefbtitle {{AAAI Conference on Artificial Intelligence}} {{AAAI Conference on Artificial Intelligence}}\ (\BVOL~34, \BPGS\ 3219--3226).
\PrintBackRefs{\CurrentBib}

\bibitem [\protect \citeauthoryear {%
Y.~Bai%
, Xu%
, Sun%
\BCBL {}\ \BBA {} Wang%
}{%
Y.~Bai%
\ \protect \BOthers {.}}{%
{\protect \APACyear {2021}}%
}]{%
GLSearch}
\APACinsertmetastar {%
GLSearch}%
\begin{APACrefauthors}%
Bai, Y.%
, Xu, D.%
, Sun, Y.%
\BCBL {} Wang, W.%
\end{APACrefauthors}%
\unskip\
\newblock
\APACrefYearMonthDay{2021}{}{}.
\newblock
{\BBOQ}\APACrefatitle {{GLSearch}: Maximum common subgraph detection via learning to search} {{GLSearch}: Maximum common subgraph detection via learning to search}.{\BBCQ}
\newblock
 \APACrefbtitle {{International Conference on Machine Learning}} {{International Conference on Machine Learning}}\ (\BVOL~139, \BPGS\ 588--598).
\PrintBackRefs{\CurrentBib}

\bibitem [\protect \citeauthoryear {%
Blumenthal%
\ \BBA {} Gamper%
}{%
Blumenthal%
\ \BBA {} Gamper%
}{%
{\protect \APACyear {2020}}%
}]{%
astar_scale}
\APACinsertmetastar {%
astar_scale}%
\begin{APACrefauthors}%
Blumenthal, D.B.%
\BCBT {}\ \BBA {} Gamper, J.%
\end{APACrefauthors}%
\unskip\
\newblock
\APACrefYearMonthDay{2020}{}{}.
\newblock
{\BBOQ}\APACrefatitle {On the exact computation of the graph edit distance} {On the exact computation of the graph edit distance}.{\BBCQ}
\newblock
\APACjournalVolNumPages{Pattern Recognition Letters}{134}{}{46--57,}
\newblock
\begin{APACrefDOI} \doi{10.1016/j.patrec.2018.05.002} \end{APACrefDOI}
\newblock

\newblock

\PrintBackRefs{\CurrentBib}

\bibitem [\protect \citeauthoryear {%
Brockschmidt%
}{%
Brockschmidt%
}{%
{\protect \APACyear {2020}}%
}]{%
GNNFiLM}
\APACinsertmetastar {%
GNNFiLM}%
\begin{APACrefauthors}%
Brockschmidt, M.%
\end{APACrefauthors}%
\unskip\
\newblock
\APACrefYearMonthDay{2020}{}{}.
\newblock
{\BBOQ}\APACrefatitle {{GNN-FiLM}: Graph neural networks with feature-wise linear modulation} {{GNN-FiLM}: Graph neural networks with feature-wise linear modulation}.{\BBCQ}
\newblock
 \APACrefbtitle {{International Conference on Machine Learning}} {{International Conference on Machine Learning}}\ (\BPGS\ 1144--1152).
\PrintBackRefs{\CurrentBib}

\bibitem [\protect \citeauthoryear {%
Bromley%
, Guyon%
, LeCun%
, S{\"a}ckinger%
\BCBL {}\ \BBA {} Shah%
}{%
Bromley%
\ \protect \BOthers {.}}{%
{\protect \APACyear {1993}}%
}]{%
siamese}
\APACinsertmetastar {%
siamese}%
\begin{APACrefauthors}%
Bromley, J.%
, Guyon, I.%
, LeCun, Y.%
, S{\"a}ckinger, E.%
\BCBL {} Shah, R.%
\end{APACrefauthors}%
\unskip\
\newblock
\APACrefYearMonthDay{1993}{}{}.
\newblock
{\BBOQ}\APACrefatitle {Signature verification using a ``{Siamese}'' time delay neural network} {Signature verification using a ``{Siamese}'' time delay neural network}.{\BBCQ}
\newblock
 \APACrefbtitle {{Advances in Neural Information Processing Systems}} {{Advances in Neural Information Processing Systems}}\ (\BVOL~6, \BPGS\ 737--744).
\PrintBackRefs{\CurrentBib}

\bibitem [\protect \citeauthoryear {%
Bunke%
\ \BBA {} Allermann%
}{%
Bunke%
\ \BBA {} Allermann%
}{%
{\protect \APACyear {1983}}%
}]{%
ged}
\APACinsertmetastar {%
ged}%
\begin{APACrefauthors}%
Bunke, H.%
\BCBT {}\ \BBA {} Allermann, G.%
\end{APACrefauthors}%
\unskip\
\newblock
\APACrefYearMonthDay{1983}{}{}.
\newblock
{\BBOQ}\APACrefatitle {Inexact graph matching for structural pattern recognition} {Inexact graph matching for structural pattern recognition}.{\BBCQ}
\newblock
\APACjournalVolNumPages{Pattern Recognition Letters}{1}{4}{245--253,}
\newblock
\begin{APACrefDOI} \doi{10.1016/0167-8655(83)90033-8} \end{APACrefDOI}
\newblock

\newblock

\PrintBackRefs{\CurrentBib}

\bibitem [\protect \citeauthoryear {%
Bunke%
\ \BBA {} Shearer%
}{%
Bunke%
\ \BBA {} Shearer%
}{%
{\protect \APACyear {1998}}%
}]{%
mcs}
\APACinsertmetastar {%
mcs}%
\begin{APACrefauthors}%
Bunke, H.%
\BCBT {}\ \BBA {} Shearer, K.%
\end{APACrefauthors}%
\unskip\
\newblock
\APACrefYearMonthDay{1998}{}{}.
\newblock
{\BBOQ}\APACrefatitle {A graph distance metric based on the maximal common subgraph} {A graph distance metric based on the maximal common subgraph}.{\BBCQ}
\newblock
\APACjournalVolNumPages{Pattern Recognition Letters}{19}{3--4}{255--259,}
\newblock
\begin{APACrefDOI} \doi{10.1016/S0167-8655(97)00179-7} \end{APACrefDOI}
\newblock

\newblock

\PrintBackRefs{\CurrentBib}

\bibitem [\protect \citeauthoryear {%
Cai%
\ \BBA {} Wang%
}{%
Cai%
\ \BBA {} Wang%
}{%
{\protect \APACyear {2018}}%
}]{%
LDP}
\APACinsertmetastar {%
LDP}%
\begin{APACrefauthors}%
Cai, C.%
\BCBT {}\ \BBA {} Wang, Y.%
\end{APACrefauthors}%
\unskip\
\newblock
\APACrefYearMonthDay{2018}{}{}.
\newblock
\APACrefbtitle {A simple yet effective baseline for non-attributed graph classification.} {A simple yet effective baseline for non-attributed graph classification.}
\newblock
\APAChowpublished {arXiv preprint 1811.03508}.
\PrintBackRefs{\CurrentBib}

\bibitem [\protect \citeauthoryear {%
Cheng%
, Yan%
, Wu%
, Huang%
\BCBL {}\ \BBA {} Zhang%
}{%
Cheng%
\ \protect \BOthers {.}}{%
{\protect \APACyear {2025}}%
}]{%
GEDIOT}
\APACinsertmetastar {%
GEDIOT}%
\begin{APACrefauthors}%
Cheng, Q.%
, Yan, D.%
, Wu, T.%
, Huang, Z.%
\BCBL {} Zhang, Q.%
\end{APACrefauthors}%
\unskip\
\newblock
\APACrefYearMonthDay{2025}{}{}.
\newblock
{\BBOQ}\APACrefatitle {Computing approximate graph edit distance via optimal transport} {Computing approximate graph edit distance via optimal transport}.{\BBCQ}
\newblock
\APACjournalVolNumPages{Proceedings of the ACM on Management of Data}{3}{1}{1--26,}
\newblock
\begin{APACrefDOI} \doi{10.1145/3709673} \end{APACrefDOI}
\newblock

\newblock

\PrintBackRefs{\CurrentBib}

\bibitem [\protect \citeauthoryear {%
Cho%
\ \protect \BOthers {.}}{%
Cho%
\ \protect \BOthers {.}}{%
{\protect \APACyear {2014}}%
}]{%
GRU}
\APACinsertmetastar {%
GRU}%
\begin{APACrefauthors}%
Cho, K.%
, van Merri{\"e}nboer, B.%
, Gulcehre, C.%
, Bahdanau, D.%
, Bougares, F.%
, Schwenk, H.%
\BCBL {} Bengio, Y.%
\end{APACrefauthors}%
\unskip\
\newblock
\APACrefYearMonthDay{2014}{}{}.
\newblock
{\BBOQ}\APACrefatitle {Learning phrase representations using {RNN} encoder--decoder for statistical machine translation} {Learning phrase representations using {RNN} encoder--decoder for statistical machine translation}.{\BBCQ}
\newblock
 \APACrefbtitle {{Conference on Empirical Methods in Natural Language Processing}} {{Conference on Empirical Methods in Natural Language Processing}}\ (\BPGS\ 1724--1734).
\PrintBackRefs{\CurrentBib}

\bibitem [\protect \citeauthoryear {%
Dwivedi%
\ \protect \BOthers {.}}{%
Dwivedi%
\ \protect \BOthers {.}}{%
{\protect \APACyear {2023}}%
}]{%
DwivediBenchmark}
\APACinsertmetastar {%
DwivediBenchmark}%
\begin{APACrefauthors}%
Dwivedi, V.P.%
, Joshi, C.K.%
, Luu, A.T.%
, Laurent, T.%
, Bengio, Y.%
\BCBL {} Bresson, X.%
\end{APACrefauthors}%
\unskip\
\newblock
\APACrefYearMonthDay{2023}{}{}.
\newblock
{\BBOQ}\APACrefatitle {Benchmarking graph neural networks} {Benchmarking graph neural networks}.{\BBCQ}
\newblock
\APACjournalVolNumPages{Journal of Machine Learning Research}{24}{43}{1--48,}
\newblock

\newblock

\PrintBackRefs{\CurrentBib}

\bibitem [\protect \citeauthoryear {%
Dwivedi%
, Luu%
, Laurent%
, Bengio%
\BCBL {}\ \BBA {} Bresson%
}{%
Dwivedi%
\ \protect \BOthers {.}}{%
{\protect \APACyear {2022}}%
}]{%
RWPE}
\APACinsertmetastar {%
RWPE}%
\begin{APACrefauthors}%
Dwivedi, V.P.%
, Luu, A.T.%
, Laurent, T.%
, Bengio, Y.%
\BCBL {} Bresson, X.%
\end{APACrefauthors}%
\unskip\
\newblock
\APACrefYearMonthDay{2022}{}{}.
\newblock
{\BBOQ}\APACrefatitle {Graph neural networks with learnable structural and positional representations} {Graph neural networks with learnable structural and positional representations}.{\BBCQ}
\newblock
 \APACrefbtitle {{International Conference on Learning Representations}.} {{International Conference on Learning Representations}.}
\PrintBackRefs{\CurrentBib}

\bibitem [\protect \citeauthoryear {%
Fankhauser%
, Riesen%
\BCBL {}\ \BBA {} Bunke%
}{%
Fankhauser%
\ \protect \BOthers {.}}{%
{\protect \APACyear {2011}}%
}]{%
VJ}
\APACinsertmetastar {%
VJ}%
\begin{APACrefauthors}%
Fankhauser, S.%
, Riesen, K.%
\BCBL {} Bunke, H.%
\end{APACrefauthors}%
\unskip\
\newblock
\APACrefYearMonthDay{2011}{}{}.
\newblock
{\BBOQ}\APACrefatitle {Speeding up graph edit distance computation through fast bipartite matching} {Speeding up graph edit distance computation through fast bipartite matching}.{\BBCQ}
\newblock
 \APACrefbtitle {{International Workshop on Graph-Based Representations in Pattern Recognition}} {{International Workshop on Graph-Based Representations in Pattern Recognition}}\ (\BPGS\ 102--111).
\PrintBackRefs{\CurrentBib}

\bibitem [\protect \citeauthoryear {%
Fey%
\ \BBA {} Lenssen%
}{%
Fey%
\ \BBA {} Lenssen%
}{%
{\protect \APACyear {2019}}%
}]{%
pyg}
\APACinsertmetastar {%
pyg}%
\begin{APACrefauthors}%
Fey, M.%
\BCBT {}\ \BBA {} Lenssen, J.E.%
\end{APACrefauthors}%
\unskip\
\newblock
\APACrefYearMonthDay{2019}{}{}.
\newblock
{\BBOQ}\APACrefatitle {Fast graph representation learning with {PyTorch Geometric}} {Fast graph representation learning with {PyTorch Geometric}}.{\BBCQ}
\newblock
 \APACrefbtitle {{International Conference on Learning Representations Workshop on Representation Learning on Graphs and Manifolds}.} {{International Conference on Learning Representations Workshop on Representation Learning on Graphs and Manifolds}.}
\PrintBackRefs{\CurrentBib}

\bibitem [\protect \citeauthoryear {%
Gori%
, Monfardini%
\BCBL {}\ \BBA {} Scarselli%
}{%
Gori%
\ \protect \BOthers {.}}{%
{\protect \APACyear {2005}}%
}]{%
gnn_1}
\APACinsertmetastar {%
gnn_1}%
\begin{APACrefauthors}%
Gori, M.%
, Monfardini, G.%
\BCBL {} Scarselli, F.%
\end{APACrefauthors}%
\unskip\
\newblock
\APACrefYearMonthDay{2005}{}{}.
\newblock
{\BBOQ}\APACrefatitle {A new model for learning in graph domains} {A new model for learning in graph domains}.{\BBCQ}
\newblock
 \APACrefbtitle {{IEEE International Joint Conference on Neural Networks}} {{IEEE International Joint Conference on Neural Networks}}\ (\BVOL~2, \BPGS\ 729--734).
\PrintBackRefs{\CurrentBib}

\bibitem [\protect \citeauthoryear {%
Hart%
, Nilsson%
\BCBL {}\ \BBA {} Raphael%
}{%
Hart%
\ \protect \BOthers {.}}{%
{\protect \APACyear {1968}}%
}]{%
astar}
\APACinsertmetastar {%
astar}%
\begin{APACrefauthors}%
Hart, P.E.%
, Nilsson, N.J.%
\BCBL {} Raphael, B.%
\end{APACrefauthors}%
\unskip\
\newblock
\APACrefYearMonthDay{1968}{}{}.
\newblock
{\BBOQ}\APACrefatitle {A formal basis for the heuristic determination of minimum cost paths} {A formal basis for the heuristic determination of minimum cost paths}.{\BBCQ}
\newblock
\APACjournalVolNumPages{IEEE Transactions on Systems Science and Cybernetics}{4}{2}{100--107,}
\newblock
\begin{APACrefDOI} \doi{10.1109/TSSC.1968.300136} \end{APACrefDOI}
\newblock

\newblock

\PrintBackRefs{\CurrentBib}

\bibitem [\protect \citeauthoryear {%
He%
, Zhang%
, Ren%
\BCBL {}\ \BBA {} Sun%
}{%
He%
\ \protect \BOthers {.}}{%
{\protect \APACyear {2016}}%
}]{%
he2016deep}
\APACinsertmetastar {%
he2016deep}%
\begin{APACrefauthors}%
He, K.%
, Zhang, X.%
, Ren, S.%
\BCBL {} Sun, J.%
\end{APACrefauthors}%
\unskip\
\newblock
\APACrefYearMonthDay{2016}{}{}.
\newblock
{\BBOQ}\APACrefatitle {Deep residual learning for image recognition} {Deep residual learning for image recognition}.{\BBCQ}
\newblock
 \APACrefbtitle {{IEEE Conference on Computer Vision and Pattern Recognition}} {{IEEE Conference on Computer Vision and Pattern Recognition}}\ (\BPGS\ 770--778).
\PrintBackRefs{\CurrentBib}

\bibitem [\protect \citeauthoryear {%
Hu%
\ \protect \BOthers {.}}{%
Hu%
\ \protect \BOthers {.}}{%
{\protect \APACyear {2020}}%
}]{%
ogb_datasets}
\APACinsertmetastar {%
ogb_datasets}%
\begin{APACrefauthors}%
Hu, W.%
, Fey, M.%
, Zitnik, M.%
, Dong, Y.%
, Ren, H.%
, Liu, B.%
\BDBL {}Leskovec, J.%
\end{APACrefauthors}%
\unskip\
\newblock
\APACrefYearMonthDay{2020}{}{}.
\newblock
{\BBOQ}\APACrefatitle {{Open Graph Benchmark}: Datasets for machine learning on graphs} {{Open Graph Benchmark}: Datasets for machine learning on graphs}.{\BBCQ}
\newblock
 \APACrefbtitle {{Advances in Neural Information Processing Systems}.} {{Advances in Neural Information Processing Systems}.}
\PrintBackRefs{\CurrentBib}

\bibitem [\protect \citeauthoryear {%
Jain%
, Roy%
, Meher%
, Chakrabarti%
\BCBL {}\ \BBA {} De%
}{%
Jain%
\ \protect \BOthers {.}}{%
{\protect \APACyear {2024}}%
}]{%
GraphEdX}
\APACinsertmetastar {%
GraphEdX}%
\begin{APACrefauthors}%
Jain, E.%
, Roy, I.%
, Meher, S.%
, Chakrabarti, S.%
\BCBL {} De, A.%
\end{APACrefauthors}%
\unskip\
\newblock
\APACrefYearMonthDay{2024}{}{}.
\newblock
{\BBOQ}\APACrefatitle {Graph edit distance with general costs using neural set divergence} {Graph edit distance with general costs using neural set divergence}.{\BBCQ}
\newblock
 \APACrefbtitle {{Advances in Neural Information Processing Systems}} {{Advances in Neural Information Processing Systems}}\ (\BVOL~37, \BPGS\ 73399--73438).
\PrintBackRefs{\CurrentBib}

\bibitem [\protect \citeauthoryear {%
Jiang%
, Ning%
\BCBL {}\ \BBA {} Dong%
}{%
Jiang%
\ \protect \BOthers {.}}{%
{\protect \APACyear {2023}}%
}]{%
survey_scale}
\APACinsertmetastar {%
survey_scale}%
\begin{APACrefauthors}%
Jiang, N.%
, Ning, B.%
\BCBL {} Dong, J.%
\end{APACrefauthors}%
\unskip\
\newblock
\APACrefYearMonthDay{2023}{}{}.
\newblock

\newblock
 \APACrefbtitle {{International Conference on Image, Vision and Computing}, title={A survey of {GNN}-based graph similarity learning}} {{International Conference on Image, Vision and Computing}, title={A survey of {GNN}-based graph similarity learning}}\ (\BPGS\ 650--654).
\PrintBackRefs{\CurrentBib}

\bibitem [\protect \citeauthoryear {%
Kendall%
}{%
Kendall%
}{%
{\protect \APACyear {1938}}%
}]{%
kendall}
\APACinsertmetastar {%
kendall}%
\begin{APACrefauthors}%
Kendall, M.G.%
\end{APACrefauthors}%
\unskip\
\newblock
\APACrefYearMonthDay{1938}{}{}.
\newblock
{\BBOQ}\APACrefatitle {A new measure of rank correlation} {A new measure of rank correlation}.{\BBCQ}
\newblock
\APACjournalVolNumPages{Biometrika}{30}{1--2}{81--93,}
\newblock
\begin{APACrefDOI} \doi{10.1093/biomet/30.1-2.81} \end{APACrefDOI}
\newblock

\newblock

\PrintBackRefs{\CurrentBib}

\bibitem [\protect \citeauthoryear {%
Kipf%
\ \BBA {} Welling%
}{%
Kipf%
\ \BBA {} Welling%
}{%
{\protect \APACyear {2017}}%
}]{%
GCN}
\APACinsertmetastar {%
GCN}%
\begin{APACrefauthors}%
Kipf, T.N.%
\BCBT {}\ \BBA {} Welling, M.%
\end{APACrefauthors}%
\unskip\
\newblock
\APACrefYearMonthDay{2017}{}{}.
\newblock
{\BBOQ}\APACrefatitle {Semi-supervised classification with graph convolutional networks} {Semi-supervised classification with graph convolutional networks}.{\BBCQ}
\newblock
 \APACrefbtitle {{International Conference on Learning Representations}.} {{International Conference on Learning Representations}.}
\PrintBackRefs{\CurrentBib}

\bibitem [\protect \citeauthoryear {%
Lan%
, Hong%
, Ma%
\BCBL {}\ \BBA {} Ma%
}{%
Lan%
\ \protect \BOthers {.}}{%
{\protect \APACyear {2024}}%
}]{%
INFMCS}
\APACinsertmetastar {%
INFMCS}%
\begin{APACrefauthors}%
Lan, Z.%
, Hong, B.%
, Ma, Y.%
\BCBL {} Ma, F.%
\end{APACrefauthors}%
\unskip\
\newblock
\APACrefYearMonthDay{2024}{}{}.
\newblock
{\BBOQ}\APACrefatitle {More interpretable graph similarity computation via maximum common subgraph inference} {More interpretable graph similarity computation via maximum common subgraph inference}.{\BBCQ}
\newblock
\APACjournalVolNumPages{IEEE Transactions on Knowledge and Data Engineering}{36}{11}{6588--6599,}
\newblock
\begin{APACrefDOI} \doi{10.1109/TKDE.2024.3387044} \end{APACrefDOI}
\newblock

\newblock

\PrintBackRefs{\CurrentBib}

\bibitem [\protect \citeauthoryear {%
Li%
, Gu%
, Dullien%
, Vinyals%
\BCBL {}\ \BBA {} Kohli%
}{%
Li%
\ \protect \BOthers {.}}{%
{\protect \APACyear {2019}}%
}]{%
GMN}
\APACinsertmetastar {%
GMN}%
\begin{APACrefauthors}%
Li, Y.%
, Gu, C.%
, Dullien, T.%
, Vinyals, O.%
\BCBL {} Kohli, P.%
\end{APACrefauthors}%
\unskip\
\newblock
\APACrefYearMonthDay{2019}{}{}.
\newblock
{\BBOQ}\APACrefatitle {Graph matching networks for learning the similarity of graph structured objects} {Graph matching networks for learning the similarity of graph structured objects}.{\BBCQ}
\newblock
 \APACrefbtitle {{International Conference on Machine Learning}} {{International Conference on Machine Learning}}\ (\BVOL~97, \BPGS\ 3835--3845).
\PrintBackRefs{\CurrentBib}

\bibitem [\protect \citeauthoryear {%
Li%
, Zemel%
, Brockschmidt%
\BCBL {}\ \BBA {} Tarlow%
}{%
Li%
\ \protect \BOthers {.}}{%
{\protect \APACyear {2016}}%
}]{%
GATED}
\APACinsertmetastar {%
GATED}%
\begin{APACrefauthors}%
Li, Y.%
, Zemel, R.%
, Brockschmidt, M.%
\BCBL {} Tarlow, D.%
\end{APACrefauthors}%
\unskip\
\newblock
\APACrefYearMonthDay{2016}{}{}.
\newblock
{\BBOQ}\APACrefatitle {Gated graph sequence neural networks} {Gated graph sequence neural networks}.{\BBCQ}
\newblock
 \APACrefbtitle {{International Conference on Learning Representations}.} {{International Conference on Learning Representations}.}
\PrintBackRefs{\CurrentBib}

\bibitem [\protect \citeauthoryear {%
Liu%
, Chen%
, Liu%
, He%
\BCBL {}\ \BBA {} Li%
}{%
Liu%
, Chen%
\BCBL {}\ \protect \BOthers {.}}{%
{\protect \APACyear {2025}}%
}]{%
G2R}
\APACinsertmetastar {%
G2R}%
\begin{APACrefauthors}%
Liu, Z.%
, Chen, Y.%
, Liu, N.%
, He, J.%
\BCBL {} Li, D.%
\end{APACrefauthors}%
\unskip\
\newblock
\APACrefYearMonthDay{2025}{}{}.
\newblock
{\BBOQ}\APACrefatitle {{Graph2Region}: Efficient graph similarity learning with structure and scale restoration} {{Graph2Region}: Efficient graph similarity learning with structure and scale restoration}.{\BBCQ}
\newblock
\APACjournalVolNumPages{IEEE Transactions on Knowledge and Data Engineering}{}{}{,}
\newblock
\begin{APACrefDOI} \doi{10.1109/TKDE.2025.3617461} \end{APACrefDOI}
\newblock

\newblock

\PrintBackRefs{\CurrentBib}

\bibitem [\protect \citeauthoryear {%
Liu%
\ \protect \BOthers {.}}{%
Liu%
\ \protect \BOthers {.}}{%
{\protect \APACyear {2026}}%
}]{%
GEN}
\APACinsertmetastar {%
GEN}%
\begin{APACrefauthors}%
Liu, Z.%
, Liu, N.%
, Chen, Y.%
, He, J.%
, Ma, S.%
\BCBL {} Li, D.%
\end{APACrefauthors}%
\unskip\
\newblock
\APACrefYearMonthDay{2026}{}{}.
\newblock
{\BBOQ}\APACrefatitle {Rethinking flexible graph similarity computation: One-step alignment with global guidance} {Rethinking flexible graph similarity computation: One-step alignment with global guidance}.{\BBCQ}
\newblock
 \APACrefbtitle {{IEEE International Conference on Data Engineering}} {{IEEE International Conference on Data Engineering}}\ (\BPGS\ 1435--1448).
\PrintBackRefs{\CurrentBib}

\bibitem [\protect \citeauthoryear {%
Liu%
, Liu%
\BCBL {}\ \protect \BOthers {.}}{%
Liu%
, Liu%
\BCBL {}\ \protect \BOthers {.}}{%
{\protect \APACyear {2025}}%
}]{%
LiuSurvey}
\APACinsertmetastar {%
LiuSurvey}%
\begin{APACrefauthors}%
Liu, Z.%
, Liu, N.%
, Chen, Y.%
, Wen, Z.%
, He, J.%
\BCBL {} Li, D.%
\end{APACrefauthors}%
\unskip\
\newblock
\APACrefYearMonthDay{2025}{}{}.
\newblock
{\BBOQ}\APACrefatitle {Graph theory-based deep graph similarity learning: A unified survey of pipeline, techniques, and challenges} {Graph theory-based deep graph similarity learning: A unified survey of pipeline, techniques, and challenges}.{\BBCQ}
\newblock
\APACjournalVolNumPages{Transactions on Machine Learning Research}{}{}{,}
\newblock

\newblock

\PrintBackRefs{\CurrentBib}

\bibitem [\protect \citeauthoryear {%
Ma%
, Ahmed%
, Willke%
\BCBL {}\ \BBA {} Yu%
}{%
Ma%
\ \protect \BOthers {.}}{%
{\protect \APACyear {2021}}%
}]{%
ma2021deep}
\APACinsertmetastar {%
ma2021deep}%
\begin{APACrefauthors}%
Ma, G.%
, Ahmed, N.K.%
, Willke, T.L.%
\BCBL {} Yu, P.S.%
\end{APACrefauthors}%
\unskip\
\newblock
\APACrefYearMonthDay{2021}{}{}.
\newblock
{\BBOQ}\APACrefatitle {Deep graph similarity learning: A survey} {Deep graph similarity learning: A survey}.{\BBCQ}
\newblock
\APACjournalVolNumPages{Data Mining and Knowledge Discovery}{35}{}{688--725,}
\newblock
\begin{APACrefDOI} \doi{10.1007/s10618-020-00733-5} \end{APACrefDOI}
\newblock

\newblock

\PrintBackRefs{\CurrentBib}

\bibitem [\protect \citeauthoryear {%
McCreesh%
, Prosser%
\BCBL {}\ \BBA {} Trimble%
}{%
McCreesh%
\ \protect \BOthers {.}}{%
{\protect \APACyear {2017}}%
}]{%
mcsplit}
\APACinsertmetastar {%
mcsplit}%
\begin{APACrefauthors}%
McCreesh, C.%
, Prosser, P.%
\BCBL {} Trimble, J.%
\end{APACrefauthors}%
\unskip\
\newblock
\APACrefYearMonthDay{2017}{}{}.
\newblock
{\BBOQ}\APACrefatitle {A partitioning algorithm for maximum common subgraph problems} {A partitioning algorithm for maximum common subgraph problems}.{\BBCQ}
\newblock
 \APACrefbtitle {{International Joint Conference on Artificial Intelligence}} {{International Joint Conference on Artificial Intelligence}}\ (\BPGS\ 712--719).
\PrintBackRefs{\CurrentBib}

\bibitem [\protect \citeauthoryear {%
Morris%
\ \protect \BOthers {.}}{%
Morris%
\ \protect \BOthers {.}}{%
{\protect \APACyear {2019}}%
}]{%
MorrisWL}
\APACinsertmetastar {%
MorrisWL}%
\begin{APACrefauthors}%
Morris, C.%
, Ritzert, M.%
, Fey, M.%
, Hamilton, W.L.%
, Lenssen, J.E.%
, Rattan, G.%
\BCBL {} Grohe, M.%
\end{APACrefauthors}%
\unskip\
\newblock
\APACrefYearMonthDay{2019}{}{}.
\newblock
{\BBOQ}\APACrefatitle {{Weisfeiler} and {Leman} go neural: Higher-order graph neural networks} {{Weisfeiler} and {Leman} go neural: Higher-order graph neural networks}.{\BBCQ}
\newblock
 \APACrefbtitle {{AAAI Conference on Artificial Intelligence}} {{AAAI Conference on Artificial Intelligence}}\ (\BVOL~33, \BPGS\ 4602--4609).
\PrintBackRefs{\CurrentBib}

\bibitem [\protect \citeauthoryear {%
Neuhaus%
, Riesen%
\BCBL {}\ \BBA {} Bunke%
}{%
Neuhaus%
\ \protect \BOthers {.}}{%
{\protect \APACyear {2006}}%
}]{%
beam_ged2}
\APACinsertmetastar {%
beam_ged2}%
\begin{APACrefauthors}%
Neuhaus, M.%
, Riesen, K.%
\BCBL {} Bunke, H.%
\end{APACrefauthors}%
\unskip\
\newblock
\APACrefYearMonthDay{2006}{}{}.
\newblock
{\BBOQ}\APACrefatitle {Fast suboptimal algorithms for the computation of graph edit distance} {Fast suboptimal algorithms for the computation of graph edit distance}.{\BBCQ}
\newblock
 \APACrefbtitle {{Joint IAPR International Workshops on Structural, Syntactic, and Statistical Pattern Recognition}} {{Joint IAPR International Workshops on Structural, Syntactic, and Statistical Pattern Recognition}}\ (\BPGS\ 163--172).
\PrintBackRefs{\CurrentBib}

\bibitem [\protect \citeauthoryear {%
Paszke%
\ \protect \BOthers {.}}{%
Paszke%
\ \protect \BOthers {.}}{%
{\protect \APACyear {2019}}%
}]{%
pytorch}
\APACinsertmetastar {%
pytorch}%
\begin{APACrefauthors}%
Paszke, A.%
, Gross, S.%
, Massa, F.%
, Lerer, A.%
, Bradbury, J.%
, Chanan, G.%
\BDBL {}Chintala, S.%
\end{APACrefauthors}%
\unskip\
\newblock
\APACrefYearMonthDay{2019}{}{}.
\newblock
{\BBOQ}\APACrefatitle {{PyTorch}: An imperative style, high-performance deep learning library} {{PyTorch}: An imperative style, high-performance deep learning library}.{\BBCQ}
\newblock
 \APACrefbtitle {{Advances in Neural Information Processing Systems}} {{Advances in Neural Information Processing Systems}}\ (\BVOL~32, \BPGS\ 8024--8035).
\PrintBackRefs{\CurrentBib}

\bibitem [\protect \citeauthoryear {%
Perez%
, Strub%
, de Vries%
, Dumoulin%
\BCBL {}\ \BBA {} Courville%
}{%
Perez%
\ \protect \BOthers {.}}{%
{\protect \APACyear {2018}}%
}]{%
FiLM}
\APACinsertmetastar {%
FiLM}%
\begin{APACrefauthors}%
Perez, E.%
, Strub, F.%
, de Vries, H.%
, Dumoulin, V.%
\BCBL {} Courville, A.%
\end{APACrefauthors}%
\unskip\
\newblock
\APACrefYearMonthDay{2018}{}{}.
\newblock
{\BBOQ}\APACrefatitle {{FiLM}: Visual reasoning with a general conditioning layer} {{FiLM}: Visual reasoning with a general conditioning layer}.{\BBCQ}
\newblock
 \APACrefbtitle {{AAAI Conference on Artificial Intelligence}} {{AAAI Conference on Artificial Intelligence}}\ (\BVOL~32).
\PrintBackRefs{\CurrentBib}

\bibitem [\protect \citeauthoryear {%
Piao%
\ \protect \BOthers {.}}{%
Piao%
\ \protect \BOthers {.}}{%
{\protect \APACyear {2023}}%
}]{%
GEDGNN}
\APACinsertmetastar {%
GEDGNN}%
\begin{APACrefauthors}%
Piao, C.%
, Xu, T.%
, Sun, X.%
, Rong, Y.%
, Zhao, K.%
\BCBL {} Cheng, H.%
\end{APACrefauthors}%
\unskip\
\newblock
\APACrefYearMonthDay{2023}{}{}.
\newblock
{\BBOQ}\APACrefatitle {Computing graph edit distance via neural graph matching} {Computing graph edit distance via neural graph matching}.{\BBCQ}
\newblock
\APACjournalVolNumPages{Proceedings of the VLDB Endowment}{16}{8}{1817--1829,}
\newblock
\begin{APACrefDOI} \doi{10.14778/3594512.3594514} \end{APACrefDOI}
\newblock

\newblock

\PrintBackRefs{\CurrentBib}

\bibitem [\protect \citeauthoryear {%
Qin%
\ \protect \BOthers {.}}{%
Qin%
\ \protect \BOthers {.}}{%
{\protect \APACyear {2021}}%
}]{%
EGSC}
\APACinsertmetastar {%
EGSC}%
\begin{APACrefauthors}%
Qin, C.%
, Zhao, H.%
, Wang, L.%
, Wang, H.%
, Zhang, Y.%
\BCBL {} Fu, Y.%
\end{APACrefauthors}%
\unskip\
\newblock
\APACrefYearMonthDay{2021}{}{}.
\newblock
{\BBOQ}\APACrefatitle {Slow learning and fast inference: Efficient graph similarity computation via knowledge distillation} {Slow learning and fast inference: Efficient graph similarity computation via knowledge distillation}.{\BBCQ}
\newblock
 \APACrefbtitle {{Advances in Neural Information Processing Systems}} {{Advances in Neural Information Processing Systems}}\ (\BVOL~34, \BPGS\ 14110--14121).
\PrintBackRefs{\CurrentBib}

\bibitem [\protect \citeauthoryear {%
Ranjan%
\ \protect \BOthers {.}}{%
Ranjan%
\ \protect \BOthers {.}}{%
{\protect \APACyear {2022}}%
}]{%
GREED}
\APACinsertmetastar {%
GREED}%
\begin{APACrefauthors}%
Ranjan, R.%
, Grover, S.%
, Medya, S.%
, Chakaravarthy, V.%
, Sabharwal, Y.%
\BCBL {} Ranu, S.%
\end{APACrefauthors}%
\unskip\
\newblock
\APACrefYearMonthDay{2022}{}{}.
\newblock
{\BBOQ}\APACrefatitle {{GREED}: A neural framework for learning graph distance functions} {{GREED}: A neural framework for learning graph distance functions}.{\BBCQ}
\newblock
 \APACrefbtitle {{Advances in Neural Information Processing Systems}} {{Advances in Neural Information Processing Systems}}\ (\BVOL~35, \BPGS\ 22518--22530).
\PrintBackRefs{\CurrentBib}

\bibitem [\protect \citeauthoryear {%
Riesen%
\ \BBA {} Bunke%
}{%
Riesen%
\ \BBA {} Bunke%
}{%
{\protect \APACyear {2009}}%
}]{%
beam_ged}
\APACinsertmetastar {%
beam_ged}%
\begin{APACrefauthors}%
Riesen, K.%
\BCBT {}\ \BBA {} Bunke, H.%
\end{APACrefauthors}%
\unskip\
\newblock
\APACrefYearMonthDay{2009}{}{}.
\newblock
{\BBOQ}\APACrefatitle {Approximate graph edit distance computation by means of bipartite graph matching} {Approximate graph edit distance computation by means of bipartite graph matching}.{\BBCQ}
\newblock
\APACjournalVolNumPages{Image and Vision Computing}{27}{7}{950--959,}
\newblock
\begin{APACrefDOI} \doi{10.1016/j.imavis.2008.04.004} \end{APACrefDOI}
\newblock

\newblock

\PrintBackRefs{\CurrentBib}

\bibitem [\protect \citeauthoryear {%
Riesen%
, Fankhauser%
\BCBL {}\ \BBA {} Bunke%
}{%
Riesen%
\ \protect \BOthers {.}}{%
{\protect \APACyear {2007}}%
}]{%
astar_ged}
\APACinsertmetastar {%
astar_ged}%
\begin{APACrefauthors}%
Riesen, K.%
, Fankhauser, S.%
\BCBL {} Bunke, H.%
\end{APACrefauthors}%
\unskip\
\newblock
\APACrefYearMonthDay{2007}{}{}.
\newblock
{\BBOQ}\APACrefatitle {Speeding up graph edit distance computation with a bipartite heuristic} {Speeding up graph edit distance computation with a bipartite heuristic}.{\BBCQ}
\newblock
 \APACrefbtitle {{Mining and Learning with Graphs}} {{Mining and Learning with Graphs}}\ (\BPGS\ 21--24).
\PrintBackRefs{\CurrentBib}

\bibitem [\protect \citeauthoryear {%
Scarselli%
, Gori%
, Tsoi%
, Hagenbuchner%
\BCBL {}\ \BBA {} Monfardini%
}{%
Scarselli%
\ \protect \BOthers {.}}{%
{\protect \APACyear {2009}}%
}]{%
gnn_2}
\APACinsertmetastar {%
gnn_2}%
\begin{APACrefauthors}%
Scarselli, F.%
, Gori, M.%
, Tsoi, A.C.%
, Hagenbuchner, M.%
\BCBL {} Monfardini, G.%
\end{APACrefauthors}%
\unskip\
\newblock
\APACrefYearMonthDay{2009}{}{}.
\newblock
{\BBOQ}\APACrefatitle {The graph neural network model} {The graph neural network model}.{\BBCQ}
\newblock
\APACjournalVolNumPages{IEEE Transactions on Neural Networks}{20}{1}{61--80,}
\newblock
\begin{APACrefDOI} \doi{10.1109/TNN.2008.2005605} \end{APACrefDOI}
\newblock

\newblock

\PrintBackRefs{\CurrentBib}

\bibitem [\protect \citeauthoryear {%
Shchur%
, Mumme%
, Bojchevski%
\BCBL {}\ \BBA {} G{\"u}nnemann%
}{%
Shchur%
\ \protect \BOthers {.}}{%
{\protect \APACyear {2018}}%
}]{%
amazon_datasets}
\APACinsertmetastar {%
amazon_datasets}%
\begin{APACrefauthors}%
Shchur, O.%
, Mumme, M.%
, Bojchevski, A.%
\BCBL {} G{\"u}nnemann, S.%
\end{APACrefauthors}%
\unskip\
\newblock
\APACrefYearMonthDay{2018}{}{}.
\newblock
\APACrefbtitle {Pitfalls of graph neural network evaluation.} {Pitfalls of graph neural network evaluation.}
\newblock
\APAChowpublished {arXiv preprint 1811.05868}.
\PrintBackRefs{\CurrentBib}

\bibitem [\protect \citeauthoryear {%
Shrivastava%
\ \BBA {} Li%
}{%
Shrivastava%
\ \BBA {} Li%
}{%
{\protect \APACyear {2014}}%
}]{%
ptc}
\APACinsertmetastar {%
ptc}%
\begin{APACrefauthors}%
Shrivastava, A.%
\BCBT {}\ \BBA {} Li, P.%
\end{APACrefauthors}%
\unskip\
\newblock
\APACrefYearMonthDay{2014}{}{}.
\newblock
{\BBOQ}\APACrefatitle {A new space for comparing graphs} {A new space for comparing graphs}.{\BBCQ}
\newblock
 \APACrefbtitle {{IEEE/ACM International Conference on Advances in Social Networks Analysis and Mining}} {{IEEE/ACM International Conference on Advances in Social Networks Analysis and Mining}}\ (\BPGS\ 62--71).
\PrintBackRefs{\CurrentBib}

\bibitem [\protect \citeauthoryear {%
Socher%
, Chen%
, Manning%
\BCBL {}\ \BBA {} Ng%
}{%
Socher%
\ \protect \BOthers {.}}{%
{\protect \APACyear {2013}}%
}]{%
NTN}
\APACinsertmetastar {%
NTN}%
\begin{APACrefauthors}%
Socher, R.%
, Chen, D.%
, Manning, C.D.%
\BCBL {} Ng, A.%
\end{APACrefauthors}%
\unskip\
\newblock
\APACrefYearMonthDay{2013}{}{}.
\newblock
{\BBOQ}\APACrefatitle {Reasoning with neural tensor networks for knowledge base completion} {Reasoning with neural tensor networks for knowledge base completion}.{\BBCQ}
\newblock
 \APACrefbtitle {{Advances in Neural Information Processing Systems}} {{Advances in Neural Information Processing Systems}}\ (\BVOL~26, \BPGS\ 926--934).
\PrintBackRefs{\CurrentBib}

\bibitem [\protect \citeauthoryear {%
Spearman%
}{%
Spearman%
}{%
{\protect \APACyear {1904}}%
}]{%
spearman}
\APACinsertmetastar {%
spearman}%
\begin{APACrefauthors}%
Spearman, C.%
\end{APACrefauthors}%
\unskip\
\newblock
\APACrefYearMonthDay{1904}{}{}.
\newblock
{\BBOQ}\APACrefatitle {The proof and measurement of association between two things} {The proof and measurement of association between two things}.{\BBCQ}
\newblock
\APACjournalVolNumPages{The American Journal of Psychology}{15}{}{72--101,}
\newblock
\begin{APACrefDOI} \doi{10.2307/1412159} \end{APACrefDOI}
\newblock

\newblock

\PrintBackRefs{\CurrentBib}

\bibitem [\protect \citeauthoryear {%
Tang%
, Zhao%
, Kong%
, Zhou%
\BCBL {}\ \BBA {} Li%
}{%
Tang%
\ \protect \BOthers {.}}{%
{\protect \APACyear {2025}}%
}]{%
FGWAlign}
\APACinsertmetastar {%
FGWAlign}%
\begin{APACrefauthors}%
Tang, J.%
, Zhao, X.%
, Kong, L.%
, Zhou, X.%
\BCBL {} Li, J.%
\end{APACrefauthors}%
\unskip\
\newblock
\APACrefYearMonthDay{2025}{}{}.
\newblock
{\BBOQ}\APACrefatitle {Fused {Gromov-Wasserstein} alignment for graph edit distance computation and beyond} {Fused {Gromov-Wasserstein} alignment for graph edit distance computation and beyond}.{\BBCQ}
\newblock
\APACjournalVolNumPages{Proceedings of the VLDB Endowment}{18}{10}{3641--3654,}
\newblock
\begin{APACrefDOI} \doi{10.14778/3748191.3748221} \end{APACrefDOI}
\newblock

\newblock

\PrintBackRefs{\CurrentBib}

\bibitem [\protect \citeauthoryear {%
Virtanen%
\ \protect \BOthers {.}}{%
Virtanen%
\ \protect \BOthers {.}}{%
{\protect \APACyear {2020}}%
}]{%
scipy}
\APACinsertmetastar {%
scipy}%
\begin{APACrefauthors}%
Virtanen, P.%
, Gommers, R.%
, Oliphant, T.E.%
, Haberland, M.%
, Reddy, T.%
, Cournapeau, D.%
\BDBL {}{SciPy 1.0 Contributors}%
\end{APACrefauthors}%
\unskip\
\newblock
\APACrefYearMonthDay{2020}{}{}.
\newblock
{\BBOQ}\APACrefatitle {{SciPy} 1.0: Fundamental algorithms for scientific computing in {Python}} {{SciPy} 1.0: Fundamental algorithms for scientific computing in {Python}}.{\BBCQ}
\newblock
\APACjournalVolNumPages{Nature Methods}{17}{3}{261--272,}
\newblock
\begin{APACrefDOI} \doi{10.1038/s41592-019-0686-2} \end{APACrefDOI}
\newblock

\newblock

\PrintBackRefs{\CurrentBib}

\bibitem [\protect \citeauthoryear {%
R.~Wang%
, Zhang%
, Yu%
, Yan%
\BCBL {}\ \BBA {} Yang%
}{%
R.~Wang%
\ \protect \BOthers {.}}{%
{\protect \APACyear {2021}}%
}]{%
GENN}
\APACinsertmetastar {%
GENN}%
\begin{APACrefauthors}%
Wang, R.%
, Zhang, T.%
, Yu, T.%
, Yan, J.%
\BCBL {} Yang, X.%
\end{APACrefauthors}%
\unskip\
\newblock
\APACrefYearMonthDay{2021}{}{}.
\newblock
{\BBOQ}\APACrefatitle {Combinatorial learning of graph edit distance via dynamic embedding} {Combinatorial learning of graph edit distance via dynamic embedding}.{\BBCQ}
\newblock
 \APACrefbtitle {{IEEE Conference on Computer Vision and Pattern Recognition}} {{IEEE Conference on Computer Vision and Pattern Recognition}}\ (\BPGS\ 5237--5246).
\PrintBackRefs{\CurrentBib}

\bibitem [\protect \citeauthoryear {%
X.~Wang%
, Ding%
, Tung%
, Ying%
\BCBL {}\ \BBA {} Jin%
}{%
X.~Wang%
\ \protect \BOthers {.}}{%
{\protect \APACyear {2012}}%
}]{%
aids_linux}
\APACinsertmetastar {%
aids_linux}%
\begin{APACrefauthors}%
Wang, X.%
, Ding, X.%
, Tung, A.K.H.%
, Ying, S.%
\BCBL {} Jin, H.%
\end{APACrefauthors}%
\unskip\
\newblock
\APACrefYearMonthDay{2012}{}{}.
\newblock
{\BBOQ}\APACrefatitle {An efficient graph indexing method} {An efficient graph indexing method}.{\BBCQ}
\newblock
 \APACrefbtitle {{IEEE International Conference on Data Engineering}} {{IEEE International Conference on Data Engineering}}\ (\BPGS\ 210--221).
\PrintBackRefs{\CurrentBib}

\bibitem [\protect \citeauthoryear {%
Witten%
, Frank%
\BCBL {}\ \BBA {} Hall%
}{%
Witten%
\ \protect \BOthers {.}}{%
{\protect \APACyear {2011}}%
}]{%
witten}
\APACinsertmetastar {%
witten}%
\begin{APACrefauthors}%
Witten, I.H.%
, Frank, E.%
\BCBL {} Hall, M.A.%
\end{APACrefauthors}%
\unskip\
\newblock
\APACrefYear{2011}.
\newblock
\APACrefbtitle {Data mining: Practical machine learning tools and techniques} {Data mining: Practical machine learning tools and techniques}\ (\PrintOrdinal{3rd}\ \BEd).
\newblock
\APACaddressPublisher{}{Morgan Kaufmann}.
\PrintBackRefs{\CurrentBib}

\bibitem [\protect \citeauthoryear {%
K.~Xu%
, Hu%
, Leskovec%
\BCBL {}\ \BBA {} Jegelka%
}{%
K.~Xu%
\ \protect \BOthers {.}}{%
{\protect \APACyear {2019}}%
}]{%
GIN}
\APACinsertmetastar {%
GIN}%
\begin{APACrefauthors}%
Xu, K.%
, Hu, W.%
, Leskovec, J.%
\BCBL {} Jegelka, S.%
\end{APACrefauthors}%
\unskip\
\newblock
\APACrefYearMonthDay{2019}{}{}.
\newblock
{\BBOQ}\APACrefatitle {How powerful are graph neural networks?} {How powerful are graph neural networks?}{\BBCQ}
\newblock
 \APACrefbtitle {{International Conference on Learning Representations}.} {{International Conference on Learning Representations}.}
\PrintBackRefs{\CurrentBib}

\bibitem [\protect \citeauthoryear {%
M.~Xu%
\ \BBA {} Chang%
}{%
M.~Xu%
\ \BBA {} Chang%
}{%
{\protect \APACyear {2025}}%
}]{%
XuChang}
\APACinsertmetastar {%
XuChang}%
\begin{APACrefauthors}%
Xu, M.%
\BCBT {}\ \BBA {} Chang, L.%
\end{APACrefauthors}%
\unskip\
\newblock
\APACrefYearMonthDay{2025}{}{}.
\newblock
{\BBOQ}\APACrefatitle {Graph edit distance estimation: A new heuristic and a holistic evaluation of learning-based methods} {Graph edit distance estimation: A new heuristic and a holistic evaluation of learning-based methods}.{\BBCQ}
\newblock
\APACjournalVolNumPages{Proceedings of the ACM on Management of Data}{3}{3}{1--24,}
\newblock
\begin{APACrefDOI} \doi{10.1145/3725304} \end{APACrefDOI}
\newblock

\newblock

\PrintBackRefs{\CurrentBib}

\bibitem [\protect \citeauthoryear {%
Yanardag%
\ \BBA {} Vishwanathan%
}{%
Yanardag%
\ \BBA {} Vishwanathan%
}{%
{\protect \APACyear {2015}}%
}]{%
imdb}
\APACinsertmetastar {%
imdb}%
\begin{APACrefauthors}%
Yanardag, P.%
\BCBT {}\ \BBA {} Vishwanathan, S.V.N.%
\end{APACrefauthors}%
\unskip\
\newblock
\APACrefYearMonthDay{2015}{}{}.
\newblock
{\BBOQ}\APACrefatitle {Deep graph kernels} {Deep graph kernels}.{\BBCQ}
\newblock
 \APACrefbtitle {{ACM SIGKDD International Conference on Knowledge Discovery and Data Mining}} {{ACM SIGKDD International Conference on Knowledge Discovery and Data Mining}}\ (\BPGS\ 1365--1374).
\PrintBackRefs{\CurrentBib}

\bibitem [\protect \citeauthoryear {%
J.~Yang%
\ \BBA {} Leskovec%
}{%
J.~Yang%
\ \BBA {} Leskovec%
}{%
{\protect \APACyear {2012}}%
}]{%
snap_datasets}
\APACinsertmetastar {%
snap_datasets}%
\begin{APACrefauthors}%
Yang, J.%
\BCBT {}\ \BBA {} Leskovec, J.%
\end{APACrefauthors}%
\unskip\
\newblock
\APACrefYearMonthDay{2012}{}{}.
\newblock
{\BBOQ}\APACrefatitle {Defining and evaluating network communities based on ground-truth} {Defining and evaluating network communities based on ground-truth}.{\BBCQ}
\newblock
 \APACrefbtitle {{IEEE International Conference on Data Mining}} {{IEEE International Conference on Data Mining}}\ (\BPGS\ 745--754).
\PrintBackRefs{\CurrentBib}

\bibitem [\protect \citeauthoryear {%
L.~Yang%
\ \BBA {} Zou%
}{%
L.~Yang%
\ \BBA {} Zou%
}{%
{\protect \APACyear {2021}}%
}]{%
Noah}
\APACinsertmetastar {%
Noah}%
\begin{APACrefauthors}%
Yang, L.%
\BCBT {}\ \BBA {} Zou, L.%
\end{APACrefauthors}%
\unskip\
\newblock
\APACrefYearMonthDay{2021}{}{}.
\newblock
{\BBOQ}\APACrefatitle {{Noah}: Neural-optimized {A*} search algorithm for graph edit distance computation} {{Noah}: Neural-optimized {A*} search algorithm for graph edit distance computation}.{\BBCQ}
\newblock
 \APACrefbtitle {{IEEE International Conference on Data Engineering}} {{IEEE International Conference on Data Engineering}}\ (\BPGS\ 576--587).
\PrintBackRefs{\CurrentBib}

\bibitem [\protect \citeauthoryear {%
P.~Yang%
\ \protect \BOthers {.}}{%
P.~Yang%
\ \protect \BOthers {.}}{%
{\protect \APACyear {2024}}%
}]{%
yang2024deep}
\APACinsertmetastar {%
yang2024deep}%
\begin{APACrefauthors}%
Yang, P.%
, Wang, H.%
, Yang, J.%
, Qian, Z.%
, Zhang, Y.%
\BCBL {} Lin, X.%
\end{APACrefauthors}%
\unskip\
\newblock
\APACrefYearMonthDay{2024}{}{}.
\newblock
{\BBOQ}\APACrefatitle {Deep learning approaches for similarity computation: A survey} {Deep learning approaches for similarity computation: A survey}.{\BBCQ}
\newblock
\APACjournalVolNumPages{IEEE Transactions on Knowledge and Data Engineering}{36}{12}{7893--7912,}
\newblock
\begin{APACrefDOI} \doi{10.1109/TKDE.2024.3422484} \end{APACrefDOI}
\newblock

\newblock

\PrintBackRefs{\CurrentBib}

\bibitem [\protect \citeauthoryear {%
R.~Yang%
\ \protect \BOthers {.}}{%
R.~Yang%
\ \protect \BOthers {.}}{%
{\protect \APACyear {2020}}%
}]{%
attributed_datasets}
\APACinsertmetastar {%
attributed_datasets}%
\begin{APACrefauthors}%
Yang, R.%
, Shi, J.%
, Xiao, X.%
, Yang, Y.%
, Liu, J.%
\BCBL {} Bhowmick, S.S.%
\end{APACrefauthors}%
\unskip\
\newblock
\APACrefYearMonthDay{2020}{}{}.
\newblock
{\BBOQ}\APACrefatitle {Scaling attributed network embedding to massive graphs} {Scaling attributed network embedding to massive graphs}.{\BBCQ}
\newblock
\APACjournalVolNumPages{Proceedings of the VLDB Endowment}{14}{1}{37--49,}
\newblock
\begin{APACrefDOI} \doi{10.14778/3421424.3421430} \end{APACrefDOI}
\newblock

\newblock

\PrintBackRefs{\CurrentBib}

\bibitem [\protect \citeauthoryear {%
Z.~Yang%
, Cohen%
\BCBL {}\ \BBA {} Salakhutdinov%
}{%
Z.~Yang%
\ \protect \BOthers {.}}{%
{\protect \APACyear {2016}}%
}]{%
planetoid_datasets}
\APACinsertmetastar {%
planetoid_datasets}%
\begin{APACrefauthors}%
Yang, Z.%
, Cohen, W.%
\BCBL {} Salakhutdinov, R.%
\end{APACrefauthors}%
\unskip\
\newblock
\APACrefYearMonthDay{2016}{}{}.
\newblock
{\BBOQ}\APACrefatitle {Revisiting semi-supervised learning with graph embeddings} {Revisiting semi-supervised learning with graph embeddings}.{\BBCQ}
\newblock
 \APACrefbtitle {{International Conference on Machine Learning}} {{International Conference on Machine Learning}}\ (\BPGS\ 40--48).
\PrintBackRefs{\CurrentBib}

\bibitem [\protect \citeauthoryear {%
Yu%
, Liu%
, Fang%
\BCBL {}\ \BBA {} Zhang%
}{%
Yu%
\ \protect \BOthers {.}}{%
{\protect \APACyear {2023}}%
}]{%
CountGNN}
\APACinsertmetastar {%
CountGNN}%
\begin{APACrefauthors}%
Yu, X.%
, Liu, Z.%
, Fang, Y.%
\BCBL {} Zhang, X.%
\end{APACrefauthors}%
\unskip\
\newblock
\APACrefYearMonthDay{2023}{}{}.
\newblock
{\BBOQ}\APACrefatitle {Learning to count isomorphisms with graph neural networks} {Learning to count isomorphisms with graph neural networks}.{\BBCQ}
\newblock
 \APACrefbtitle {{AAAI Conference on Artificial Intelligence}} {{AAAI Conference on Artificial Intelligence}}\ (\BVOL~37, \BPGS\ 4845--4853).
\PrintBackRefs{\CurrentBib}

\bibitem [\protect \citeauthoryear {%
Zeng%
, Tung%
, Wang%
, Feng%
\BCBL {}\ \BBA {} Zhou%
}{%
Zeng%
\ \protect \BOthers {.}}{%
{\protect \APACyear {2009}}%
}]{%
aids_1}
\APACinsertmetastar {%
aids_1}%
\begin{APACrefauthors}%
Zeng, Z.%
, Tung, A.K.H.%
, Wang, J.%
, Feng, J.%
\BCBL {} Zhou, L.%
\end{APACrefauthors}%
\unskip\
\newblock
\APACrefYearMonthDay{2009}{}{}.
\newblock
{\BBOQ}\APACrefatitle {Comparing stars: On approximating graph edit distance} {Comparing stars: On approximating graph edit distance}.{\BBCQ}
\newblock
\APACjournalVolNumPages{Proceedings of the VLDB Endowment}{2}{1}{25--36,}
\newblock
\begin{APACrefDOI} \doi{10.14778/1687627.1687631} \end{APACrefDOI}
\newblock

\newblock

\PrintBackRefs{\CurrentBib}

\bibitem [\protect \citeauthoryear {%
Zheng%
, Shi%
\BCBL {}\ \BBA {} Yang%
}{%
Zheng%
\ \protect \BOthers {.}}{%
{\protect \APACyear {2025}}%
}]{%
GraSP}
\APACinsertmetastar {%
GraSP}%
\begin{APACrefauthors}%
Zheng, H.%
, Shi, J.%
\BCBL {} Yang, R.%
\end{APACrefauthors}%
\unskip\
\newblock
\APACrefYearMonthDay{2025}{}{}.
\newblock
{\BBOQ}\APACrefatitle {{GraSP}: Simple yet effective graph similarity predictions} {{GraSP}: Simple yet effective graph similarity predictions}.{\BBCQ}
\newblock
 \APACrefbtitle {{AAAI Conference on Artificial Intelligence}} {{AAAI Conference on Artificial Intelligence}}\ (\BVOL~39, \BPGS\ 22884--22892).
\PrintBackRefs{\CurrentBib}

\bibitem [\protect \citeauthoryear {%
Zhou%
\ \protect \BOthers {.}}{%
Zhou%
\ \protect \BOthers {.}}{%
{\protect \APACyear {2020}}%
}]{%
gnn_paper}
\APACinsertmetastar {%
gnn_paper}%
\begin{APACrefauthors}%
Zhou, J.%
, Cui, G.%
, Hu, S.%
, Zhang, Z.%
, Yang, C.%
, Liu, Z.%
\BDBL {}Sun, M.%
\end{APACrefauthors}%
\unskip\
\newblock
\APACrefYearMonthDay{2020}{}{}.
\newblock
{\BBOQ}\APACrefatitle {Graph neural networks: A review of methods and applications} {Graph neural networks: A review of methods and applications}.{\BBCQ}
\newblock
\APACjournalVolNumPages{AI Open}{1}{}{57--81,}
\newblock
\begin{APACrefDOI} \doi{10.1016/j.aiopen.2021.01.001} \end{APACrefDOI}
\newblock

\newblock

\PrintBackRefs{\CurrentBib}

\bibitem [\protect \citeauthoryear {%
Zhuo%
\ \BBA {} Tan%
}{%
Zhuo%
\ \BBA {} Tan%
}{%
{\protect \APACyear {2022}}%
}]{%
ERIC}
\APACinsertmetastar {%
ERIC}%
\begin{APACrefauthors}%
Zhuo, W.%
\BCBT {}\ \BBA {} Tan, G.%
\end{APACrefauthors}%
\unskip\
\newblock
\APACrefYearMonthDay{2022}{}{}.
\newblock
{\BBOQ}\APACrefatitle {Efficient graph similarity computation with alignment regularization} {Efficient graph similarity computation with alignment regularization}.{\BBCQ}
\newblock
 \APACrefbtitle {{Advances in Neural Information Processing Systems}} {{Advances in Neural Information Processing Systems}}\ (\BVOL~35, \BPGS\ 30181--30193).
\PrintBackRefs{\CurrentBib}

\end{thebibliography}
